\documentclass[conference]{IEEEtran}
\usepackage{graphicx}
\usepackage{float}
\usepackage{subfigure}
\usepackage{algorithm}
\usepackage{algorithmicx}
\usepackage{algpseudocode}
\usepackage{multirow}
\usepackage{colortbl} 
\usepackage{soul,color}
\usepackage{hyperref}
\newcommand{\name}{\texttt{Decaf}}
\usepackage{tikz}
\usepackage{amsmath}
\usepackage[framemethod=TikZ]{mdframed}

\ifCLASSINFOpdf
\else
\fi
\begin{document}
%
\title{\name: Data \underline{D}istribution D\underline{ec}ompose \underline{A}ttack \\ against \underline{F}ederated Learning}


\author{\IEEEauthorblockN{Zhiyang Dai}
\IEEEauthorblockA{Nanjing University of Science and Technology\\
dzy@njust.edu.cn}
\and
\IEEEauthorblockN{Chunyi Zhou}
\IEEEauthorblockA{Nanjing University of Science and Technology\\
zhouchunyi@njust.edu.cn}
\and
\IEEEauthorblockN{Anmin Fu}
\IEEEauthorblockA{School of Computer Science and Engineering,\\ Nanjing University of Science and Technology\\
fuam@njust.edu.cn}}



%


\IEEEoverridecommandlockouts
\makeatletter\def\@IEEEpubidpullup{6.5\baselineskip}\makeatother

\maketitle

\begin{abstract}
In contrast to prevalent Federated Learning (FL) privacy inference techniques such as generative adversarial networks attacks, membership inference attacks, property inference attacks, and model inversion attacks, we devise an innovative privacy threat: the Data Distribution Decompose Attack on FL, termed \texttt{Decaf}. This attack enables an honest-but-curious FL server to meticulously profile the proportion of each class owned by the victim FL user, divulging sensitive information like local market item distribution and business competitiveness. The crux of \texttt{Decaf} lies in the profound observation that the magnitude of local model gradient changes closely mirrors the underlying data distribution, including the proportion of each class. \texttt{Decaf} addresses two crucial challenges: accurately identify the missing/null class(es) given by any victim user as a premise and then quantify the precise relationship between gradient changes and each remaining non-null class. Notably, \texttt{Decaf} operates stealthily, rendering it entirely passive and undetectable to victim users regarding the infringement of their data distribution privacy. Experimental validation on five benchmark datasets (MNIST, FASHION-MNIST, CIFAR-10, FER-2013, and SkinCancer) employing diverse model architectures, including customized convolutional networks, standardized VGG16, and ResNet18, demonstrates \texttt{Decaf}'s efficacy. Results indicate its ability to accurately decompose local user data distribution, regardless of whether it is IID or non-IID distributed. Specifically, the dissimilarity measured using $L_{\infty}$ distance between the distribution decomposed by \texttt{Decaf} and ground truth is consistently below 5\% when no null classes exist. Moreover, \texttt{Decaf} achieves 100\% accuracy in determining any victim user's null classes, validated through formal proof.
\end{abstract}


%

\section{Introduction}
Opposed to centralized deep learning (DL), federated learning (FL) entitles users to train models locally without gathering their private data into the server, significantly reducing the privacy encroaching on user data.
It is thus attracting ever-increasing attention in various fields, especially privacy-sensitive applications such as product recommendation systems \cite{46DBLP:series/lncs/YangTZCY20, 47DBLP:conf/icsim/QinLQ21}, social networks \cite{48DBLP:conf/sdm/YangC19, 49DBLP:conf/hicss/AyoraHK21, 50DBLP:conf/sac/AnelliDNFN21}, and medical detection systems \cite{51DBLP:journals/titb/HossenPKABI23, 52DBLP:journals/titb/AouediSPM23, 53DBLP:journals/bdcc/SinghVF23}.
\begin{figure}[h]
  \centering
  \includegraphics[width=\linewidth]{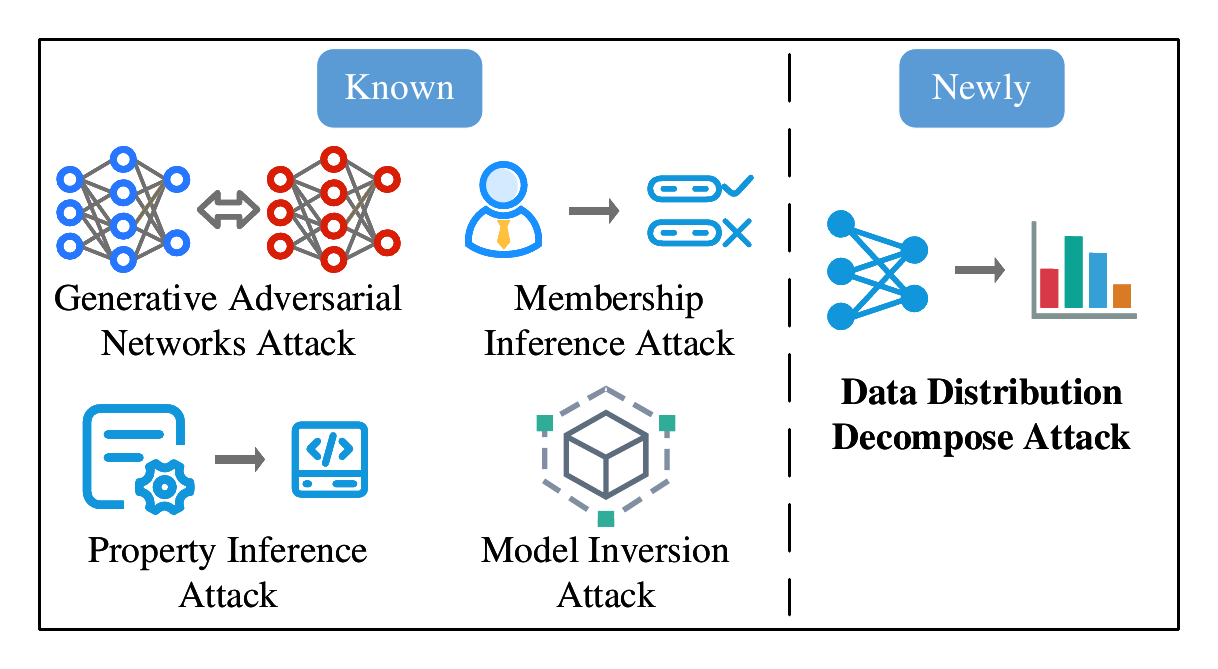}
  \caption{Privacy inference attacks against FL.}
  \label{Fig_privacy_attacks}
\end{figure}

However, FL is frustrating to completely mitigate privacy leakage risks due to advanced inference attacks\cite{31DBLP:conf/sp/NasrSH19}.  
As summarized in \autoref{Fig_privacy_attacks}, existing mainstream FL privacy inference attacks are Generative Adversarial Network (GAN) attacks, membership inference attacks, property inference attacks, and model inversion attacks, which can respectively expose various private information of the victim user. 
The GAN attack \cite{4DBLP:conf/ccs/HitajAP17, 56DBLP:conf/infocom/WangSZSWQ19} results in that any attacker disguised as a general participant can reconstruct the sensitive information of the local user from the victim's local data by exploiting the GAN technique. 
The membership inference attack \cite{9DBLP:conf/sp/ShokriSSS17, 10DBLP:conf/csfw/YeomGFJ18, 11DBLP:conf/ndss/Salem0HBF019, 12DBLP:conf/uss/SongM21, 13DBLP:conf/icml/Choquette-ChooT21, 14DBLP:conf/ccs/LiZ21, 15DBLP:conf/sp/ChenJW20, 16DBLP:conf/cvpr/LiXZYL20} aims to infer whether a particular data sample record is contained in a model's training dataset. 
The property inference attack \cite{17DBLP:conf/ccs/GanjuWYGB18, 18DBLP:conf/sp/MelisSCS19, 19DBLP:conf/ndss/Zhou00022, 20DBLP:conf/ccs/WangW22} can infer properties of training data, even if the properties are irrelevant to the main FL learning task. 
Model inversion attack \cite{21DBLP:conf/ccs/FredriksonJR15, 22DBLP:conf/uss/TramerZJRR16, 22DBLP:conf/uss/TramerZJRR16, 23DBLP:journals/tifs/KhosravyNHNB22, 24DBLP:journals/tifs/ZhuYZLZ23, 25DBLP:conf/cvpr/KahlaCJJ22} can restore the ground-truth labels and confidence coefficients by querying a large amount of predictive data to the target model, and then reconstruct the training dataset.

\noindent\textbf{Limitation of Existing Attacks}. None of those above mainstream FL privacy inference attacks is capable of inferring the local user data distribution or composition amongst one of the most interesting private information \cite{chaudhari2023snap}, e.g., the composition of different sold items of a store, bought items by a customer (e.g., FASHION-MNIST), one's personality (e.g., FER-2013), gender of an institution, fraction of cancer patients (e.g., SkinCancer).  
The data composition is the proportion per class given a user's local dataset. To ease understanding, assume that a user has three classes: 50 samples of class A, 50 samples of class B, and 100 samples of class C. The data composition is A:B:C = 0.25:0.25:0.5. Exposure to data distribution puts the victim under disadvantageous business competition, for example, different items sold in a store, and reputation/political concerns, for example, the gender rate of employees. 

In this context, we ask the following research questions:

\begin{mdframed}[backgroundcolor=black!10,rightline=false,leftline=false,topline=false,bottomline=false,roundcorner=2mm]
Is it practical to stealthily decompose the private data distribution of the FL local user? If so, to what extent has such private information been leaked?
\end{mdframed}


\noindent\textbf{Data Distribution Decompose Attack:} We reveal a new type of FL privacy inference attack, a data distribution decompose attack, dubbed \texttt{Decaf}, to infer the victim user's local data distribution information accurately. The \texttt{Decaf} is orthogonal to other mainstream attacks as shown in \autoref{Fig_privacy_attacks}. Fulfilling \texttt{Decaf} is, however, non-trivial, which confronts two significant challenges (\textbf{Cs}):
\begin{itemize}
\item \textbf{C1:} How do we first identify and remove FL users' potential null classes to avoid confusion following data distribution decomposition on remaining classes?
\item \textbf{C2:} How do we then accurately decompose the remaining non-null classes' distribution upon the victim's local model update?
\end{itemize}

The \textbf{C1} is raised because local data can be non-IID distributed. Each user unnecessarily owns data samples from all classes. Those null classes have to be identified. Otherwise, the \texttt{Decaf} accuracy will be substantially degraded or messed up (detailed ablation in \autoref{sec:without_removal}). The \textbf{C2} is because samples per class are often imbalanced to FL users. Moreover, the imbalance can be severe.

\noindent\textit{Our Key Observation and Solution}: DL models memorize the data they have been trained on during training and react to it as gradient changes \cite{1DBLP:conf/ndss/ZhouGFCD0X023,31DBLP:conf/sp/NasrSH19}. Based on this recognition, our first new observation is that different neurons, especially in the last linear layer of the model, are dominated by different classes' sampling data during the FL training. The second observation is that the gradient changes magnitude is dependent on the proportion of each non-null class.

These two observations or insights enable us to constructively address \textbf{C1} and \textbf{C2}, respectively. More specifically, upon the first observation, there will be no positive gradient changes given a null class in its corresponding neuron positions of the last fully connected layer if an FL user has no such class (formally proved in \autoref{sec:null_classes} and experimentally shown in \autoref{sec:target_layer}). We can thus firmly (100\%) determine the null classes to address \textbf{C1}.

Upon the second observation, we can establish a quantitative relationship between the gradient changes and the proportion of each of the remaining non-null classes of any FL user's local dataset. The proportion of non-classes is inferred through an optimization formulation after gradient changes per class are extracted from the victim user's local model with a tiny auxiliary per-class sample (e.g., only 2 samples per class). Identifying null classes and profiling proportions of non-null classes fulfill the \texttt{Decaf} attack.

\noindent\textbf{Contributions:} Our main contributions are threefold:
\begin{itemize}
\item We propose an innovative privacy inference attack based on model gradient changes, namely \texttt{Decaf}, which decomposes any FL user's local data distribution.
\item We devise \texttt{Decaf} as a completely \textit{passive} privacy inference attack that induces neither global model accuracy drop nor FL training latency, which is stealthy and cannot be spotted by victim FL users. We fulfill \texttt{Decaf} in four steps: Gradient Change Extraction, Null Classes Removal, Gradient Bases Construction, and Remaining Classes Decomposition to circumvent \textbf{C1} and \textbf{C2}.
\item We extensively evaluate \texttt{Decaf} performance on five benchmark datasets (MNIST, CIFAR-10, FASHION-MNIST, FER-2013, and SkinCancer datasets) with various model architectures, including custom models, ResNet, and VGG. The experimental results and additional ablation studies affirm that \texttt{Decaf} can consistently achieve high accuracy regardless of whether user data is distributed IID or non-IID.
\end{itemize}

\section{Background and Related Work}
This section first provides background on FL and then concisely describes related privacy inference attacks on it.
\subsection{Federated Learning}
FL~\cite{2DBLP:conf/aistats/McMahanMRHA17} shifts the machine learning paradigm from centralized DL training to distributed DL training manner without aggregating and directly accessing user's private data into a centralized server. According to the data partition, FL can be divided into horizontal federation, vertical federation, and migration learning \cite{3DBLP:journals/tist/YangLCT19, 27DBLP:journals/tsp/PillutlaKH22}. In FL training, participated users download the initial global model $\theta_{global}$ provided by the FL server/coordinator and then train models locally upon their private data point $(x,y)$ to update local models for the current FL round:
\begin{equation}
    \theta_{\rm local}^i=\theta_{\rm global}-\alpha\nabla\ell(x,y),
\end{equation}
where $\alpha$ is the local model training learning rate, and $\ell$ is the loss function. Upon updated local models, for the typical \texttt{FedAvg} FL aggregation \cite{2DBLP:conf/aistats/McMahanMRHA17} used in this work, the server updates in an weighted manner expressed as:
\begin{equation}
    \theta_{\rm global}=\sum_{i}^{n} \frac{D_i}{D} \theta_{\rm local}^i,
\end{equation}
where n users participate in the FL, each possessing a local dataset $D_i$, and $\sum_{i}^{n}D_{i} = D$. Users can then download the updated global model for the next FL training round. 
Opposed to centralized machine learning, FL users do not need to upload their private data to the server, greatly reducing their risk of data privacy leakage. 
Note that the FL server allows all users to synchronize the local model update to the server---not all users must stay online all the time. That is, the FL server can still update the global model even if some users are disconnected due to communication or computation issues or drop due to unavailability. Thus, it is suitable for parallel training scenarios with multiple sources of heterogeneous data, such as smart traffic \cite{28DBLP:journals/iotj/LiuYKNZ20}, smart medical \cite{29DBLP:journals/iotj/ZhangZLWZSWLW21}, and malicious code detection\cite{309772293}.

\subsection{Mainstream Privacy Attacks on Federated Learning}
Despite the fact that FL substantially reduces the privacy leakage risks confronted by centralized DL that directly exposes raw data, it has been shown that some privacy information can be leaked due to privacy inference attacks on FL, such as GAN attack, membership inference attack, property inference attack, and model inversion attack.

\textbf{Generative Adversarial Networks Attack.} Hitaj \textit{et al.} \cite{4DBLP:conf/ccs/HitajAP17} devise a privacy attack method based on Generative Adversarial Networks. The attacker first trains a generator network locally during FL training, which is used to generate training samples of the target class, and obtains the user privacy data contained in the DL model by gaming the generator model with the discriminator model. Wang \textit{et al.} \cite{56DBLP:conf/infocom/WangSZSWQ19} propose a framework incorporating GAN with a multi-task discriminator, which simultaneously discriminates class, reality, and client identity of input samples. 

\textbf{Membership Inference Attack.} Membership inference attack \cite{9DBLP:conf/sp/ShokriSSS17, 10DBLP:conf/csfw/YeomGFJ18, 11DBLP:conf/ndss/Salem0HBF019, 12DBLP:conf/uss/SongM21} is an inferential attack on the degree of affiliation of data samples. The attacker uses the target model output's posterior probability as the attack model's input to determine whether the sample is a member of the training set of the target model. Recently, Choo  \textit{et al.} \cite{13DBLP:conf/icml/Choquette-ChooT21} and Li  \textit{et al.} \cite{14DBLP:conf/ccs/LiZ21} concurrently identify that even only the predicted labeling result can also leak membership information.

\textbf{Property Inference Attack.} Property inference attack \cite{17DBLP:conf/ccs/GanjuWYGB18, 18DBLP:conf/sp/MelisSCS19, 20DBLP:conf/ccs/WangW22} can infer data properties of other members that are not the model task without affecting the training process. Recently, Zhou \textit{et al.} \cite{19DBLP:conf/ndss/Zhou00022} present the first property inference attack against GANs. The attack aims to infer macro-level training data properties, i.e., the proportion of samples used to train the target GAN for a given attribute. Note that though their attack infers similar data proportion properties with our attack, \texttt{Decaf} can get the more comprehensive distribution information. Moreover, \texttt{Decaf} can work for each user in FL, not limited to GANs.

\textbf{Model Inversion Attack.} Model inversion attack \cite{21DBLP:conf/ccs/FredriksonJR15, 22DBLP:conf/uss/TramerZJRR16, 23DBLP:journals/tifs/KhosravyNHNB22} aims to obtain information about models' training data or test data through its prediction output. The attacker can use the application program interface to send large amounts of prediction data to models and then re-model the class labels and confidence coefficients returned by target models. Recently, both Zhu \textit{et al.} \cite{24DBLP:journals/tifs/ZhuYZLZ23} and Kahla \textit{et al.} \cite{25DBLP:conf/cvpr/KahlaCJJ22} find that model inversion attack can be performed using only the data labels rather than the model output's confidence scores, which is more applicable to most current application scenarios.

\textbf{Preference Profiling Attack.} In addition, we note that a recent Preference Profiling Attack (PPA) \cite{1DBLP:conf/ndss/ZhouGFCD0X023} to infer the class with the typically highest proportion in a victim local user's dataset has been revealed. However, PPA cannot decompose the more interesting local data distribution as the \texttt{Decaf} does. Moreover, \texttt{Decaf} has a distinct attacking methodology compared to PPA. PPA exploits model sensitivity to infer FL users' data preference, which is the top-$k$ preference, especially effective for $k=1$. The model sensitivity is the sum of gradient changes, which differs from gradient changes distribution in \texttt{Decaf}. They also actively operate a selective and grouped local model aggregation to enhance their attack performance, which makes their attack less stealthy.

\section{Data Distribution Decompose Attack}
\label{sec:figs}

In this section, we first define the threat model of \texttt{Decaf} and clarify the attack targets. Then we elaborate on the details per \texttt{Decaf} step.
\begin{figure}[h]
  \centering
  \includegraphics[width=\linewidth]{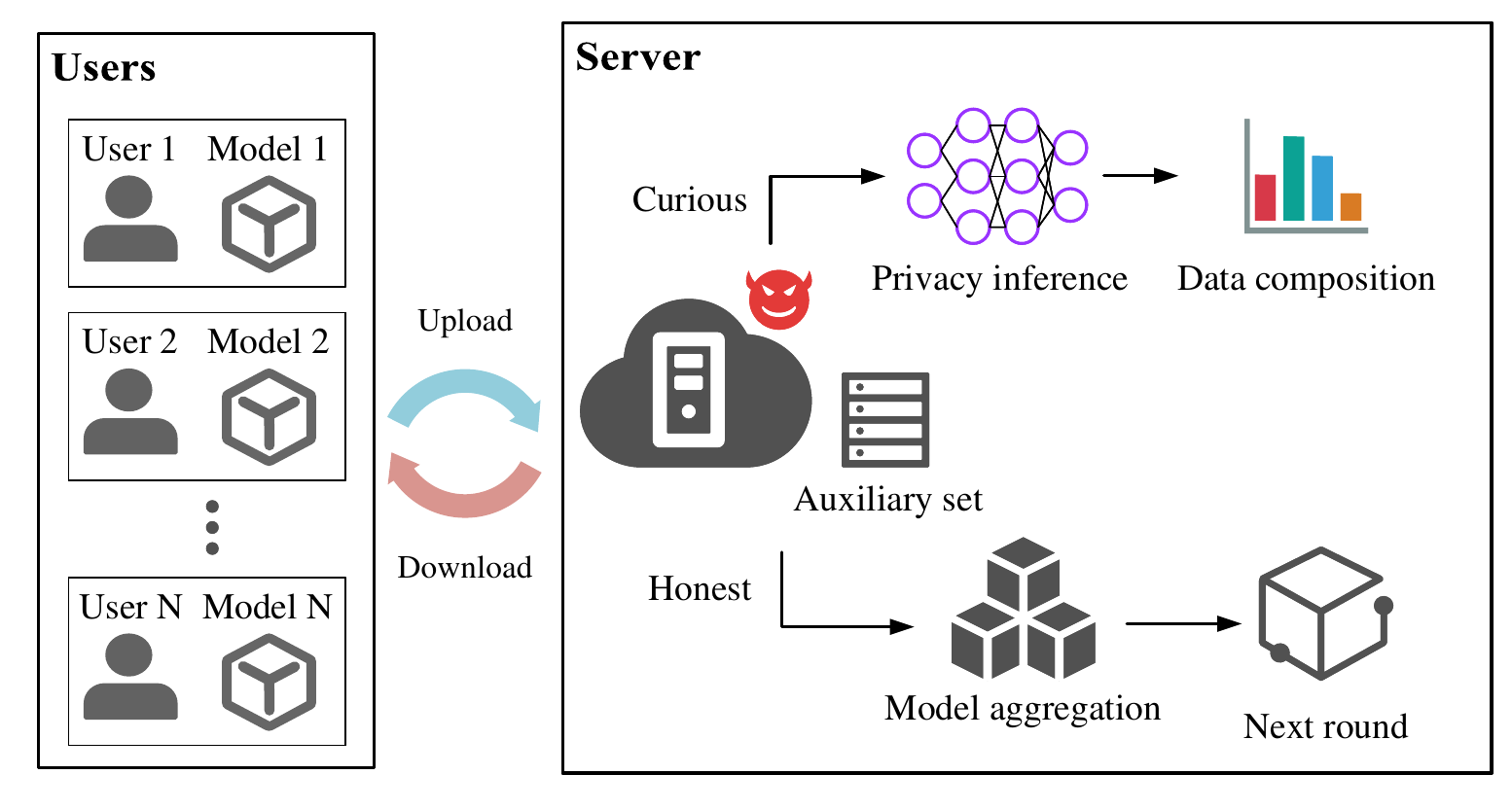}
  \caption{Threat model overview of \texttt{Decaf}.}
  \label{Fig_threat_model}
\end{figure}
\subsection{Threat Model and Attack Targets}
\textbf{Threat Model}. As shown in \autoref{Fig_threat_model}, FL has two types of entities: the users and the server. 
Users are not necessarily limited to individuals but are organizations with private data, such as shopping malls, enterprises, or hospitals. 
The server is assumed to be honest but curious, which is aligned with other mainstream FL privacy information attacks \cite{1DBLP:conf/ndss/ZhouGFCD0X023, 18DBLP:conf/sp/MelisSCS19, 57DBLP:conf/ccs/MalekzadehBG21}. It honestly performs FL global model aggregation while curiously stealing private data composition information from any victim user's local models. 
\texttt{Decaf} targets the plaintext FL aggregation, which is also consistent with existing work on privacy attacks \cite{1DBLP:conf/ndss/ZhouGFCD0X023, 4DBLP:conf/ccs/HitajAP17, 9DBLP:conf/sp/ShokriSSS17, 17DBLP:conf/ccs/GanjuWYGB18, 18DBLP:conf/sp/MelisSCS19, 31DBLP:conf/sp/NasrSH19}.
\begin{figure*}[h]
  \centering
  \includegraphics[width=\linewidth]{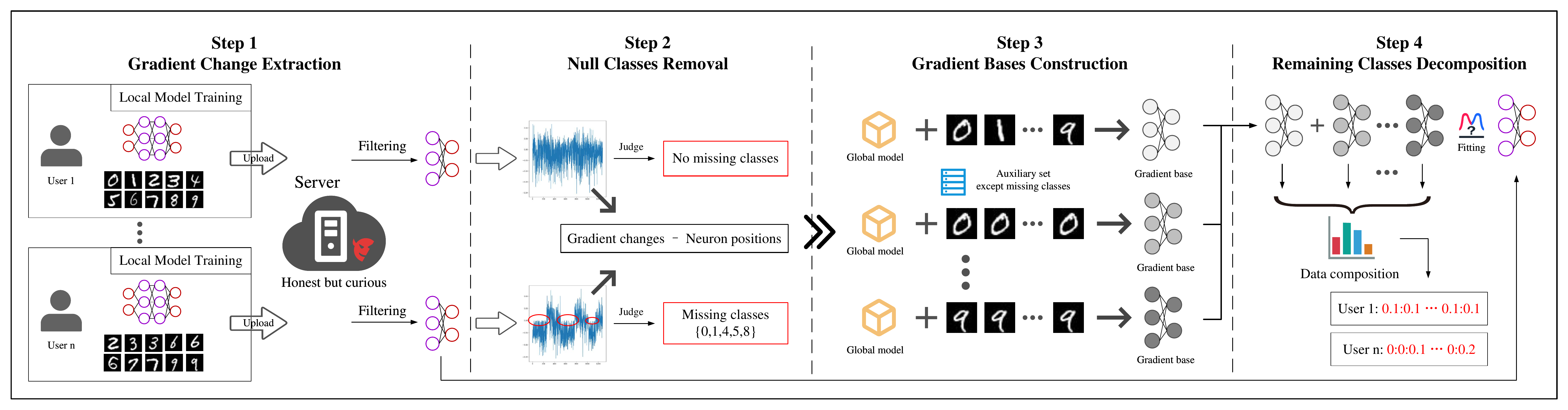}
  \caption{\texttt{Decaf} attack overview. 
  }
  \label{Fig_Decaf}
\end{figure*}
\begin{itemize}
    \item Victim (User): A victim is an FL user whose private data distribution is interested in the \texttt{Decaf} attacker.
    The users are considered honest. They will strictly follow the FL training process to train models locally and upload model parameters to the server for aggregation to harness FL's learning capability on rich, isolated data. Due to the data heterogeneity, users are expected to have different local data distributions.
    \item Attacker (Server): The attacker is the FL server, who is honest but curious. It has full access to local model parameters before and after local model training. The server regularly aggregates the collected user local models and disseminates the new global model for the next training round. However, it might curiously infer users' data composition from the local model parameters. We assume that the server has a tiny auxiliary set (e.g., only 20 images out of 60,000 training images in total for FASHION-MNIST as validated in \autoref{sec:auxiliary}) with samples from all classes, which is reasonably not intersected by users' datasets and under the capability of the server to collect it by itself. For example, the server can collect it from public sources or buy it from the data market. This is completely affordable by the server, as this auxiliary set is very small. Alternatively, the server can ask FL users first to upload a small amount of data to train the initial model. This assumption is essentially aligned with existing works~\cite{1DBLP:conf/ndss/ZhouGFCD0X023, 31DBLP:conf/sp/NasrSH19, 32DBLP:journals/corr/abs-1806-00582, 33DBLP:conf/ndss/CaoF0G21, 34DBLP:conf/esorics/DongCLWZ21, 5910179336, 6010.1145/3576915.3623212, 61287324, 62}. We also note that using a small auxiliary dataset is a practice to facilitate overcoming the non-IID data distribution induced FL accuracy degradation \cite{wang2021addressing,sattler2021fedaux}. In addition, the FL server often needs a small fraction of testing samples to evaluate the global model performance, e.g., to stop. 
\end{itemize}
\noindent\textbf{Attack Targets}. The goal of \texttt{Decaf} is to infer users' data composition from the local model gradients uploaded by users in each FL round. To evaluate the effectiveness of \texttt{Decaf}, we set the following three attack targets:
\begin{itemize}
    \item \textbf{Accuracy}. The data distribution decomposition should reach a high accuracy, measured by a distance metric. We leverage $L_p$ distance, e.g., $L_1$, $L_2$ and $L_\infty$ to quantify the distance between the inferred data distribution composition and the ground-truth data distribution\cite{14DBLP:conf/ccs/LiZ21}. Following , the formal definition of $L_p$ is as below.
    
    Let the feature space $X$ take values $n$-dimensional real vectors, $x_i, x_j \in X=R^n, x_i=(x_i^{(1)}, x_i^{(2)}, ..., x_i^{(n)})^T, x_j=(x_j^{(1)}, x_j^{(2)}, ..., x_j^{(n)})^T$,
    \begin{equation}
    \label{Lp}
        L_p(x_i, x_j) = (\sum_{m=1}^{n}|x_i^{(m)} - x_j^{(m)}|^p)^{\frac{1}{p}}
    \end{equation}
    Specifically, when $p=1, 2, \infty$, it is the Manhattan distance, the European distance, and the Chebyshev distance, respectively. According to the \autoref{Lp}, $L_1$ quantifies the accumulated errors of all classes between predicted proportion and ground truth per class. $L_2$ demonstrates a general error of the data composition. $L_\infty$ measures the largest error of a class out of all classes.
    \item \textbf{Stealthiness}. To ensure that the global model's availability and utility are not affected by \texttt{Decaf}, we will explore the attack stealthiness in terms of FL model accuracy and running time (e.g., no latency is induced per FL round because of \texttt{Decaf}). In addition, the attack should be passive rather than active for staying stealthy. In other words, the FL learning has no difference to the user before and after the \texttt{Decaf}. The victim user is infeasible in detecting malicious behavior from the FL server when their data distribution composition is inferred.
    \item \textbf{Stability}. \texttt{Decaf} should be applicable in diverse FL scenarios, especially fitting the data heterogeneity nature of FL. We set FL users with different data distributions to evaluate \texttt{Decaf}. The \texttt{Decaf} should perform stably even though the attack is only upon any single-round FL learning snapshot, even though that attack upon multiple rounds is better.
\end{itemize}
\begin{table}[h]
\centering
\caption{Notation Summary.}
\resizebox{0.9\linewidth}{!}{
\begin{tabular}{|c|c|}
\hline
\textbf{Notation}            & \textbf{Description}                                                                                                 \\ \hline
$num$                        & \begin{tabular}[c]{@{}c@{}}Average number of neurons of a class\\ in the classification layer\end{tabular}           \\ \hline
$N$                          & Number of data classes                                                                                               \\ \hline
$len()$                      & Length calculation function                                                                                          \\ \hline
$g_{\rm{target}}^t$          & Target gradient changes in round $t$                                                                                 \\ \hline
$flag$                       & Whether the class is missing                                                                                         \\ \hline
$neu$                        & Neuron number                                                                                                        \\ \hline
$C_{\rm{miss}}$              & Set of missing classes                                                                                               \\ \hline
$\theta_{\rm{global}}^{t-1}$ & Parameters of global model in round $t-1$                                                                            \\ \hline
$(x_c, y_c)$                 & A dataset with data of class $c$                                                                                     \\ \hline
$f$                          & Model training function                                                                                              \\ \hline
$\ell$                       & Loss function                                                                                                        \\ \hline
$\nabla$                     & Gradient calculation                                                                                                 \\ \hline
$g_c^t$                      & \begin{tabular}[c]{@{}c@{}}Gradient base constructed by a dataset\\ with data of class $c$ in round $t$\end{tabular} \\ \hline
$g_u^t$                      & \begin{tabular}[c]{@{}c@{}}Gradient base constructed by\\ an union dataset in round $t$\end{tabular}                 \\ \hline
$S_{\rm{bases}}$             & Set of gradient bases                                                                                                \\ \hline
$r_c$                        & The ratio of class $c$                                                                                               \\ \hline
\end{tabular}
}
\label{tab:notation}
\end{table}

\subsection{\texttt{Decaf} Overview}
\texttt{Decaf} is based on a key observation: Although FL model aggregation causes local models to have certain generalization properties and dilutes the privacy of each user, the original gradient changes of local models can still leak their private data information. Thus, \texttt{Decaf} focuses more on the gradient changes of users’ local models, which are entirely influenced by the users’ private data. \texttt{Decaf} is generally divided into four steps, as shown in \autoref{Fig_Decaf}, including gradient change extraction, null classes removal, gradient bases construction, and remaining classes decomposition. 
Notations used for the following descriptions are summarized in \autoref{tab:notation}.
\begin{enumerate}
    \item \textbf{Gradient Change Extraction.} 
    After receiving the local model from a victim user, the server extracts the gradient change of the local model. Since the number of parameters, in particular, can be very large, gradient extraction per neuron is likely to be costly. It is unnecessary, as gradient change extraction from a single layer (e.g., the last fully connected layer) is sufficient for \texttt{Decaf} to decompose the victim's data distribution accurately.
    \item \textbf{Null Classes Removal.} The victim user's local data often does not have all the classes that the global data has. It is crucial to remove those null classes (in particular, those classes have an exact 0\% proportion) from the local data distribution as a prior step before decomposing existing classes accurately.
    \item \textbf{Gradient Bases Construction.} After inferring users' null classes, the server can retrain the global model to construct gradient bases with an auxiliary set. The gradient bases simulate the gradient changes of the victim's local model, which is the foundation for the following step.
    \item \textbf{Remaining Classes Decomposition.} The remaining class decomposition is formulated as an optimization, specifically linear regression, by setting an objective upon the constructed gradient bases. The optimization is guided by a so-called loss computed upon the objective and the extracted gradient changes of the victim's local model. The per-class proportion of remaining classes is obtained by resolving the optimization.
\end{enumerate}
\subsection{Gradient Change Extraction}
The gradient change extraction of victim user $i$ is expressed as follows:
\begin{equation}
    G_i^t = (\theta_i^t - \theta_{\rm{global}}^{t-1}) / \alpha
\end{equation}
where $\theta_i^t$ is the weight vector of $i_{\rm th}$ user's local model at round $t$, $\theta_{\rm{global}}^{t-1}$ is the weight vector of the global model at round $t-1$, and $\alpha$ is the learning rate. Despite that, it is feasible to extract the gradient change for all weights in a DL model, and we found that it is unnecessary to do so as gradient change from one layer is sufficient for \texttt{Decal} performance. This work specifically exploits the last fully connected classification layer as the gradient change extraction layer for the attack expedition (See \autoref{sec:target_layer} for more details). We denote the extracted gradient changes used for later steps of \texttt{Decaf} as $g_{\rm{target}}^t$, which explicitly memorizes the private local user's data distribution information.
\begin{algorithm}
    \caption{Null Classes Removal}
    \label{algo:1}      
    \begin{algorithmic}[1]
        \State \textbf{Input: The user's gradient changes of target layer $g_{\rm target}^t$, the number of classes in the dataset $N$}
        \State \textbf{Output: The set of missing classes for the user $C_{miss}$}
        \State $num\leftarrow len(g_{\rm{target}}^t)/N$
        \For{each $i\in\{1,2,...,N\}$}
        \State $flag\leftarrow 0$
        \For{each $neu\in\{1,2,...,num\}$}
        \If{$g_{\rm{target}}^t[(i-1)*num+neu]>0$}
        \State $flag\leftarrow 1$
        \State \textbf{break}
        \EndIf
        \EndFor
        \If{$flag=0$}
        \State The user misses class $i$, put $i$ into the set $C_{\rm{miss}}$
        \EndIf
        \EndFor
        \State \textbf{return} $C_{\rm{miss}}$
    \end{algorithmic}
\end{algorithm}
\subsection{Null Classes Removal}
\label{sec:null_classes}
The victim user often does not have the total number of global classes. We denote those missed global classes of a victim user as null classes. First, It is essential to identify and remove null classes to achieve accurate data decomposition for non-null classes. It is challenging, which is circumvented through our two newly revealed key observations affirmed with formal proof (experimental validations are in \autoref{sec:target_layer}). Firstly, we reveal that a specific group of neurons in the last fully connected layer of a classification model is specifically responsible for a class. Secondly, given that a class is null in the victim's local data, we found that the gradient changes will not exhibit positive values, which is provable, as we will formally do shortly.
Upon these two key observations, we can precisely identify the null classes of a victim user. For example, given that the last fully connected layer is with $len(g^t_{\rm target})=120$ neurons given a $N=10$ classes, then the number of neurons corresponding to each class is $\frac{len(g^t_{\rm target})}{N} = 12$. The first 12 neurons are for class $0$, the second 12 neurons are for class $1$, etc.
Algorithm \ref{algo:1} elaborates the details of null class removal once the group of neurons per class is determined, \texttt{Line 6-11} iterates each neuron corresponding to a specific class. This class is regarded as a null class conditioned on the fact that none of its corresponding neurons exhibit a positive gradient change. Otherwise, the \texttt{flag} will be set to 1, and this class is regarded as a non-null class.

The following proves the null class relationship with the non-positive gradient values of the last fully connected layer.

\noindent{\bf Proof.} Assume the weight of the last fully connected layer---the layer before softmax---is $m \times n$, where $m$ is the number of inputs and $n$ is the number of outputs (i.e., the number of classes). We define the input $X=[x_1, x_2, ..., x_m]^T$, and the ground-truth label of the sample is $Y=[y_1, y_2, ..., y_n]^T$ with one-hot encoding.

The weight $W$ is:
\begin{equation}
    \begin{bmatrix} 
    w_{11} & w_{12} & \cdots & w_{1m}\\
    w_{21} & w_{22} & \cdots & w_{2m}\\
    \vdots & \vdots &\ddots & \vdots\\
    w_{n1} & w_{n2} & \cdots & w_{nm}
    \end{bmatrix},
\end{equation}
and the bias is $B=[b_1, b_2, ..., b_n]$.

The prediction can then be expressed as:
\begin{equation}
Y^{'}=[y_1^{'}, y_2^{'}, ..., y_n^{'}]=WX+B,
\end{equation}
where
\begin{equation}
y_i^{'}=\sum_{j=1}^{m}w_{ij}x_j + b_i, (1 \leq i \leq n, i \in Z)
\end{equation} 

Then we can normalize the $Y^{'}$:
\begin{equation}
    Y^{''}=[y_1^{''}, y_2^{''}, ..., y_n^{''}]^T
\end{equation}
with a softmax function:
\begin{equation}
    y_i^{''}=\frac{e^{y_i^{'}}}{\sum_{k=1}^{n}e^{y_k^{'}}}. (1 \leq i \leq n, i \in Z).
\end{equation}

Now the cross-entropy loss function is expressed as
\begin{equation}
    L=-\sum_{i=1}^{n}y_iln(y_i^{''}).
\end{equation}

Let $c$ be the label of the sample, where $1 \leq c \leq n$, we can get
\begin{equation}
    y_c=1, y_i=0, i \ne c.
\end{equation} 

The loss function can be then simplified as:
\begin{equation}
    L=-ln(y_c{''}).
\end{equation}

To this end, we can calculate the gradient of $w_{ij}$:
\begin{equation}
    \frac{\partial L}{\partial w_{ij}}=\frac{\partial L}{\partial y_c^{''}} \cdot \frac{\partial y_c^{''}}{\partial y_i^{'}} \cdot \frac{\partial y_i^{'}}{\partial w_{ij}}=-\frac{1}{y_c^{''}} \cdot \frac{\partial y_c^{''}}{\partial y_i^{'}} \cdot x_j.
\end{equation}

If $i \ne c$, we have
\begin{equation}
    \frac{\partial y_c^{''}}{\partial y_i^{'}}=-\frac{e^{y_i^{'}}}{(\sum_{k=1}^{n}e^{y_k^{'}})^{2}} < 0.
\end{equation}

Consider that the input $X$ is not only the input of the current layer but also the output of the previous layer. And it must be $\ge 0$ after applying the activation function of \textsf{relu} (value range is $[0, \infty)$), or \textsf{sigmod} (value range is (0, 1)). Since $-\frac{1}{y_c^{''}} < 0$, the gradient is $\frac{\partial L}{\partial w_{ij}} \geq 0$. Thus, we can get
\begin{equation}
    w_{ij}^{t} = w_{ij}^{t-1} - \alpha \frac{\partial L}{\partial w_{ij}}
\end{equation}
\begin{equation}
\label{eq1}
    g^{t}= \frac{w_{ij}^{t}-w_{ij}^{t-1}}{\alpha}= -\frac{\partial L}{\partial w_{ij}} \leq 0,
\end{equation}
where $g^{t}$ is the gradient of epoch $t$ that can be calculated by the attacker server, and $\alpha$ is the learning rate. As a result, $w_{ij}$ always exhibits either 0 or negative gradient changes, if $i \ne c$, where $1 \leq i \leq n$, $1 \leq j \leq m$, $i, j \in Z$. 

Otherwise, if $i = c$, we have
\begin{equation}
    \frac{\partial y_c^{''}}{\partial y_i^{'}}=\frac{\partial y_c^{''}}{\partial y_c^{'}}=\frac{e^{y_c^{'}}}{(\sum_{k=1}^{n}e^{y_k^{'}})^{2}} > 0.
\end{equation}

Thus, we can get
\begin{equation}
\label{eq2}
    g^{t}= \frac{w_{cj}^{t}-w_{cj}^{t-1}}{\alpha}= -\frac{\partial L}{\partial w_{cj}} \geq 0.
\end{equation}

As a result, $w_{ij}$ always show positive gradient changes, if $i = c$, where $1 \leq j \leq m$, $i, j \in Z$. 

To this end, we have proved that the parameters $w_{cj}$ are indeed responsible for the classification of class $c$, which is consistent with our first observation in \autoref{sec:null_classes}. Furthermore, the second observation has been proved by the \autoref{eq1} and \autoref{eq2}. If there are no class $c$ samples in the user's dataset, then its corresponding $w_{cj}$ must not have positive gradient changes, either 0 or negative.

\begin{algorithm}
    \caption{Gradient Bases Construction}
    \label{algo:2}      
    \begin{algorithmic}[1]
        \State \textbf{Input: The global model in round {t-1} $\theta_{\rm{global}}^{t-1}$, learning rate $\alpha$, auxiliary set $(x_c, y_c)$, where $c \in \{1,2,...,N\} \setminus C_{\rm{miss}}$}
        \State \textbf{Output: The set of gradient bases $S_{\rm{bases}}$}
        \For{each $c$}
        \State $\theta_c^t \leftarrow \theta_{\rm{global}}^{t-1} - \alpha \nabla \ell (f(x_c,\theta_{\rm{global}}^{t-1}),y_c)$
        \State Calculate the same gradient changes of target layer $g_c^t$ and put it into the set $S_{\rm{bases}}$
        \EndFor
        \State $(x_u, y_u) \leftarrow \sum_c (x_c, y_c)$, where $c \in \{1,2,...,N\} \setminus C_{\rm{miss}}$
        \State $\theta_u^t \leftarrow \theta_{\rm{global}}^{t-1} - \alpha \nabla \ell (f(x_u,\theta_{\rm{global}}^{t-1}),y_u)$
        \State Calculate the same gradient changes of target layer $g_u^t$ and put it into the set $S_{\rm{bases}}$
        \State \textbf{return} $S_{\rm{bases}}$
    \end{algorithmic}
\end{algorithm}

\subsection{Gradient Bases Construction} 
Gradient bases are required to facilitate the following non-null class data distribution decomposition. The gradient bases are generally a set of gradient references extracted from a simulated local model update \texttt{per non-null class} given that all classes are evenly distributed---the number of each class is equal. As shown in Algorithm \ref{algo:2}, to do so, the server utilizes auxiliary set $(x_c, y_c)$ to train the global model at the $t-1$ round, here $(x_c, y_c)$ is the dataset only consisting of samples from $c_{\rm th}$ class and $c\in \{1,2,...,N\} \setminus C_{\rm{null}}$---any class within \texttt{non-null} classes. Then the server extracts the gradient change of the target layer (e.g., the last fully connected layer) for class $c$, serving as the gradient base of class $c$, which process is repeated for all non-null classes, respectively (\texttt{Line 3-6}).
In addition, as in \texttt{Line 8-10}, the server utilizes the unified auxiliary dataset $(x_u, y_u)$ to simulate a global model to obtain a unified gradient base $g_u^t$, where $(x_u, y_u)$ is the union of the dataset $(x_c, y_c)$. 
The gradient base $g_u^t$ is used to simulate the general trend of the gradient changes of the victim model $g_{\rm{target}}^t$. In contrast, the gradient bases $g_c^t$ are specifically used to distinguish the proportion between them (e.g., by identifying the corresponding scaling factors).

\begin{algorithm}
    \caption{Remaining Classes Decomposition}
    \label{algo:3}      
    \begin{algorithmic}[1]
        \State \textbf{Input: The target gradient changes $g_{\rm{target}}^t$, gradient bases $S_{\rm{bases}}$, learning rate $\beta$, number of rounds $epoch$ }
        \State \textbf{Output: The data composition $R$, where $r_c \in R$ and $c \in \{1,2,...,N\} \setminus C_{\rm{miss}}$}
        \State Initialize scalar factors $\eta_c$ and $\eta_u$, where $\eta_c, \eta_u \ge 0$ and $c \in \{1,2,...,N\} \setminus C_{\rm{miss}}$
        \State Define the objective function $h = \sum_c \eta_c g_c^t + \eta_u g_u^t$, where $g_c^t,g_u^t \in S_{\rm{bases}}$
        \For{each $e \in \{1,2,...,epoch\}$}
        \State Define the loss function $\ell = (h - g_{\rm{target}}^t)^2$
        \State $\eta_c \leftarrow \eta_c - \beta \nabla \ell$
        \State $\eta_u \leftarrow \eta_u - \beta \nabla \ell$
        \EndFor
        \For{each $c$}
        \State $r_c = (\eta_c + \eta_u) / \sum_c(\eta_c + \eta_u)$
        \State Put $r_c$ into $R$
        \EndFor
        \State \textbf{return} $R$
    \end{algorithmic}
\end{algorithm}

\subsection{Remaining Classes Decomposition}
\label{RCD}
The complete process of remaining classes decomposition is detailed in Algorithm \ref{algo:3}. We formulate the remaining classes decomposition as an optimization problem. The objective function is expressed as:
\begin{equation}
    h = \sum_c \eta_c g_c^t + \eta_u g_u^t,
\end{equation}
where $g_c^t$ and $g_u^t$ are per-class gradient bases and unified gradient base extracted with auxiliary data in the previous step. $\eta_c, \eta_u \ge 0$ and $c \in \{1,2,...,N\} \setminus C_{\rm{miss}}$, and $\eta_c, \eta_u \ge 0$ are scalar factors that are going to be solved. Once identified, these scalar factors can be used to compute the data distribution decomposition. The loss is based on the mean square error between the $h$ and $g_{\rm{target}}^t$ that is from the victim user, expressed as:
\begin{equation}
    \ell = (h - g_{\rm{target}}^t)^2.
\end{equation}

Upon the formulated objective function and loss guidance, scalar factors of $\eta_c, \eta_u \ge 0$ are resolved via gradient descent after a number of epochs (e.g., 1000).
Consequentially, the proportion per remaining class, $r_c$, is obtained through:
\begin{equation}
    r_c = (\eta_c + \eta_u) / \sum_c(\eta_c + \eta_u).
\end{equation}

Note that $\eta_u$ corresponding to $g_u^t$ is alike a calibrator to eliminate large variance when the $\eta_c, c \in \{1,2,...,N\} \setminus C_{\rm{miss}}$ are solely computed through $g_c^t$. The significance of this collaborator is validated in \autoref{sec:calibrator}.

\begin{table}[htbp]
\centering
\caption{Model Structure.}
\resizebox{0.9\linewidth}{!}{
\begin{tabular}{|c|c|}
\hline
\textbf{Datasets} & \textbf{Model Structure}               \\ \hline
MNIST             & 2 conv + 1 pool + 2 fn                 \\ \hline
FASHION-MNIST     & 2 conv + 1 pool + 2 fn                 \\ \hline
CIFAR-10          & 3 conv modul. + 2 fn (ResNet18, VGG16) \\ \hline
FER-2013          & 4 conv modul. + 2 fn (ResNet18, VGG16) \\ \hline
SkinCancer        & 3 conv modul. + 2 fn                   \\ \hline
\end{tabular}
}
\label{tab:model_structure}
\end{table}

\section{Experiments}

This section extensively evaluates \texttt{Decaf} and analyzes its attacking performance in terms of attack accuracy, stealthiness, and generalization. The rationale for utilizing the last fully connected layer as the target layer for gradient change extraction is then explained. In addition, \texttt{Decaf} is evaluated through complex models (e.g., ResNet18, VGG16). Lastly, we experiment and discuss the key insights of \texttt{Decaf}, including defenses and real scenario applications.
\subsection{Experimental Settings}
\noindent\textbf{Datasets.}
We evaluate the \texttt{Decaf} on five benchmark Datasets: MNIST, FASHION-MNIST, CIFAR-10, FER-2013, and SkinCancer. Among them, note that FASHION-MNIST can resemble commodity recommendation scenarios; FER-2013 resembles facial expression recognition application, which both data distribution are sensitive to local users in practice. Medical dataset of SkinCancer also contains private information.

\begin{itemize}
\item MNIST~\cite{35DBLP:journals/pieee/LeCunBBH98} is a formatted 0-9 handwritten digit recognition grayscale image dataset. It includes 60,000 training data images with 10,000 test data images. Each image is with a size of $28 \times 28\times1$.

\item FASHION-MNIST~\cite{36DBLP:journals/corr/abs-1708-07747} consists of ten classes of images that include t-shirts, jeans, pullovers, skirts, jackets, sandals, shirts, sneakers, bags, and boots. In the format of grayscale images with the size of $28\times 28\times 1$, a total of 60,000 images of training data and 10,000 images of test data are included.

\item CIFAR-10~\cite{37krizhevsky2009learning} consists of ten classes. It has 60,000 images with a size of $32\times 32 \times 3$, with 6,000 images in each class. Among them, there are 50,000 training images and 10,000 test images.

\item FER-2013~\cite{38} is a face expression dataset with approximately 30,000 diverse grayscale facial images with the size of 48 $\times 48 \times 1$. The expression dataset has seven classes: anger, disgust, fear, happiness, neutrality, sadness, and surprise. Among them, there are 32,298 training images and 3,754 test images.

\item SkinCancer \cite{39DBLP:journals/corr/abs-1902-03368, 40DBLP:journals/corr/abs-1803-10417} consists of 10,015 dermoscopic images, and cases include a representative collection of all critical diagnostic classes in the field of pigmented lesions. It has seven classes that are actinic keratosis and intraepithelial carcinoma/Bowen's disease (akiec), basal cell carcinoma (bcc), benign keratosis-like lesions (bkl), dermatofibrosarcoma (df), melanoma (mel), melanocytic nevus (nv), and vascular lesions (vasc). Among them, there are 7,010 training images and 3,005 test images. Each image is with a size of $28 \times 28 \times 3$.
\end{itemize}

\begin{table}[h]
\centering
\caption{The results of \texttt{Decaf} on MNIST.}
\resizebox{\linewidth}{!}{
\begin{tabular}{|c|cccccccccc|}
\hline
                                & \multicolumn{10}{c|}{}                                                                                                                                                                                                                                                                                                                                                                                                                                                                                                     \\
\multirow{-2}{*}{\textbf{User}} & \multicolumn{10}{c|}{\multirow{-2}{*}{\textbf{Data Composition (\%)}}}                                                                                                                                                                                                                                                                                                                                                                                                                                                     \\ \hline
                                & \multicolumn{1}{c|}{10}                            & \multicolumn{1}{c|}{10}                            & \multicolumn{1}{c|}{10}                            & \multicolumn{1}{c|}{10}                            & \multicolumn{1}{c|}{10}                            & \multicolumn{1}{c|}{10}                            & \multicolumn{1}{c|}{10}                            & \multicolumn{1}{c|}{10}                            & \multicolumn{1}{c|}{10}                            & 10                            \\ \cline{2-11} 
\multirow{-2}{*}{\textbf{1}}    & \multicolumn{1}{c|}{\cellcolor[HTML]{C0C0C0}9.43}  & \multicolumn{1}{c|}{\cellcolor[HTML]{C0C0C0}10.21} & \multicolumn{1}{c|}{\cellcolor[HTML]{C0C0C0}9.84}  & \multicolumn{1}{c|}{\cellcolor[HTML]{C0C0C0}9.68}  & \multicolumn{1}{c|}{\cellcolor[HTML]{C0C0C0}9.85}  & \multicolumn{1}{c|}{\cellcolor[HTML]{C0C0C0}9.59}  & \multicolumn{1}{c|}{\cellcolor[HTML]{C0C0C0}10.29} & \multicolumn{1}{c|}{\cellcolor[HTML]{C0C0C0}10.48} & \multicolumn{1}{c|}{\cellcolor[HTML]{C0C0C0}11.37} & \cellcolor[HTML]{C0C0C0}9.26  \\ \hline
                                & \multicolumn{1}{c|}{6.67}                          & \multicolumn{1}{c|}{10}                            & \multicolumn{1}{c|}{8.33}                          & \multicolumn{1}{c|}{7.5}                           & \multicolumn{1}{c|}{13.33}                         & \multicolumn{1}{c|}{15.83}                         & \multicolumn{1}{c|}{5.83}                          & \multicolumn{1}{c|}{10}                            & \multicolumn{1}{c|}{12.5}                          & 10                            \\ \cline{2-11} 
\multirow{-2}{*}{\textbf{2}}    & \multicolumn{1}{c|}{\cellcolor[HTML]{C0C0C0}10.65} & \multicolumn{1}{c|}{\cellcolor[HTML]{C0C0C0}9.76}  & \multicolumn{1}{c|}{\cellcolor[HTML]{C0C0C0}9.65}  & \multicolumn{1}{c|}{\cellcolor[HTML]{C0C0C0}9.36}  & \multicolumn{1}{c|}{\cellcolor[HTML]{C0C0C0}10.74} & \multicolumn{1}{c|}{\cellcolor[HTML]{C0C0C0}13.11} & \multicolumn{1}{c|}{\cellcolor[HTML]{C0C0C0}7.77}  & \multicolumn{1}{c|}{\cellcolor[HTML]{C0C0C0}10.83} & \multicolumn{1}{c|}{\cellcolor[HTML]{C0C0C0}10.4}  & \cellcolor[HTML]{C0C0C0}7.75  \\ \hline
                                & \multicolumn{1}{c|}{15}                            & \multicolumn{1}{c|}{15.83}                         & \multicolumn{1}{c|}{7.5}                           & \multicolumn{1}{c|}{5}                             & \multicolumn{1}{c|}{15}                            & \multicolumn{1}{c|}{12.5}                          & \multicolumn{1}{c|}{13.33}                         & \multicolumn{1}{c|}{6.67}                          & \multicolumn{1}{c|}{4.17}                          & 5                             \\ \cline{2-11} 
\multirow{-2}{*}{\textbf{3}}    & \multicolumn{1}{c|}{\cellcolor[HTML]{C0C0C0}12.25} & \multicolumn{1}{c|}{\cellcolor[HTML]{C0C0C0}13.76} & \multicolumn{1}{c|}{\cellcolor[HTML]{C0C0C0}8.62}  & \multicolumn{1}{c|}{\cellcolor[HTML]{C0C0C0}9.74}  & \multicolumn{1}{c|}{\cellcolor[HTML]{C0C0C0}10.78} & \multicolumn{1}{c|}{\cellcolor[HTML]{C0C0C0}9.86}  & \multicolumn{1}{c|}{\cellcolor[HTML]{C0C0C0}13.1}  & \multicolumn{1}{c|}{\cellcolor[HTML]{C0C0C0}10.59} & \multicolumn{1}{c|}{\cellcolor[HTML]{C0C0C0}5.66}  & \cellcolor[HTML]{C0C0C0}5.66  \\ \hline
                                & \multicolumn{1}{c|}{13.33}                         & \multicolumn{1}{c|}{3.33}                          & \multicolumn{1}{c|}{2.5}                           & \multicolumn{1}{c|}{18.33}                         & \multicolumn{1}{c|}{10}                            & \multicolumn{1}{c|}{13.33}                         & \multicolumn{1}{c|}{5.83}                          & \multicolumn{1}{c|}{1.67}                          & \multicolumn{1}{c|}{8.33}                          & 23.33                         \\ \cline{2-11} 
\multirow{-2}{*}{\textbf{4}}    & \multicolumn{1}{c|}{\cellcolor[HTML]{C0C0C0}12.65} & \multicolumn{1}{c|}{\cellcolor[HTML]{C0C0C0}5.91}  & \multicolumn{1}{c|}{\cellcolor[HTML]{C0C0C0}8.73}  & \multicolumn{1}{c|}{\cellcolor[HTML]{C0C0C0}11.95} & \multicolumn{1}{c|}{\cellcolor[HTML]{C0C0C0}9.34}  & \multicolumn{1}{c|}{\cellcolor[HTML]{C0C0C0}12.33} & \multicolumn{1}{c|}{\cellcolor[HTML]{C0C0C0}6.34}  & \multicolumn{1}{c|}{\cellcolor[HTML]{C0C0C0}2.54}  & \multicolumn{1}{c|}{\cellcolor[HTML]{C0C0C0}11.18} & \cellcolor[HTML]{C0C0C0}19.03 \\ \hline
                                & \multicolumn{1}{c|}{5}                             & \multicolumn{1}{c|}{18.33}                         & \multicolumn{1}{c|}{-}                             & \multicolumn{1}{c|}{8.33}                          & \multicolumn{1}{c|}{10}                            & \multicolumn{1}{c|}{12.5}                          & \multicolumn{1}{c|}{25}                            & \multicolumn{1}{c|}{8.33}                          & \multicolumn{1}{c|}{6.67}                          & 5.83                          \\ \cline{2-11} 
\multirow{-2}{*}{\textbf{5}}    & \multicolumn{1}{c|}{\cellcolor[HTML]{C0C0C0}9.93}  & \multicolumn{1}{c|}{\cellcolor[HTML]{C0C0C0}16.23} & \multicolumn{1}{c|}{\cellcolor[HTML]{C0C0C0}-}     & \multicolumn{1}{c|}{\cellcolor[HTML]{C0C0C0}8.52}  & \multicolumn{1}{c|}{\cellcolor[HTML]{C0C0C0}13.56} & \multicolumn{1}{c|}{\cellcolor[HTML]{C0C0C0}10.54} & \multicolumn{1}{c|}{\cellcolor[HTML]{C0C0C0}15.9}  & \multicolumn{1}{c|}{\cellcolor[HTML]{C0C0C0}12.01} & \multicolumn{1}{c|}{\cellcolor[HTML]{C0C0C0}7.77}  & \cellcolor[HTML]{C0C0C0}5.53  \\ \hline
                                & \multicolumn{1}{c|}{6.67}                          & \multicolumn{1}{c|}{10.83}                         & \multicolumn{1}{c|}{15}                            & \multicolumn{1}{c|}{5}                             & \multicolumn{1}{c|}{16.67}                         & \multicolumn{1}{c|}{-}                             & \multicolumn{1}{c|}{12.5}                          & \multicolumn{1}{c|}{8.33}                          & \multicolumn{1}{c|}{25}                            & -                             \\ \cline{2-11} 
\multirow{-2}{*}{\textbf{6}}    & \multicolumn{1}{c|}{\cellcolor[HTML]{C0C0C0}12.6}  & \multicolumn{1}{c|}{\cellcolor[HTML]{C0C0C0}11.67} & \multicolumn{1}{c|}{\cellcolor[HTML]{C0C0C0}12.17} & \multicolumn{1}{c|}{\cellcolor[HTML]{C0C0C0}10.07} & \multicolumn{1}{c|}{\cellcolor[HTML]{C0C0C0}14.36} & \multicolumn{1}{c|}{\cellcolor[HTML]{C0C0C0}-}     & \multicolumn{1}{c|}{\cellcolor[HTML]{C0C0C0}12.63} & \multicolumn{1}{c|}{\cellcolor[HTML]{C0C0C0}8.18}  & \multicolumn{1}{c|}{\cellcolor[HTML]{C0C0C0}18.33} & \cellcolor[HTML]{C0C0C0}-     \\ \hline
                                & \multicolumn{1}{c|}{16.67}                         & \multicolumn{1}{c|}{7.5}                           & \multicolumn{1}{c|}{13.33}                         & \multicolumn{1}{c|}{-}                             & \multicolumn{1}{c|}{7.5}                           & \multicolumn{1}{c|}{25}                            & \multicolumn{1}{c|}{-}                             & \multicolumn{1}{c|}{-}                             & \multicolumn{1}{c|}{26.67}                         & 3.33                          \\ \cline{2-11} 
\multirow{-2}{*}{\textbf{7}}    & \multicolumn{1}{c|}{\cellcolor[HTML]{C0C0C0}12.4}  & \multicolumn{1}{c|}{\cellcolor[HTML]{C0C0C0}12.46} & \multicolumn{1}{c|}{\cellcolor[HTML]{C0C0C0}15.5}  & \multicolumn{1}{c|}{\cellcolor[HTML]{C0C0C0}-}     & \multicolumn{1}{c|}{\cellcolor[HTML]{C0C0C0}11.92} & \multicolumn{1}{c|}{\cellcolor[HTML]{C0C0C0}16.67} & \multicolumn{1}{c|}{\cellcolor[HTML]{C0C0C0}-}     & \multicolumn{1}{c|}{\cellcolor[HTML]{C0C0C0}-}     & \multicolumn{1}{c|}{\cellcolor[HTML]{C0C0C0}21.69} & \cellcolor[HTML]{C0C0C0}9.37  \\ \hline
                                & \multicolumn{1}{c|}{-}                             & \multicolumn{1}{c|}{-}                             & \multicolumn{1}{c|}{33.33}                         & \multicolumn{1}{c|}{5}                             & \multicolumn{1}{c|}{-}                             & \multicolumn{1}{c|}{-}                             & \multicolumn{1}{c|}{26.67}                         & \multicolumn{1}{c|}{3.33}                          & \multicolumn{1}{c|}{-}                             & 31.67                         \\ \cline{2-11} 
\multirow{-2}{*}{\textbf{8}}    & \multicolumn{1}{c|}{\cellcolor[HTML]{C0C0C0}-}     & \multicolumn{1}{c|}{\cellcolor[HTML]{C0C0C0}-}     & \multicolumn{1}{c|}{\cellcolor[HTML]{C0C0C0}28.83} & \multicolumn{1}{c|}{\cellcolor[HTML]{C0C0C0}12.37} & \multicolumn{1}{c|}{\cellcolor[HTML]{C0C0C0}-}     & \multicolumn{1}{c|}{\cellcolor[HTML]{C0C0C0}-}     & \multicolumn{1}{c|}{\cellcolor[HTML]{C0C0C0}26}    & \multicolumn{1}{c|}{\cellcolor[HTML]{C0C0C0}5.13}  & \multicolumn{1}{c|}{\cellcolor[HTML]{C0C0C0}-}     & \cellcolor[HTML]{C0C0C0}27.67 \\ \hline
                                & \multicolumn{1}{c|}{-}                             & \multicolumn{1}{c|}{-}                             & \multicolumn{1}{c|}{-}                             & \multicolumn{1}{c|}{41.67}                         & \multicolumn{1}{c|}{-}                             & \multicolumn{1}{c|}{8.33}                          & \multicolumn{1}{c|}{-}                             & \multicolumn{1}{c|}{-}                             & \multicolumn{1}{c|}{-}                             & 50                            \\ \cline{2-11} 
\multirow{-2}{*}{\textbf{9}}    & \multicolumn{1}{c|}{\cellcolor[HTML]{C0C0C0}-}     & \multicolumn{1}{c|}{\cellcolor[HTML]{C0C0C0}-}     & \multicolumn{1}{c|}{\cellcolor[HTML]{C0C0C0}-}     & \multicolumn{1}{c|}{\cellcolor[HTML]{C0C0C0}40.89} & \multicolumn{1}{c|}{\cellcolor[HTML]{C0C0C0}-}     & \multicolumn{1}{c|}{\cellcolor[HTML]{C0C0C0}11.39} & \multicolumn{1}{c|}{\cellcolor[HTML]{C0C0C0}-}     & \multicolumn{1}{c|}{\cellcolor[HTML]{C0C0C0}-}     & \multicolumn{1}{c|}{\cellcolor[HTML]{C0C0C0}-}     & \cellcolor[HTML]{C0C0C0}47.71 \\ \hline
                                & \multicolumn{1}{c|}{-}                             & \multicolumn{1}{c|}{-}                             & \multicolumn{1}{c|}{-}                             & \multicolumn{1}{c|}{-}                             & \multicolumn{1}{c|}{-}                             & \multicolumn{1}{c|}{-}                             & \multicolumn{1}{c|}{-}                             & \multicolumn{1}{c|}{100}                           & \multicolumn{1}{c|}{-}                             & -                             \\ \cline{2-11} 
\multirow{-2}{*}{\textbf{10}}   & \multicolumn{1}{c|}{\cellcolor[HTML]{C0C0C0}-}     & \multicolumn{1}{c|}{\cellcolor[HTML]{C0C0C0}-}     & \multicolumn{1}{c|}{\cellcolor[HTML]{C0C0C0}-}     & \multicolumn{1}{c|}{\cellcolor[HTML]{C0C0C0}-}     & \multicolumn{1}{c|}{\cellcolor[HTML]{C0C0C0}-}     & \multicolumn{1}{c|}{\cellcolor[HTML]{C0C0C0}-}     & \multicolumn{1}{c|}{\cellcolor[HTML]{C0C0C0}-}     & \multicolumn{1}{c|}{\cellcolor[HTML]{C0C0C0}100}   & \multicolumn{1}{c|}{\cellcolor[HTML]{C0C0C0}-}     & \cellcolor[HTML]{C0C0C0}-     \\ \hline
\end{tabular}
}
\label{tab:mnist_res}
\end{table}

\begin{table}[h]
\centering
\caption{The results of \texttt{Decaf} on FASHION-MNIST.}
\resizebox{\linewidth}{!}{
\begin{tabular}{|c|cccccccccc|}
\hline
                                & \multicolumn{10}{c|}{}                                                                                                                                                                                                                                                                                                                                                                                                                                                                                                                                                                                                                                                                                                                                           \\
\multirow{-2}{*}{\textbf{User}} & \multicolumn{10}{c|}{\multirow{-2}{*}{\textbf{Data Composition (\%)}}}                                                                                                                                                                                                                                                                                                                                                                                                                                                                                                                                                                                                                                                                                           \\ \hline
                                & \multicolumn{1}{c|}{10}                                                   & \multicolumn{1}{c|}{10}                                                   & \multicolumn{1}{c|}{10}                                                   & \multicolumn{1}{c|}{10}                                                   & \multicolumn{1}{c|}{10}                                                   & \multicolumn{1}{c|}{10}                                                   & \multicolumn{1}{c|}{10}                                                   & \multicolumn{1}{c|}{10}                                                   & \multicolumn{1}{c|}{10}                                                   & 10                                                   \\ \cline{2-11} 
\multirow{-2}{*}{\textbf{1}}    & \multicolumn{1}{c|}{\cellcolor[HTML]{C0C0C0}{\color[HTML]{000000} 10.52}} & \multicolumn{1}{c|}{\cellcolor[HTML]{C0C0C0}{\color[HTML]{000000} 9.66}}  & \multicolumn{1}{c|}{\cellcolor[HTML]{C0C0C0}{\color[HTML]{000000} 10.5}}  & \multicolumn{1}{c|}{\cellcolor[HTML]{C0C0C0}{\color[HTML]{000000} 9.74}}  & \multicolumn{1}{c|}{\cellcolor[HTML]{C0C0C0}{\color[HTML]{000000} 10.2}}  & \multicolumn{1}{c|}{\cellcolor[HTML]{C0C0C0}{\color[HTML]{000000} 9.66}}  & \multicolumn{1}{c|}{\cellcolor[HTML]{C0C0C0}{\color[HTML]{000000} 10.36}} & \multicolumn{1}{c|}{\cellcolor[HTML]{C0C0C0}{\color[HTML]{000000} 9.81}}  & \multicolumn{1}{c|}{\cellcolor[HTML]{C0C0C0}{\color[HTML]{000000} 9.91}}  & \cellcolor[HTML]{C0C0C0}{\color[HTML]{000000} 9.66}  \\ \hline
                                & \multicolumn{1}{c|}{6.67}                                                 & \multicolumn{1}{c|}{10}                                                   & \multicolumn{1}{c|}{8.33}                                                 & \multicolumn{1}{c|}{7.5}                                                  & \multicolumn{1}{c|}{13.33}                                                & \multicolumn{1}{c|}{15.83}                                                & \multicolumn{1}{c|}{5.83}                                                 & \multicolumn{1}{c|}{10}                                                   & \multicolumn{1}{c|}{12.5}                                                 & 10                                                   \\ \cline{2-11} 
\multirow{-2}{*}{\textbf{2}}    & \multicolumn{1}{c|}{\cellcolor[HTML]{C0C0C0}{\color[HTML]{000000} 8.26}}  & \multicolumn{1}{c|}{\cellcolor[HTML]{C0C0C0}{\color[HTML]{000000} 10.91}} & \multicolumn{1}{c|}{\cellcolor[HTML]{C0C0C0}{\color[HTML]{000000} 8.66}}  & \multicolumn{1}{c|}{\cellcolor[HTML]{C0C0C0}{\color[HTML]{000000} 11.03}} & \multicolumn{1}{c|}{\cellcolor[HTML]{C0C0C0}{\color[HTML]{000000} 12.7}}  & \multicolumn{1}{c|}{\cellcolor[HTML]{C0C0C0}{\color[HTML]{000000} 10.89}} & \multicolumn{1}{c|}{\cellcolor[HTML]{C0C0C0}{\color[HTML]{000000} 5.99}}  & \multicolumn{1}{c|}{\cellcolor[HTML]{C0C0C0}{\color[HTML]{000000} 10.14}} & \multicolumn{1}{c|}{\cellcolor[HTML]{C0C0C0}{\color[HTML]{000000} 11.55}} & \cellcolor[HTML]{C0C0C0}{\color[HTML]{000000} 9.87}  \\ \hline
                                & \multicolumn{1}{c|}{15}                                                   & \multicolumn{1}{c|}{15.83}                                                & \multicolumn{1}{c|}{7.5}                                                  & \multicolumn{1}{c|}{5}                                                    & \multicolumn{1}{c|}{15}                                                   & \multicolumn{1}{c|}{12.5}                                                 & \multicolumn{1}{c|}{13.33}                                                & \multicolumn{1}{c|}{6.67}                                                 & \multicolumn{1}{c|}{4.17}                                                 & 5                                                    \\ \cline{2-11} 
\multirow{-2}{*}{\textbf{3}}    & \multicolumn{1}{c|}{\cellcolor[HTML]{C0C0C0}{\color[HTML]{000000} 15.74}} & \multicolumn{1}{c|}{\cellcolor[HTML]{C0C0C0}{\color[HTML]{000000} 12.45}} & \multicolumn{1}{c|}{\cellcolor[HTML]{C0C0C0}{\color[HTML]{000000} 8.21}}  & \multicolumn{1}{c|}{\cellcolor[HTML]{C0C0C0}{\color[HTML]{000000} 6.79}}  & \multicolumn{1}{c|}{\cellcolor[HTML]{C0C0C0}{\color[HTML]{000000} 12.35}} & \multicolumn{1}{c|}{\cellcolor[HTML]{C0C0C0}{\color[HTML]{000000} 13.99}} & \multicolumn{1}{c|}{\cellcolor[HTML]{C0C0C0}{\color[HTML]{000000} 9.61}}  & \multicolumn{1}{c|}{\cellcolor[HTML]{C0C0C0}{\color[HTML]{000000} 8.03}}  & \multicolumn{1}{c|}{\cellcolor[HTML]{C0C0C0}{\color[HTML]{000000} 6.42}}  & \cellcolor[HTML]{C0C0C0}{\color[HTML]{000000} 6.42}  \\ \hline
                                & \multicolumn{1}{c|}{13.33}                                                & \multicolumn{1}{c|}{3.33}                                                 & \multicolumn{1}{c|}{2.5}                                                  & \multicolumn{1}{c|}{18.33}                                                & \multicolumn{1}{c|}{10}                                                   & \multicolumn{1}{c|}{13.33}                                                & \multicolumn{1}{c|}{5.83}                                                 & \multicolumn{1}{c|}{1.67}                                                 & \multicolumn{1}{c|}{8.33}                                                 & 23.33                                                \\ \cline{2-11} 
\multirow{-2}{*}{\textbf{4}}    & \multicolumn{1}{c|}{\cellcolor[HTML]{C0C0C0}{\color[HTML]{000000} 18.96}} & \multicolumn{1}{c|}{\cellcolor[HTML]{C0C0C0}{\color[HTML]{000000} 8.65}}  & \multicolumn{1}{c|}{\cellcolor[HTML]{C0C0C0}{\color[HTML]{000000} 2.02}}  & \multicolumn{1}{c|}{\cellcolor[HTML]{C0C0C0}{\color[HTML]{000000} 16.9}}  & \multicolumn{1}{c|}{\cellcolor[HTML]{C0C0C0}{\color[HTML]{000000} 12.23}} & \multicolumn{1}{c|}{\cellcolor[HTML]{C0C0C0}{\color[HTML]{000000} 11.79}} & \multicolumn{1}{c|}{\cellcolor[HTML]{C0C0C0}{\color[HTML]{000000} 0.56}}  & \multicolumn{1}{c|}{\cellcolor[HTML]{C0C0C0}{\color[HTML]{000000} 2.09}}  & \multicolumn{1}{c|}{\cellcolor[HTML]{C0C0C0}{\color[HTML]{000000} 11.76}} & \cellcolor[HTML]{C0C0C0}{\color[HTML]{000000} 15.03} \\ \hline
                                & \multicolumn{1}{c|}{5}                                                    & \multicolumn{1}{c|}{18.33}                                                & \multicolumn{1}{c|}{-}                                                    & \multicolumn{1}{c|}{8.33}                                                 & \multicolumn{1}{c|}{10}                                                   & \multicolumn{1}{c|}{12.5}                                                 & \multicolumn{1}{c|}{25}                                                   & \multicolumn{1}{c|}{8.33}                                                 & \multicolumn{1}{c|}{6.67}                                                 & 5.83                                                 \\ \cline{2-11} 
\multirow{-2}{*}{\textbf{5}}    & \multicolumn{1}{c|}{\cellcolor[HTML]{C0C0C0}{\color[HTML]{000000} 11.69}} & \multicolumn{1}{c|}{\cellcolor[HTML]{C0C0C0}{\color[HTML]{000000} 11.68}} & \multicolumn{1}{c|}{\cellcolor[HTML]{C0C0C0}{\color[HTML]{000000} -}}     & \multicolumn{1}{c|}{\cellcolor[HTML]{C0C0C0}{\color[HTML]{000000} 9}}     & \multicolumn{1}{c|}{\cellcolor[HTML]{C0C0C0}{\color[HTML]{000000} 10.44}} & \multicolumn{1}{c|}{\cellcolor[HTML]{C0C0C0}{\color[HTML]{000000} 10.68}} & \multicolumn{1}{c|}{\cellcolor[HTML]{C0C0C0}{\color[HTML]{000000} 16.83}} & \multicolumn{1}{c|}{\cellcolor[HTML]{C0C0C0}{\color[HTML]{000000} 10.57}} & \multicolumn{1}{c|}{\cellcolor[HTML]{C0C0C0}{\color[HTML]{000000} 10.1}}  & \cellcolor[HTML]{C0C0C0}{\color[HTML]{000000} 9}     \\ \hline
                                & \multicolumn{1}{c|}{6.67}                                                 & \multicolumn{1}{c|}{10.83}                                                & \multicolumn{1}{c|}{15}                                                   & \multicolumn{1}{c|}{5}                                                    & \multicolumn{1}{c|}{16.67}                                                & \multicolumn{1}{c|}{-}                                                    & \multicolumn{1}{c|}{12.5}                                                 & \multicolumn{1}{c|}{8.33}                                                 & \multicolumn{1}{c|}{25}                                                   & -                                                    \\ \cline{2-11} 
\multirow{-2}{*}{\textbf{6}}    & \multicolumn{1}{c|}{\cellcolor[HTML]{C0C0C0}{\color[HTML]{000000} 10.41}} & \multicolumn{1}{c|}{\cellcolor[HTML]{C0C0C0}{\color[HTML]{000000} 13.46}} & \multicolumn{1}{c|}{\cellcolor[HTML]{C0C0C0}{\color[HTML]{000000} 17.59}} & \multicolumn{1}{c|}{\cellcolor[HTML]{C0C0C0}{\color[HTML]{000000} 9.02}}  & \multicolumn{1}{c|}{\cellcolor[HTML]{C0C0C0}{\color[HTML]{000000} 16.65}} & \multicolumn{1}{c|}{\cellcolor[HTML]{C0C0C0}{\color[HTML]{000000} -}}     & \multicolumn{1}{c|}{\cellcolor[HTML]{C0C0C0}{\color[HTML]{000000} 9.02}}  & \multicolumn{1}{c|}{\cellcolor[HTML]{C0C0C0}{\color[HTML]{000000} 11.11}} & \multicolumn{1}{c|}{\cellcolor[HTML]{C0C0C0}{\color[HTML]{000000} 12.72}} & \cellcolor[HTML]{C0C0C0}{\color[HTML]{000000} -}     \\ \hline
                                & \multicolumn{1}{c|}{16.67}                                                & \multicolumn{1}{c|}{7.5}                                                  & \multicolumn{1}{c|}{13.33}                                                & \multicolumn{1}{c|}{-}                                                    & \multicolumn{1}{c|}{7.5}                                                  & \multicolumn{1}{c|}{25}                                                   & \multicolumn{1}{c|}{-}                                                    & \multicolumn{1}{c|}{-}                                                    & \multicolumn{1}{c|}{26.67}                                                & 3.33                                                 \\ \cline{2-11} 
\multirow{-2}{*}{\textbf{7}}    & \multicolumn{1}{c|}{\cellcolor[HTML]{C0C0C0}{\color[HTML]{000000} 15.8}}  & \multicolumn{1}{c|}{\cellcolor[HTML]{C0C0C0}{\color[HTML]{000000} 11.83}} & \multicolumn{1}{c|}{\cellcolor[HTML]{C0C0C0}{\color[HTML]{000000} 16.77}} & \multicolumn{1}{c|}{\cellcolor[HTML]{C0C0C0}{\color[HTML]{000000} -}}     & \multicolumn{1}{c|}{\cellcolor[HTML]{C0C0C0}{\color[HTML]{000000} 13.67}} & \multicolumn{1}{c|}{\cellcolor[HTML]{C0C0C0}{\color[HTML]{000000} 16.74}} & \multicolumn{1}{c|}{\cellcolor[HTML]{C0C0C0}{\color[HTML]{000000} -}}     & \multicolumn{1}{c|}{\cellcolor[HTML]{C0C0C0}{\color[HTML]{000000} -}}     & \multicolumn{1}{c|}{\cellcolor[HTML]{C0C0C0}{\color[HTML]{000000} 12.98}} & \cellcolor[HTML]{C0C0C0}{\color[HTML]{000000} 12.2}  \\ \hline
                                & \multicolumn{1}{c|}{-}                                                    & \multicolumn{1}{c|}{-}                                                    & \multicolumn{1}{c|}{33.33}                                                & \multicolumn{1}{c|}{5}                                                    & \multicolumn{1}{c|}{-}                                                    & \multicolumn{1}{c|}{-}                                                    & \multicolumn{1}{c|}{26.67}                                                & \multicolumn{1}{c|}{3.33}                                                 & \multicolumn{1}{c|}{-}                                                    & 31.67                                                \\ \cline{2-11} 
\multirow{-2}{*}{\textbf{8}}    & \multicolumn{1}{c|}{\cellcolor[HTML]{C0C0C0}{\color[HTML]{000000} -}}     & \multicolumn{1}{c|}{\cellcolor[HTML]{C0C0C0}{\color[HTML]{000000} -}}     & \multicolumn{1}{c|}{\cellcolor[HTML]{C0C0C0}{\color[HTML]{000000} 24.54}} & \multicolumn{1}{c|}{\cellcolor[HTML]{C0C0C0}{\color[HTML]{000000} 17.32}} & \multicolumn{1}{c|}{\cellcolor[HTML]{C0C0C0}{\color[HTML]{000000} -}}     & \multicolumn{1}{c|}{\cellcolor[HTML]{C0C0C0}{\color[HTML]{000000} -}}     & \multicolumn{1}{c|}{\cellcolor[HTML]{C0C0C0}{\color[HTML]{000000} 23.63}} & \multicolumn{1}{c|}{\cellcolor[HTML]{C0C0C0}{\color[HTML]{000000} 15.51}} & \multicolumn{1}{c|}{\cellcolor[HTML]{C0C0C0}{\color[HTML]{000000} -}}     & \cellcolor[HTML]{C0C0C0}{\color[HTML]{000000} 19.08} \\ \hline
                                & \multicolumn{1}{c|}{-}                                                    & \multicolumn{1}{c|}{-}                                                    & \multicolumn{1}{c|}{-}                                                    & \multicolumn{1}{c|}{41.67}                                                & \multicolumn{1}{c|}{-}                                                    & \multicolumn{1}{c|}{8.33}                                                 & \multicolumn{1}{c|}{-}                                                    & \multicolumn{1}{c|}{-}                                                    & \multicolumn{1}{c|}{-}                                                    & 50                                                   \\ \cline{2-11} 
\multirow{-2}{*}{\textbf{9}}    & \multicolumn{1}{c|}{\cellcolor[HTML]{C0C0C0}{\color[HTML]{000000} -}}     & \multicolumn{1}{c|}{\cellcolor[HTML]{C0C0C0}{\color[HTML]{000000} -}}     & \multicolumn{1}{c|}{\cellcolor[HTML]{C0C0C0}{\color[HTML]{000000} -}}     & \multicolumn{1}{c|}{\cellcolor[HTML]{C0C0C0}{\color[HTML]{000000} 38.99}} & \multicolumn{1}{c|}{\cellcolor[HTML]{C0C0C0}{\color[HTML]{000000} -}}     & \multicolumn{1}{c|}{\cellcolor[HTML]{C0C0C0}{\color[HTML]{000000} 24.89}} & \multicolumn{1}{c|}{\cellcolor[HTML]{C0C0C0}{\color[HTML]{000000} -}}     & \multicolumn{1}{c|}{\cellcolor[HTML]{C0C0C0}{\color[HTML]{000000} -}}     & \multicolumn{1}{c|}{\cellcolor[HTML]{C0C0C0}{\color[HTML]{000000} -}}     & \cellcolor[HTML]{C0C0C0}{\color[HTML]{000000} 36.13} \\ \hline
                                & \multicolumn{1}{c|}{-}                                                    & \multicolumn{1}{c|}{-}                                                    & \multicolumn{1}{c|}{-}                                                    & \multicolumn{1}{c|}{-}                                                    & \multicolumn{1}{c|}{-}                                                    & \multicolumn{1}{c|}{-}                                                    & \multicolumn{1}{c|}{-}                                                    & \multicolumn{1}{c|}{100}                                                  & \multicolumn{1}{c|}{-}                                                    & -                                                    \\ \cline{2-11} 
\multirow{-2}{*}{\textbf{10}}   & \multicolumn{1}{c|}{\cellcolor[HTML]{C0C0C0}{\color[HTML]{000000} -}}     & \multicolumn{1}{c|}{\cellcolor[HTML]{C0C0C0}{\color[HTML]{000000} -}}     & \multicolumn{1}{c|}{\cellcolor[HTML]{C0C0C0}{\color[HTML]{000000} -}}     & \multicolumn{1}{c|}{\cellcolor[HTML]{C0C0C0}{\color[HTML]{000000} -}}     & \multicolumn{1}{c|}{\cellcolor[HTML]{C0C0C0}{\color[HTML]{000000} -}}     & \multicolumn{1}{c|}{\cellcolor[HTML]{C0C0C0}{\color[HTML]{000000} -}}     & \multicolumn{1}{c|}{\cellcolor[HTML]{C0C0C0}{\color[HTML]{000000} -}}     & \multicolumn{1}{c|}{\cellcolor[HTML]{C0C0C0}{\color[HTML]{000000} 100}}   & \multicolumn{1}{c|}{\cellcolor[HTML]{C0C0C0}{\color[HTML]{000000} -}}     & \cellcolor[HTML]{C0C0C0}{\color[HTML]{000000} -}     \\ \hline
\end{tabular}
}
\label{tab:fam_res}
\end{table}

\begin{table}[h]
\centering
\caption{The results of \texttt{Decaf} on CIFAR-10.}
\resizebox{\linewidth}{!}{
\begin{tabular}{|c|cccccccccc|}
\hline
                                & \multicolumn{10}{c|}{}                                                                                                                                                                                                                                                                                                                                                                                                                                                                                                                                                                                                                                                                                                                                           \\
\multirow{-2}{*}{\textbf{User}} & \multicolumn{10}{c|}{\multirow{-2}{*}{\textbf{Data Composition (\%)}}}                                                                                                                                                                                                                                                                                                                                                                                                                                                                                                                                                                                                                                                                                           \\ \hline
                                & \multicolumn{1}{c|}{10}                                                   & \multicolumn{1}{c|}{10}                                                   & \multicolumn{1}{c|}{10}                                                   & \multicolumn{1}{c|}{10}                                                   & \multicolumn{1}{c|}{10}                                                   & \multicolumn{1}{c|}{10}                                                   & \multicolumn{1}{c|}{10}                                                   & \multicolumn{1}{c|}{10}                                                   & \multicolumn{1}{c|}{10}                                                   & 10                                                   \\ \cline{2-11} 
\multirow{-2}{*}{\textbf{1}}    & \multicolumn{1}{c|}{\cellcolor[HTML]{C0C0C0}{\color[HTML]{000000} 9.69}}  & \multicolumn{1}{c|}{\cellcolor[HTML]{C0C0C0}{\color[HTML]{000000} 10.47}} & \multicolumn{1}{c|}{\cellcolor[HTML]{C0C0C0}{\color[HTML]{000000} 10.62}} & \multicolumn{1}{c|}{\cellcolor[HTML]{C0C0C0}{\color[HTML]{000000} 9.72}}  & \multicolumn{1}{c|}{\cellcolor[HTML]{C0C0C0}{\color[HTML]{000000} 10.07}} & \multicolumn{1}{c|}{\cellcolor[HTML]{C0C0C0}{\color[HTML]{000000} 9.46}}  & \multicolumn{1}{c|}{\cellcolor[HTML]{C0C0C0}{\color[HTML]{000000} 9.75}}  & \multicolumn{1}{c|}{\cellcolor[HTML]{C0C0C0}{\color[HTML]{000000} 11.3}}  & \multicolumn{1}{c|}{\cellcolor[HTML]{C0C0C0}{\color[HTML]{000000} 9.48}}  & \cellcolor[HTML]{C0C0C0}{\color[HTML]{000000} 9.43}  \\ \hline
                                & \multicolumn{1}{c|}{6}                                                    & \multicolumn{1}{c|}{10}                                                   & \multicolumn{1}{c|}{8}                                                    & \multicolumn{1}{c|}{7}                                                    & \multicolumn{1}{c|}{14}                                                   & \multicolumn{1}{c|}{17}                                                   & \multicolumn{1}{c|}{5}                                                    & \multicolumn{1}{c|}{10}                                                   & \multicolumn{1}{c|}{13}                                                   & 10                                                   \\ \cline{2-11} 
\multirow{-2}{*}{\textbf{2}}    & \multicolumn{1}{c|}{\cellcolor[HTML]{C0C0C0}{\color[HTML]{000000} 8.45}}  & \multicolumn{1}{c|}{\cellcolor[HTML]{C0C0C0}{\color[HTML]{000000} 9.88}}  & \multicolumn{1}{c|}{\cellcolor[HTML]{C0C0C0}{\color[HTML]{000000} 10.56}} & \multicolumn{1}{c|}{\cellcolor[HTML]{C0C0C0}{\color[HTML]{000000} 5.88}}  & \multicolumn{1}{c|}{\cellcolor[HTML]{C0C0C0}{\color[HTML]{000000} 12.76}} & \multicolumn{1}{c|}{\cellcolor[HTML]{C0C0C0}{\color[HTML]{000000} 12.28}} & \multicolumn{1}{c|}{\cellcolor[HTML]{C0C0C0}{\color[HTML]{000000} 6}}     & \multicolumn{1}{c|}{\cellcolor[HTML]{C0C0C0}{\color[HTML]{000000} 10.71}} & \multicolumn{1}{c|}{\cellcolor[HTML]{C0C0C0}{\color[HTML]{000000} 14.02}} & \cellcolor[HTML]{C0C0C0}{\color[HTML]{000000} 9.46}  \\ \hline
                                & \multicolumn{1}{c|}{16}                                                   & \multicolumn{1}{c|}{17}                                                   & \multicolumn{1}{c|}{7}                                                    & \multicolumn{1}{c|}{4}                                                    & \multicolumn{1}{c|}{16}                                                   & \multicolumn{1}{c|}{13}                                                   & \multicolumn{1}{c|}{14}                                                   & \multicolumn{1}{c|}{6}                                                    & \multicolumn{1}{c|}{3}                                                    & 4                                                    \\ \cline{2-11} 
\multirow{-2}{*}{\textbf{3}}    & \multicolumn{1}{c|}{\cellcolor[HTML]{C0C0C0}{\color[HTML]{000000} 15.01}} & \multicolumn{1}{c|}{\cellcolor[HTML]{C0C0C0}{\color[HTML]{000000} 15.72}} & \multicolumn{1}{c|}{\cellcolor[HTML]{C0C0C0}{\color[HTML]{000000} 11.83}} & \multicolumn{1}{c|}{\cellcolor[HTML]{C0C0C0}{\color[HTML]{000000} 4.25}}  & \multicolumn{1}{c|}{\cellcolor[HTML]{C0C0C0}{\color[HTML]{000000} 17.46}} & \multicolumn{1}{c|}{\cellcolor[HTML]{C0C0C0}{\color[HTML]{000000} 13.13}} & \multicolumn{1}{c|}{\cellcolor[HTML]{C0C0C0}{\color[HTML]{000000} 11.62}} & \multicolumn{1}{c|}{\cellcolor[HTML]{C0C0C0}{\color[HTML]{000000} 5.73}}  & \multicolumn{1}{c|}{\cellcolor[HTML]{C0C0C0}{\color[HTML]{000000} 2.62}}  & \cellcolor[HTML]{C0C0C0}{\color[HTML]{000000} 2.62}  \\ \hline
                                & \multicolumn{1}{c|}{14}                                                   & \multicolumn{1}{c|}{2}                                                    & \multicolumn{1}{c|}{1}                                                    & \multicolumn{1}{c|}{20}                                                   & \multicolumn{1}{c|}{10}                                                   & \multicolumn{1}{c|}{14}                                                   & \multicolumn{1}{c|}{5}                                                    & \multicolumn{1}{c|}{2}                                                    & \multicolumn{1}{c|}{8}                                                    & 24                                                   \\ \cline{2-11} 
\multirow{-2}{*}{\textbf{4}}    & \multicolumn{1}{c|}{\cellcolor[HTML]{C0C0C0}{\color[HTML]{000000} 13.76}} & \multicolumn{1}{c|}{\cellcolor[HTML]{C0C0C0}{\color[HTML]{000000} 5.23}}  & \multicolumn{1}{c|}{\cellcolor[HTML]{C0C0C0}{\color[HTML]{000000} 0.8}}   & \multicolumn{1}{c|}{\cellcolor[HTML]{C0C0C0}{\color[HTML]{000000} 16.47}} & \multicolumn{1}{c|}{\cellcolor[HTML]{C0C0C0}{\color[HTML]{000000} 12.89}} & \multicolumn{1}{c|}{\cellcolor[HTML]{C0C0C0}{\color[HTML]{000000} 13.33}} & \multicolumn{1}{c|}{\cellcolor[HTML]{C0C0C0}{\color[HTML]{000000} 9.49}}  & \multicolumn{1}{c|}{\cellcolor[HTML]{C0C0C0}{\color[HTML]{000000} 2.29}}  & \multicolumn{1}{c|}{\cellcolor[HTML]{C0C0C0}{\color[HTML]{000000} 9.48}}  & \cellcolor[HTML]{C0C0C0}{\color[HTML]{000000} 15.65} \\ \hline
                                & \multicolumn{1}{c|}{4}                                                    & \multicolumn{1}{c|}{18}                                                   & \multicolumn{1}{c|}{-}                                                    & \multicolumn{1}{c|}{8}                                                    & \multicolumn{1}{c|}{10}                                                   & \multicolumn{1}{c|}{13}                                                   & \multicolumn{1}{c|}{28}                                                   & \multicolumn{1}{c|}{8}                                                    & \multicolumn{1}{c|}{6}                                                    & 5                                                    \\ \cline{2-11} 
\multirow{-2}{*}{\textbf{5}}    & \multicolumn{1}{c|}{\cellcolor[HTML]{C0C0C0}{\color[HTML]{000000} 6.48}}  & \multicolumn{1}{c|}{\cellcolor[HTML]{C0C0C0}{\color[HTML]{000000} 20.48}} & \multicolumn{1}{c|}{\cellcolor[HTML]{C0C0C0}{\color[HTML]{000000} -}}     & \multicolumn{1}{c|}{\cellcolor[HTML]{C0C0C0}{\color[HTML]{000000} 8.32}}  & \multicolumn{1}{c|}{\cellcolor[HTML]{C0C0C0}{\color[HTML]{000000} 10.96}} & \multicolumn{1}{c|}{\cellcolor[HTML]{C0C0C0}{\color[HTML]{000000} 14.66}} & \multicolumn{1}{c|}{\cellcolor[HTML]{C0C0C0}{\color[HTML]{000000} 17.2}}  & \multicolumn{1}{c|}{\cellcolor[HTML]{C0C0C0}{\color[HTML]{000000} 7.25}}  & \multicolumn{1}{c|}{\cellcolor[HTML]{C0C0C0}{\color[HTML]{000000} 9.35}}  & \cellcolor[HTML]{C0C0C0}{\color[HTML]{000000} 5.3}   \\ \hline
                                & \multicolumn{1}{c|}{6}                                                    & \multicolumn{1}{c|}{11}                                                   & \multicolumn{1}{c|}{16}                                                   & \multicolumn{1}{c|}{4}                                                    & \multicolumn{1}{c|}{16}                                                   & \multicolumn{1}{c|}{-}                                                    & \multicolumn{1}{c|}{13}                                                   & \multicolumn{1}{c|}{8}                                                    & \multicolumn{1}{c|}{26}                                                   & -                                                    \\ \cline{2-11} 
\multirow{-2}{*}{\textbf{6}}    & \multicolumn{1}{c|}{\cellcolor[HTML]{C0C0C0}{\color[HTML]{000000} 9.96}}  & \multicolumn{1}{c|}{\cellcolor[HTML]{C0C0C0}{\color[HTML]{000000} 11.5}}  & \multicolumn{1}{c|}{\cellcolor[HTML]{C0C0C0}{\color[HTML]{000000} 15.15}} & \multicolumn{1}{c|}{\cellcolor[HTML]{C0C0C0}{\color[HTML]{000000} 9.96}}  & \multicolumn{1}{c|}{\cellcolor[HTML]{C0C0C0}{\color[HTML]{000000} 15.81}} & \multicolumn{1}{c|}{\cellcolor[HTML]{C0C0C0}{\color[HTML]{000000} -}}     & \multicolumn{1}{c|}{\cellcolor[HTML]{C0C0C0}{\color[HTML]{000000} 10.24}} & \multicolumn{1}{c|}{\cellcolor[HTML]{C0C0C0}{\color[HTML]{000000} 10.6}}  & \multicolumn{1}{c|}{\cellcolor[HTML]{C0C0C0}{\color[HTML]{000000} 16.77}} & \cellcolor[HTML]{C0C0C0}{\color[HTML]{000000} -}     \\ \hline
                                & \multicolumn{1}{c|}{18}                                                   & \multicolumn{1}{c|}{7}                                                    & \multicolumn{1}{c|}{14}                                                   & \multicolumn{1}{c|}{-}                                                    & \multicolumn{1}{c|}{7}                                                    & \multicolumn{1}{c|}{24}                                                   & \multicolumn{1}{c|}{-}                                                    & \multicolumn{1}{c|}{-}                                                    & \multicolumn{1}{c|}{26}                                                   & 4                                                    \\ \cline{2-11} 
\multirow{-2}{*}{\textbf{7}}    & \multicolumn{1}{c|}{\cellcolor[HTML]{C0C0C0}{\color[HTML]{000000} 17.7}}  & \multicolumn{1}{c|}{\cellcolor[HTML]{C0C0C0}{\color[HTML]{000000} 10.75}} & \multicolumn{1}{c|}{\cellcolor[HTML]{C0C0C0}{\color[HTML]{000000} 19.52}} & \multicolumn{1}{c|}{\cellcolor[HTML]{C0C0C0}{\color[HTML]{000000} -}}     & \multicolumn{1}{c|}{\cellcolor[HTML]{C0C0C0}{\color[HTML]{000000} 6.56}}  & \multicolumn{1}{c|}{\cellcolor[HTML]{C0C0C0}{\color[HTML]{000000} 14.36}} & \multicolumn{1}{c|}{\cellcolor[HTML]{C0C0C0}{\color[HTML]{000000} -}}     & \multicolumn{1}{c|}{\cellcolor[HTML]{C0C0C0}{\color[HTML]{000000} -}}     & \multicolumn{1}{c|}{\cellcolor[HTML]{C0C0C0}{\color[HTML]{000000} 24.54}} & \cellcolor[HTML]{C0C0C0}{\color[HTML]{000000} 6.56}  \\ \hline
                                & \multicolumn{1}{c|}{-}                                                    & \multicolumn{1}{c|}{-}                                                    & \multicolumn{1}{c|}{34}                                                   & \multicolumn{1}{c|}{4}                                                    & \multicolumn{1}{c|}{-}                                                    & \multicolumn{1}{c|}{-}                                                    & \multicolumn{1}{c|}{26}                                                   & \multicolumn{1}{c|}{2}                                                    & \multicolumn{1}{c|}{-}                                                    & 34                                                   \\ \cline{2-11} 
\multirow{-2}{*}{\textbf{8}}    & \multicolumn{1}{c|}{\cellcolor[HTML]{C0C0C0}{\color[HTML]{000000} -}}     & \multicolumn{1}{c|}{\cellcolor[HTML]{C0C0C0}{\color[HTML]{000000} -}}     & \multicolumn{1}{c|}{\cellcolor[HTML]{C0C0C0}{\color[HTML]{000000} 30.19}} & \multicolumn{1}{c|}{\cellcolor[HTML]{C0C0C0}{\color[HTML]{000000} 11.49}} & \multicolumn{1}{c|}{\cellcolor[HTML]{C0C0C0}{\color[HTML]{000000} -}}     & \multicolumn{1}{c|}{\cellcolor[HTML]{C0C0C0}{\color[HTML]{000000} -}}     & \multicolumn{1}{c|}{\cellcolor[HTML]{C0C0C0}{\color[HTML]{000000} 25.29}} & \multicolumn{1}{c|}{\cellcolor[HTML]{C0C0C0}{\color[HTML]{000000} 9.12}}  & \multicolumn{1}{c|}{\cellcolor[HTML]{C0C0C0}{\color[HTML]{000000} -}}     & \cellcolor[HTML]{C0C0C0}{\color[HTML]{000000} 23.91} \\ \hline
                                & \multicolumn{1}{c|}{-}                                                    & \multicolumn{1}{c|}{-}                                                    & \multicolumn{1}{c|}{-}                                                    & \multicolumn{1}{c|}{40}                                                   & \multicolumn{1}{c|}{-}                                                    & \multicolumn{1}{c|}{10}                                                   & \multicolumn{1}{c|}{-}                                                    & \multicolumn{1}{c|}{-}                                                    & \multicolumn{1}{c|}{-}                                                    & 50                                                   \\ \cline{2-11} 
\multirow{-2}{*}{\textbf{9}}    & \multicolumn{1}{c|}{\cellcolor[HTML]{C0C0C0}{\color[HTML]{000000} -}}     & \multicolumn{1}{c|}{\cellcolor[HTML]{C0C0C0}{\color[HTML]{000000} -}}     & \multicolumn{1}{c|}{\cellcolor[HTML]{C0C0C0}{\color[HTML]{000000} -}}     & \multicolumn{1}{c|}{\cellcolor[HTML]{C0C0C0}{\color[HTML]{000000} 36.75}} & \multicolumn{1}{c|}{\cellcolor[HTML]{C0C0C0}{\color[HTML]{000000} -}}     & \multicolumn{1}{c|}{\cellcolor[HTML]{C0C0C0}{\color[HTML]{000000} 25.32}} & \multicolumn{1}{c|}{\cellcolor[HTML]{C0C0C0}{\color[HTML]{000000} -}}     & \multicolumn{1}{c|}{\cellcolor[HTML]{C0C0C0}{\color[HTML]{000000} -}}     & \multicolumn{1}{c|}{\cellcolor[HTML]{C0C0C0}{\color[HTML]{000000} -}}     & \cellcolor[HTML]{C0C0C0}{\color[HTML]{000000} 37.92} \\ \hline
                                & \multicolumn{1}{c|}{-}                                                    & \multicolumn{1}{c|}{-}                                                    & \multicolumn{1}{c|}{-}                                                    & \multicolumn{1}{c|}{-}                                                    & \multicolumn{1}{c|}{-}                                                    & \multicolumn{1}{c|}{-}                                                    & \multicolumn{1}{c|}{-}                                                    & \multicolumn{1}{c|}{100}                                                  & \multicolumn{1}{c|}{-}                                                    & -                                                    \\ \cline{2-11} 
\multirow{-2}{*}{\textbf{10}}   & \multicolumn{1}{c|}{\cellcolor[HTML]{C0C0C0}{\color[HTML]{000000} -}}     & \multicolumn{1}{c|}{\cellcolor[HTML]{C0C0C0}{\color[HTML]{000000} -}}     & \multicolumn{1}{c|}{\cellcolor[HTML]{C0C0C0}{\color[HTML]{000000} -}}     & \multicolumn{1}{c|}{\cellcolor[HTML]{C0C0C0}{\color[HTML]{000000} -}}     & \multicolumn{1}{c|}{\cellcolor[HTML]{C0C0C0}{\color[HTML]{000000} -}}     & \multicolumn{1}{c|}{\cellcolor[HTML]{C0C0C0}{\color[HTML]{000000} -}}     & \multicolumn{1}{c|}{\cellcolor[HTML]{C0C0C0}{\color[HTML]{000000} -}}     & \multicolumn{1}{c|}{\cellcolor[HTML]{C0C0C0}{\color[HTML]{000000} 100}}   & \multicolumn{1}{c|}{\cellcolor[HTML]{C0C0C0}{\color[HTML]{000000} -}}     & \cellcolor[HTML]{C0C0C0}{\color[HTML]{000000} -}     \\ \hline
\end{tabular}
}
\label{tab:cifar10_res}
\end{table}

\begin{table}[h]
\centering
\caption{The results of \texttt{Decaf} on FER-2013.}
\resizebox{\linewidth}{!}{
\begin{tabular}{|c|ccccccc|}
\hline
                                & \multicolumn{7}{c|}{}                                                                                                                                                                                                                                                                                                                                       \\
\multirow{-2}{*}{\textbf{User}} & \multicolumn{7}{c|}{\multirow{-2}{*}{\textbf{Data Composition (\%)}}}                                                                                                                                                                                                                                                                                       \\ \hline
                                & \multicolumn{1}{c|}{14.29}                         & \multicolumn{1}{c|}{14.29}                         & \multicolumn{1}{c|}{14.29}                         & \multicolumn{1}{c|}{14.29}                         & \multicolumn{1}{c|}{14.29}                         & \multicolumn{1}{c|}{14.29}                         & 14.29                         \\ \cline{2-8} 
\multirow{-2}{*}{\textbf{1}}    & \multicolumn{1}{c|}{\cellcolor[HTML]{C0C0C0}14.97} & \multicolumn{1}{c|}{\cellcolor[HTML]{C0C0C0}17.13} & \multicolumn{1}{c|}{\cellcolor[HTML]{C0C0C0}14.26} & \multicolumn{1}{c|}{\cellcolor[HTML]{C0C0C0}15.1}  & \multicolumn{1}{c|}{\cellcolor[HTML]{C0C0C0}12.46} & \multicolumn{1}{c|}{\cellcolor[HTML]{C0C0C0}13.71} & \cellcolor[HTML]{C0C0C0}12.37 \\ \hline
                                & \multicolumn{1}{c|}{19.23}                         & \multicolumn{1}{c|}{5.77}                          & \multicolumn{1}{c|}{15.38}                         & \multicolumn{1}{c|}{3.85}                          & \multicolumn{1}{c|}{26.92}                         & \multicolumn{1}{c|}{13.46}                         & 15.38                         \\ \cline{2-8} 
\multirow{-2}{*}{\textbf{2}}    & \multicolumn{1}{c|}{\cellcolor[HTML]{C0C0C0}21.37} & \multicolumn{1}{c|}{\cellcolor[HTML]{C0C0C0}12.77} & \multicolumn{1}{c|}{\cellcolor[HTML]{C0C0C0}14.06} & \multicolumn{1}{c|}{\cellcolor[HTML]{C0C0C0}1.75}  & \multicolumn{1}{c|}{\cellcolor[HTML]{C0C0C0}20.48} & \multicolumn{1}{c|}{\cellcolor[HTML]{C0C0C0}12.37} & \cellcolor[HTML]{C0C0C0}17.19 \\ \hline
                                & \multicolumn{1}{c|}{5.17}                          & \multicolumn{1}{c|}{7.76}                          & \multicolumn{1}{c|}{13.79}                         & \multicolumn{1}{c|}{24.14}                         & \multicolumn{1}{c|}{12.07}                         & \multicolumn{1}{c|}{19.83}                         & 17.24                         \\ \cline{2-8} 
\multirow{-2}{*}{\textbf{3}}    & \multicolumn{1}{c|}{\cellcolor[HTML]{C0C0C0}10.92} & \multicolumn{1}{c|}{\cellcolor[HTML]{C0C0C0}14.66} & \multicolumn{1}{c|}{\cellcolor[HTML]{C0C0C0}15.27} & \multicolumn{1}{c|}{\cellcolor[HTML]{C0C0C0}19.92} & \multicolumn{1}{c|}{\cellcolor[HTML]{C0C0C0}10.92} & \multicolumn{1}{c|}{\cellcolor[HTML]{C0C0C0}13.52} & \cellcolor[HTML]{C0C0C0}14.78 \\ \hline
                                & \multicolumn{1}{c|}{29.2}                          & \multicolumn{1}{c|}{-}                             & \multicolumn{1}{c|}{7.3}                           & \multicolumn{1}{c|}{21.9}                          & \multicolumn{1}{c|}{8.76}                          & \multicolumn{1}{c|}{10.95}                         & 21.9                          \\ \cline{2-8} 
\multirow{-2}{*}{\textbf{4}}    & \multicolumn{1}{c|}{\cellcolor[HTML]{C0C0C0}24.97} & \multicolumn{1}{c|}{\cellcolor[HTML]{C0C0C0}-}     & \multicolumn{1}{c|}{\cellcolor[HTML]{C0C0C0}10.3}  & \multicolumn{1}{c|}{\cellcolor[HTML]{C0C0C0}22.67} & \multicolumn{1}{c|}{\cellcolor[HTML]{C0C0C0}9.69}  & \multicolumn{1}{c|}{\cellcolor[HTML]{C0C0C0}14.39} & \cellcolor[HTML]{C0C0C0}17.98 \\ \hline
                                & \multicolumn{1}{c|}{-}                             & \multicolumn{1}{c|}{8.04}                          & \multicolumn{1}{c|}{33.93}                         & \multicolumn{1}{c|}{-}                             & \multicolumn{1}{c|}{26.79}                         & \multicolumn{1}{c|}{31.25}                         & -                             \\ \cline{2-8} 
\multirow{-2}{*}{\textbf{5}}    & \multicolumn{1}{c|}{\cellcolor[HTML]{C0C0C0}-}     & \multicolumn{1}{c|}{\cellcolor[HTML]{C0C0C0}21.44} & \multicolumn{1}{c|}{\cellcolor[HTML]{C0C0C0}27.2}  & \multicolumn{1}{c|}{\cellcolor[HTML]{C0C0C0}-}     & \multicolumn{1}{c|}{\cellcolor[HTML]{C0C0C0}25.42} & \multicolumn{1}{c|}{\cellcolor[HTML]{C0C0C0}25.94} & \cellcolor[HTML]{C0C0C0}-     \\ \hline
                                & \multicolumn{1}{c|}{27.4}                          & \multicolumn{1}{c|}{-}                             & \multicolumn{1}{c|}{47.95}                         & \multicolumn{1}{c|}{-}                             & \multicolumn{1}{c|}{24.66}                         & \multicolumn{1}{c|}{-}                             & -                             \\ \cline{2-8} 
\multirow{-2}{*}{\textbf{6}}    & \multicolumn{1}{c|}{\cellcolor[HTML]{C0C0C0}32.07} & \multicolumn{1}{c|}{\cellcolor[HTML]{C0C0C0}-}     & \multicolumn{1}{c|}{\cellcolor[HTML]{C0C0C0}38.09} & \multicolumn{1}{c|}{\cellcolor[HTML]{C0C0C0}-}     & \multicolumn{1}{c|}{\cellcolor[HTML]{C0C0C0}29.85} & \multicolumn{1}{c|}{\cellcolor[HTML]{C0C0C0}-}     & \cellcolor[HTML]{C0C0C0}-     \\ \hline
                                & \multicolumn{1}{c|}{-}                             & \multicolumn{1}{c|}{-}                             & \multicolumn{1}{c|}{-}                             & \multicolumn{1}{c|}{-}                             & \multicolumn{1}{c|}{-}                             & \multicolumn{1}{c|}{-}                             & 100                           \\ \cline{2-8} 
\multirow{-2}{*}{\textbf{7}}    & \multicolumn{1}{c|}{\cellcolor[HTML]{C0C0C0}-}     & \multicolumn{1}{c|}{\cellcolor[HTML]{C0C0C0}-}     & \multicolumn{1}{c|}{\cellcolor[HTML]{C0C0C0}-}     & \multicolumn{1}{c|}{\cellcolor[HTML]{C0C0C0}-}     & \multicolumn{1}{c|}{\cellcolor[HTML]{C0C0C0}-}     & \multicolumn{1}{c|}{\cellcolor[HTML]{C0C0C0}-}     & \cellcolor[HTML]{C0C0C0}100   \\ \hline
\end{tabular}
}
\label{tab:fer2013_res}
\end{table}

\begin{table}[h]
\centering
\caption{The results of \texttt{Decaf} on SkinCancer.}
\resizebox{\linewidth}{!}{
\begin{tabular}{|c|ccccccc|}
\hline
                                & \multicolumn{7}{c|}{}                                                                                                                                                                                                                                                                                                                                       \\
\multirow{-2}{*}{\textbf{User}} & \multicolumn{7}{c|}{\multirow{-2}{*}{\textbf{Data Composition (\%)}}}                                                                                                                                                                                                                                                                                       \\ \hline
                                & \multicolumn{1}{c|}{14.29}                         & \multicolumn{1}{c|}{14.29}                         & \multicolumn{1}{c|}{14.29}                         & \multicolumn{1}{c|}{14.29}                         & \multicolumn{1}{c|}{14.29}                         & \multicolumn{1}{c|}{14.29}                         & 14.29                         \\ \cline{2-8} 
\multirow{-2}{*}{\textbf{1}}    & \multicolumn{1}{c|}{\cellcolor[HTML]{C0C0C0}15.51} & \multicolumn{1}{c|}{\cellcolor[HTML]{C0C0C0}13.12} & \multicolumn{1}{c|}{\cellcolor[HTML]{C0C0C0}11.41} & \multicolumn{1}{c|}{\cellcolor[HTML]{C0C0C0}13.65} & \multicolumn{1}{c|}{\cellcolor[HTML]{C0C0C0}17.07} & \multicolumn{1}{c|}{\cellcolor[HTML]{C0C0C0}15.88} & \cellcolor[HTML]{C0C0C0}13.36 \\ \hline
                                & \multicolumn{1}{c|}{15.2}                          & \multicolumn{1}{c|}{20.52}                         & \multicolumn{1}{c|}{11.85}                         & \multicolumn{1}{c|}{16}                            & \multicolumn{1}{c|}{12.16}                         & \multicolumn{1}{c|}{14.44}                         & 9.88                          \\ \cline{2-8} 
\multirow{-2}{*}{\textbf{2}}    & \multicolumn{1}{c|}{\cellcolor[HTML]{C0C0C0}14.62} & \multicolumn{1}{c|}{\cellcolor[HTML]{C0C0C0}17.79} & \multicolumn{1}{c|}{\cellcolor[HTML]{C0C0C0}12.16} & \multicolumn{1}{c|}{\cellcolor[HTML]{C0C0C0}14.86} & \multicolumn{1}{c|}{\cellcolor[HTML]{C0C0C0}15.48} & \multicolumn{1}{c|}{\cellcolor[HTML]{C0C0C0}16.45} & \cellcolor[HTML]{C0C0C0}8.65  \\ \hline
                                & \multicolumn{1}{c|}{10.26}                         & \multicolumn{1}{c|}{8.97}                          & \multicolumn{1}{c|}{16.67}                         & \multicolumn{1}{c|}{12.82}                         & \multicolumn{1}{c|}{14.1}                          & \multicolumn{1}{c|}{16.03}                         & 21.15                         \\ \cline{2-8} 
\multirow{-2}{*}{\textbf{3}}    & \multicolumn{1}{c|}{\cellcolor[HTML]{C0C0C0}14.1}  & \multicolumn{1}{c|}{\cellcolor[HTML]{C0C0C0}11.99} & \multicolumn{1}{c|}{\cellcolor[HTML]{C0C0C0}13.37} & \multicolumn{1}{c|}{\cellcolor[HTML]{C0C0C0}12.38} & \multicolumn{1}{c|}{\cellcolor[HTML]{C0C0C0}17.14} & \multicolumn{1}{c|}{\cellcolor[HTML]{C0C0C0}14.87} & \cellcolor[HTML]{C0C0C0}16.16 \\ \hline
                                & \multicolumn{1}{c|}{20}                            & \multicolumn{1}{c|}{-}                             & \multicolumn{1}{c|}{38.46}                         & \multicolumn{1}{c|}{10.77}                         & \multicolumn{1}{c|}{7.69}                          & \multicolumn{1}{c|}{7.69}                          & 15.38                         \\ \cline{2-8} 
\multirow{-2}{*}{\textbf{4}}    & \multicolumn{1}{c|}{\cellcolor[HTML]{C0C0C0}15.55} & \multicolumn{1}{c|}{\cellcolor[HTML]{C0C0C0}-}     & \multicolumn{1}{c|}{\cellcolor[HTML]{C0C0C0}31.49} & \multicolumn{1}{c|}{\cellcolor[HTML]{C0C0C0}16.12} & \multicolumn{1}{c|}{\cellcolor[HTML]{C0C0C0}12.26} & \multicolumn{1}{c|}{\cellcolor[HTML]{C0C0C0}12.7}  & \cellcolor[HTML]{C0C0C0}11.88 \\ \hline
                                & \multicolumn{1}{c|}{9.09}                          & \multicolumn{1}{c|}{16.36}                         & \multicolumn{1}{c|}{-}                             & \multicolumn{1}{c|}{27.27}                         & \multicolumn{1}{c|}{-}                             & \multicolumn{1}{c|}{14.55}                         & 32.73                         \\ \cline{2-8} 
\multirow{-2}{*}{\textbf{5}}    & \multicolumn{1}{c|}{\cellcolor[HTML]{C0C0C0}17.1}  & \multicolumn{1}{c|}{\cellcolor[HTML]{C0C0C0}18.76} & \multicolumn{1}{c|}{\cellcolor[HTML]{C0C0C0}-}     & \multicolumn{1}{c|}{\cellcolor[HTML]{C0C0C0}17.14} & \multicolumn{1}{c|}{\cellcolor[HTML]{C0C0C0}-}     & \multicolumn{1}{c|}{\cellcolor[HTML]{C0C0C0}17.17} & \cellcolor[HTML]{C0C0C0}29.83 \\ \hline
                                & \multicolumn{1}{c|}{-}                             & \multicolumn{1}{c|}{18.84}                         & \multicolumn{1}{c|}{20.29}                         & \multicolumn{1}{c|}{-}                             & \multicolumn{1}{c|}{43.48}                         & \multicolumn{1}{c|}{-}                             & 17.39                         \\ \cline{2-8} 
\multirow{-2}{*}{\textbf{6}}    & \multicolumn{1}{c|}{\cellcolor[HTML]{C0C0C0}-}     & \multicolumn{1}{c|}{\cellcolor[HTML]{C0C0C0}21.15} & \multicolumn{1}{c|}{\cellcolor[HTML]{C0C0C0}22.68} & \multicolumn{1}{c|}{\cellcolor[HTML]{C0C0C0}-}     & \multicolumn{1}{c|}{\cellcolor[HTML]{C0C0C0}38.82} & \multicolumn{1}{c|}{\cellcolor[HTML]{C0C0C0}-}     & \cellcolor[HTML]{C0C0C0}17.34 \\ \hline
                                & \multicolumn{1}{c|}{-}                             & \multicolumn{1}{c|}{100}                           & \multicolumn{1}{c|}{-}                             & \multicolumn{1}{c|}{-}                             & \multicolumn{1}{c|}{-}                             & \multicolumn{1}{c|}{-}                             & -                             \\ \cline{2-8} 
\multirow{-2}{*}{\textbf{7}}    & \multicolumn{1}{c|}{\cellcolor[HTML]{C0C0C0}-}     & \multicolumn{1}{c|}{\cellcolor[HTML]{C0C0C0}100}   & \multicolumn{1}{c|}{\cellcolor[HTML]{C0C0C0}-}     & \multicolumn{1}{c|}{\cellcolor[HTML]{C0C0C0}-}     & \multicolumn{1}{c|}{\cellcolor[HTML]{C0C0C0}-}     & \multicolumn{1}{c|}{\cellcolor[HTML]{C0C0C0}-}     & \cellcolor[HTML]{C0C0C0}-     \\ \hline
\end{tabular}
}
\label{tab:sk_res}
\end{table}

\noindent\textbf{Model Architecture.} To expedite the experiments for extensive evaluations, we leverage non-complex models as summarized in \autoref{tab:model_structure}. We defer the \texttt{Decaf} generalization to relatively complex standard model architectures (e.g., ResNet18, VGG16), which is experimentally validated in \autoref{sec:complex_model}. Note that one convolution module consists of one convolution layer and one pooling layer. For FER-2013 dataset, four convolution modules are used, which is deeper (in particular, with one more convolution module) than the architecture for CIFAR-10 and SkinCancer considering its larger image size.

\noindent\textbf{Machine Hardware.} In our experiments, a personal computer is used to perform FL training and fulfil \texttt{Decaf}. Note that the \texttt{Decaf} attack does not need to be performed simultaneously with the FL training, the attacker can perform it after the completion of the FL training once all historical victim local model updates are recorded by the attacker. The computer has a CPU of 12th Gen Intel(R) Core(TM) i7-12700 2.10 GHz, a 16.0 GB RAM. The operating system is Windows 11 64-bit.

\noindent\textbf{FL Setting.} Our experiments use the most common \texttt{FedAvg} as aggregation method \cite{2DBLP:conf/aistats/McMahanMRHA17}.The optimizer is the \texttt{adadelta}. We assign each user to participate in FL datasets with different distributions to simulate the data heterogeneity in FL. For example, we set one user with an evenly distributed dataset, and three users have data for all classes, but their proportions differ. The remaining six users have datasets that are missing data for 1, 2, 3, 5, 7, and 9 classes, respectively. Specifically distribution settings can be found along with the \texttt{Decaf} result reports, see \autoref{tab:mnist_res}-\autoref{tab:sk_res}. Hence, their data are non-independently and identically (non-IID) distributed. Note that \texttt{Decaf} is insensitive to data distribution by design. Except for the experiments on the FER-2013, and SkinCancer datasets, where we set up 7 federated users to perform the attack experiments, we set up 10 federated users for each of the remaining datasets.

\begin{figure*}[h]
  \centering
  \includegraphics[width=\textwidth]{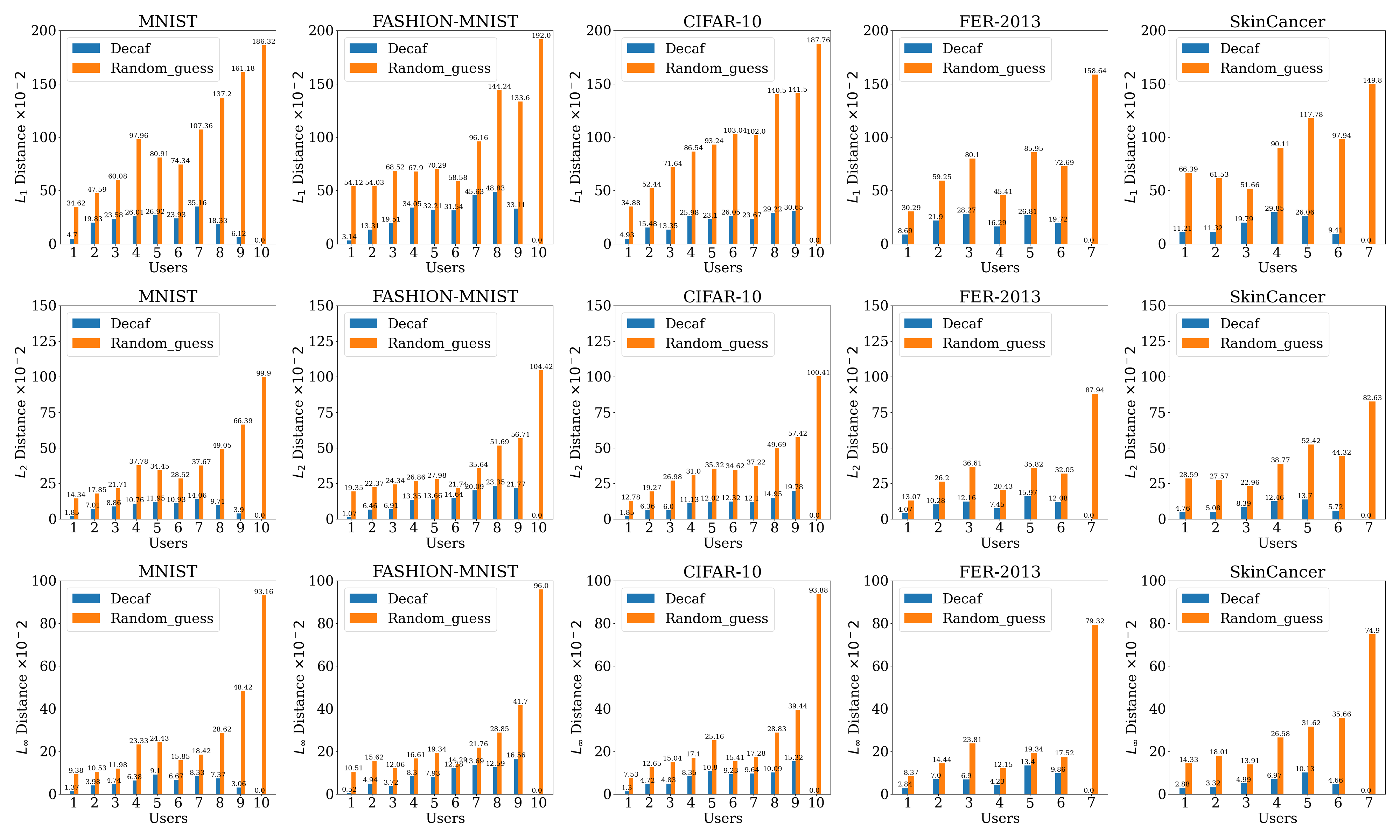}
  \caption{Comparison of \texttt{Decaf} and random guess.}
  \label{Fig_Lp_distance}
\end{figure*}

\subsection{Attack Performance}
\label{sec:attack_performance}
\textbf{Attack Accuracy}. We randomly select one round (multiple-round is also shown shortly when evaluating attack stability) to perform \texttt{Decaf} in the first few FL rounds over five datasets. Each user has data samples with different proportions, which results are detailed in \autoref{tab:mnist_res}-\autoref{tab:sk_res} per dataset. Note that for each user, the decomposed data distribution is gray-boxed. We can find that null-classes are 100\% accurately identified in all cases. The remaining classes are also inferred accurately within a small margin of error. We use the $L_p, p \in \{1, 2, \infty\}$ distances to measure the accuracy of \texttt{Decaf}. The random guesses of data composition are used for comparison shown in \autoref{Fig_Lp_distance}. The $L_p$ distances tend to more likely increase as the data is more non-IID distributed---either more classes are missed or remaining non-null classes are distributed more imbalanced. However, they are still much smaller than the random guesses. This is under expectation because the variance of extracted gradient changes tends to be increased in this context, rendering a relatively increasing difference between the decomposed and ground-truth distribution. It is worth noting that the $L_1$ distance is an accumulation of \textit{absolute differences} between decomposed and ground-truth data distribution. In fact, for users \#1-3 whose data have no null-classes, the $L_1$ distance is often within 0.2 even if the data is non-IID distributed to a large extent. The $L_1$ only deviates relatively notably given that the data is extremely non-IID distributed, and there exists null-class(es). The $L_2$ distance indicates the overall accuracy of the data composition inferences. It has been in a relatively stable state, around 0.1 in most cases. The $L_{\infty}$ distance is usually small, e.g., less than or around 0.1, even though the $L_1$ distance can go high.

\noindent\textbf{Attack Stealthiness}. \texttt{Decaf} does not tamper the user's model parameters or change the FL aggregation process at all---it is a passive attack. So whether the \texttt{Decaf} is performed during the FL learning is not a matter. The FL server can record the victim's local model updates and perform the \texttt{Decaf} anytime. Therefore, it is infeasible for the victim to detect whether the \texttt{Decaf} is launched when the victim is participating in the FL learning.

\begin{figure*}[h]
  \centering
  \includegraphics[width=\textwidth]{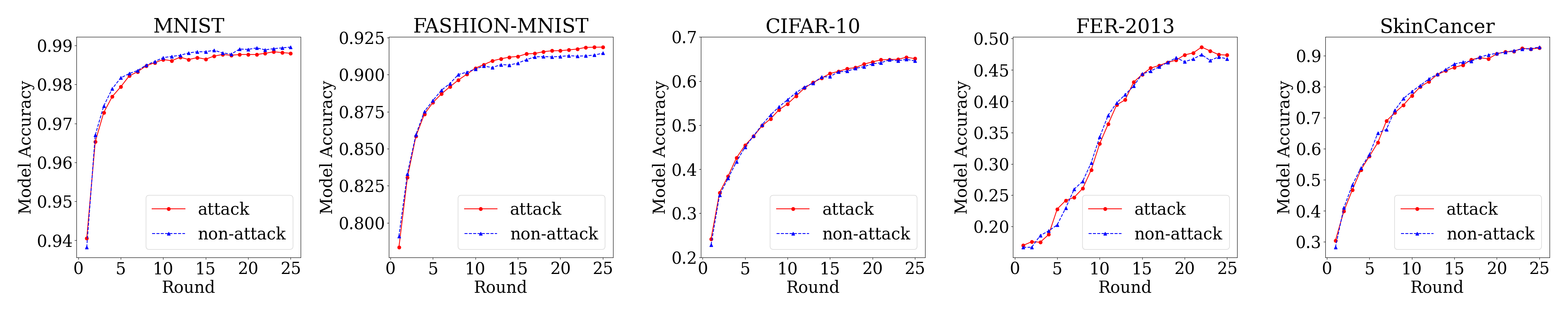}
  \caption{Accuracy of FL global model with and without \texttt{Decaf}.}
  \label{Fig_model_accuracy}
\end{figure*}

We further test the accuracy of the global model for each round until it converges during FL. \autoref{Fig_model_accuracy} shows the global model accuracy with and without \texttt{Decaf}. As can be seen, the accuracy of the global model is not affected in any way. Suppose that \texttt{Decaf} is performed simultaneously during the FL. The attacking overhead (e.g., latency) might be a concern to some extent, as reported in \autoref{tab:decaf_overhead}. It is indeed inducing latency. However, we argue that the \texttt{Decaf} is independent of the FL, which can be performed after the FL. Therefore, there will be no latency in practice. In addition, the attack computation overhead is negligible once this is done after the FL training is finished or the \texttt{Decaf} is performed through an independent machine that is different from the FL server. Note \autoref{tab:decaf_overhead}, only CPU is used, and the additional latency is less than 4 seconds. The normal FL training takes about 24 and 28 seconds per FL round. In this context, the \texttt{Decaf} latency is small or can be hidden in the FL training process even when the \texttt{Decaf} is indeed performed simultaneously with the FL training on the same FL server machine.

\begin{table}[h]
\centering
\caption{\texttt{Decaf} computation overhead. (time in seconds)}
\resizebox{\linewidth}{!}{
\begin{tabular}{|c|c|c|c|}
\hline
\textbf{Decaf (CPU)}                                                                    & \textbf{Task}                   & \textbf{MNIST} & \textbf{CIFAR10} \\ \hline
\multirow{2}{*}{\begin{tabular}[c]{@{}c@{}}FL Learning\\ (Normal)\end{tabular}}          & Regular Aggregation             & 0.009          & 0.003            \\ \cline{2-4} 
                                                                                       & Local Model Training            & 23.439         & 28.437           \\ \hline
\multirow{3}{*}{\begin{tabular}[c]{@{}c@{}}Decaf Overhead\\ (Additional)\end{tabular}} & Null Classes Removal            & 0.001          & 0.001            \\ \cline{2-4} 
                                                                                       & Gradient Bases Construction     & 0.608          & 0.478            \\ \cline{2-4} 
                                                                                       & Remaining Classes Decomposition & 3.589          & 3.196            \\ \hline
\end{tabular}
}
\label{tab:decaf_overhead}
\end{table}

\begin{figure*}[h]
	\centering
	\includegraphics[width=1.0\linewidth]{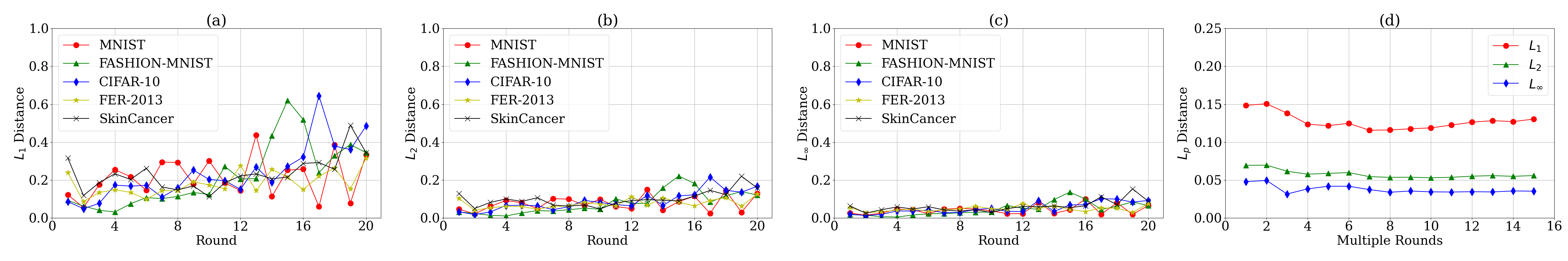}
	\caption{Attack stability of \texttt{Decaf}. Figures (a)-(c) are the single-round attack, and figure (d) is the multi-round attack.}
    \label{Fig_multi_attack}
\end{figure*}
\noindent\textbf{Attack Stability}. \texttt{Decaf} only requires a single round snap-shot of gradient changes in the user's local model so that the attack can be performed at any round. \autoref{Fig_multi_attack}(a)-(c) show the results of \texttt{Decaf} on multiple FL users with different data distributions on five datasets. Here, we compute the $L_p, p\in\{1,2,\infty\}$ distances between the \texttt{Decaf} inferred data composition and the ground truth per FL round of the first 20 rounds. We can observe that all $L_p, p\in\{1,2,\infty\}$ distances decrease in the first few FL rounds. After reaching a minimum around the third FL round, distances gradually increase. The reason is that the global model is randomly initialized. Local models will converge in different directions due to the different data compositions. Thus, there is a relatively large variance when constructing gradient bases from the local models updated upon it. Once the randomness militates after a few rounds, the \texttt{Decaf} achieves the best-attacking performance because the local model is more sensitive as knowledge from other users' data has not been well learned so far. When the FL continues, the global model starts learning global data well, diminishing global/local model sensitivity to per user's data and rendering decreasing attacking performance. In this context, the user's private data induce less gradient change given the (relatively) converged model. Despite that \texttt{Decaf} can still work in the late-stage of the FL learning (e.g., when many FL rounds have passed), it is recommended that the \texttt{Decaf} is performed via early-stage victim's local models that exhibit higher sensitivity in terms of gradient change to the local data distribution.

We also validate the multi-round attack performance of \texttt{Decaf}. Unlike a single-round-based attack, the multi-round attack combines results of $k$ (continuous) rounds of \texttt{Decaf}. The SkinCancer dataset is used. We choose one user (user 2 in \autoref{tab:sk_res}) to perform the multi-round attack with the $k$ rounds from 1 to 15 on SkinCancer. We first average $k$ rounds of decomposed data distribution, where each round is obtained through the single-round attack. We then normalize the averaged data distribution (similar to converting logits to softmax in machine learning to ensure the total probability/proportion is exactly 1.0). After that, the  $L_p$ distances are measured. The multiple-round \texttt{Decaf} attack is shown in \autoref{Fig_multi_attack}-(d). As expected, the multiple-round attack improves the attack performance as all $L_p$ distances decrease. More specifically, the $L_p$ distance first decreases with the number of multiple rounds increasing and then slowly increasing. It is because later rounds of \texttt{Decaf} get progressively worse as the global model converges (detailed in \autoref{sec:attack_performance}).

\subsection{Number of Auxiliary Sample}\label{sec:auxiliary}
\begin{figure}[h]
	\centering
	\includegraphics[width=0.75\linewidth]{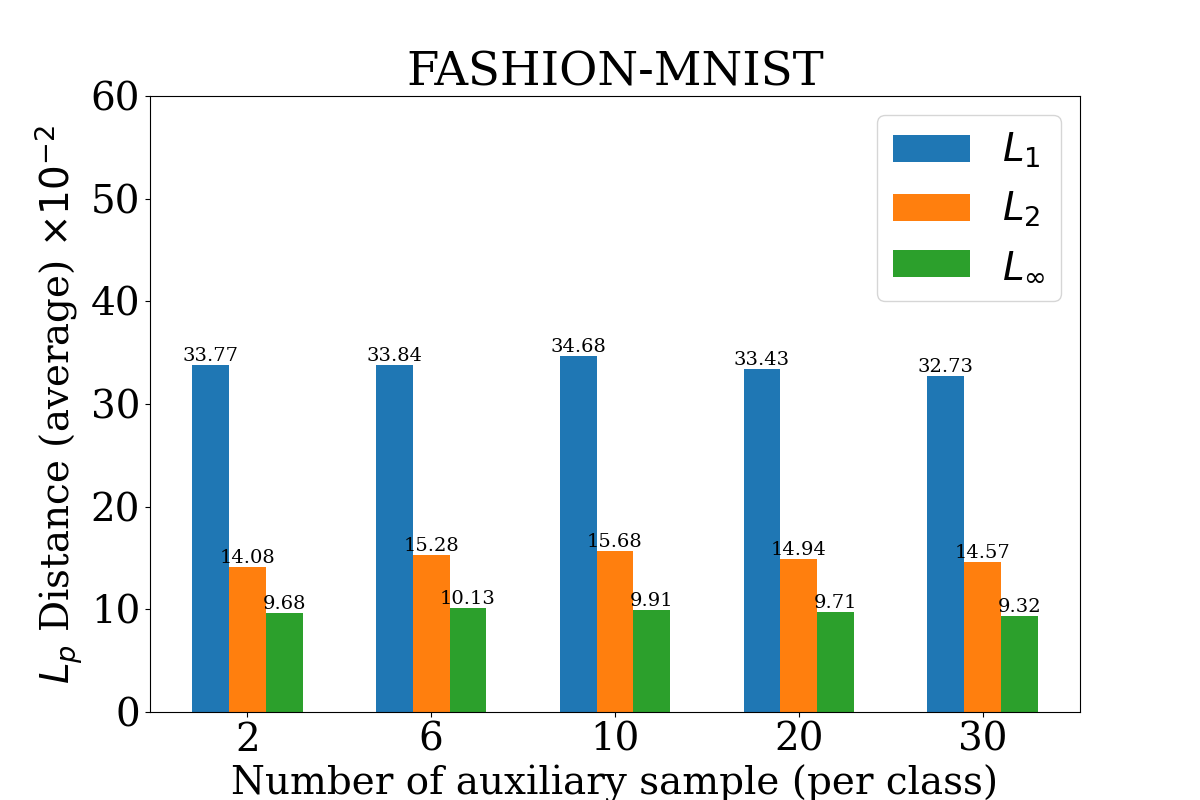}
	\caption{\texttt{Decaf} exhibit similar results on different numbers of auxiliary samples even given quite a few auxiliary samples per class, e.g., 2.}
    \label{Fig_aux}
\end{figure}

Previous experiments consistently use 100 auxiliary samples per class for extensive evaluations on \texttt{Decaf}. We here evaluate \texttt{Decaf} further with a different number of auxiliary samples per class on FASHION-MNIST, as shown in \autoref{Fig_aux}, where all other settings are intact compared to previous experiments. We calculate averaged $L_p$ distances for all users in the first 10 FL rounds. We can find that the $L_p$ distances have remained steady. There is no noticeable change as the server's number of auxiliary samples increases from 2 to 30 per class. It means that the gradient bases barely depend on the number of auxiliary samples, and \texttt{Decaf} can still work stably given few available auxiliary samples, e.g., 2, per class.
\subsection{Neuron Number of Last Fully Connected Layer}\label{sec:layernum}
\begin{figure}[h]
	\centering
	\includegraphics[width=0.75\linewidth]{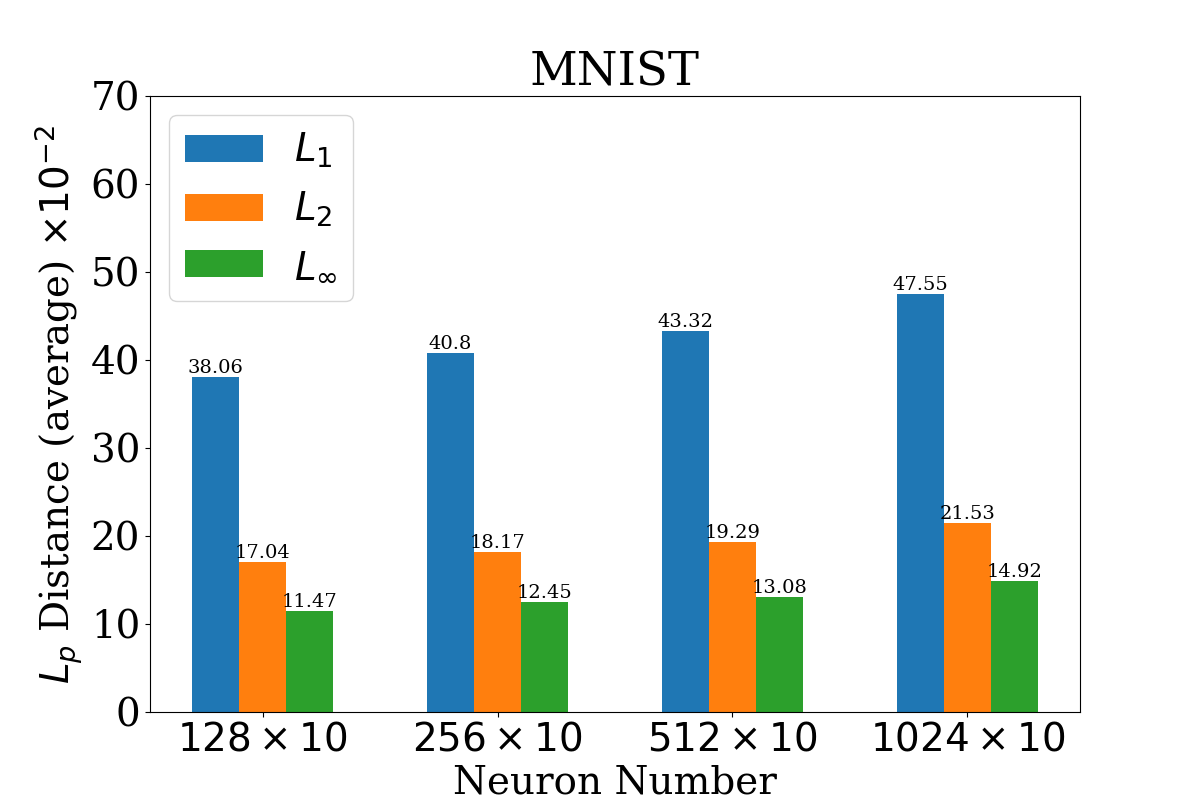}
	\caption{\texttt{Decaf} under different neuron numbers of last fully connected layer on MNIST.}
    \label{Fig_neuron_number}
\end{figure}

We further study the effect of different neuron numbers of the last fully connected layer on \texttt{Decaf}. We attack all users (consistent with users in \autoref{tab:mnist_res}) on MNIST for the first 10 FL rounds. As shown in \autoref{Fig_neuron_number}, the model structure is the same as in the \autoref{tab:model_structure} except that the last fully connected layer is set to $128 \times 10$, $256 \times 10$, $512 \times 10$, and $1024 \times 10$, respectively. We calculate the average $L_p$ distances of all users for the first 10 FL rounds. \autoref{Fig_neuron_number} shows an upward trend in all $L_p$ distances as the number of neurons increases. The smaller the number of neurons, the more accurate \texttt{Decaf} is. However, the $L_p$ distances increase slowly. It also indicates that the effectiveness of \texttt{Decaf} does not drop off a cliff as the number of neurons increases.
\subsection{Without Null Classes Removal}
\label{sec:without_removal}
\begin{figure}[h]
	\centering
	\includegraphics[width=0.75\linewidth]{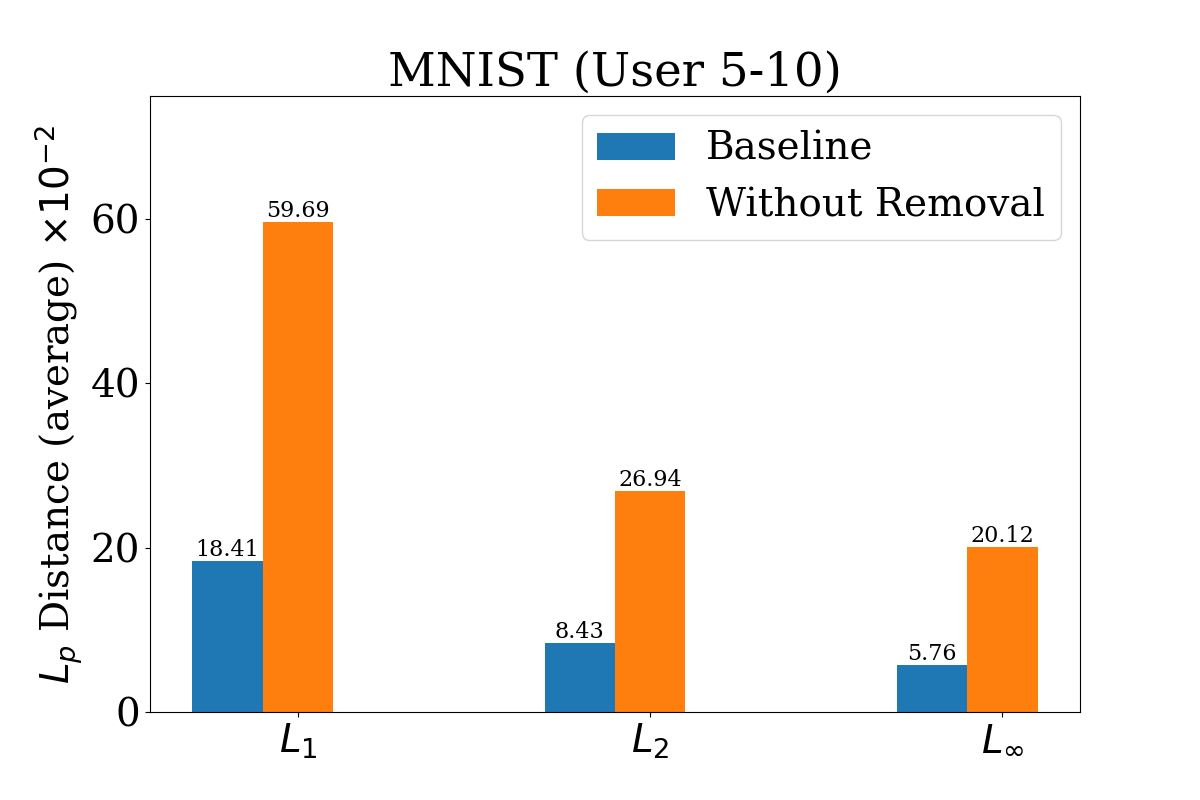}
	\caption{\texttt{Decaf} without null classes removal can not predict well when users' data have null classes.}
    \label{Fig_without_removal}
\end{figure}

The step of Null Classes Removal ensures that \texttt{Decaf} can obtain 100\% accuracy when inferring users' null classes, thus enabling the rest of the remaining classes' decomposition to be more accurately performed. We now perform and compare \texttt{Decaf} with/without the step of Null Classes Removal on MNIST (detailed in \autoref{tab:without_removal}). We can see that \texttt{Decaf} can still accurately attack the first four users with all classes without Null Classes Removal because the step does not affect users who \textit{do not have null classes}. However, the $L_p$ distance of users 5-10 who do own null classes, as shown in \autoref{Fig_without_removal}, is times larger than the baseline attack with the Null Classes Removal step applied.
\subsection{Without Calibrator}
\label{sec:calibrator}
\begin{figure}[h]
	\centering
	\includegraphics[width=0.75\linewidth]{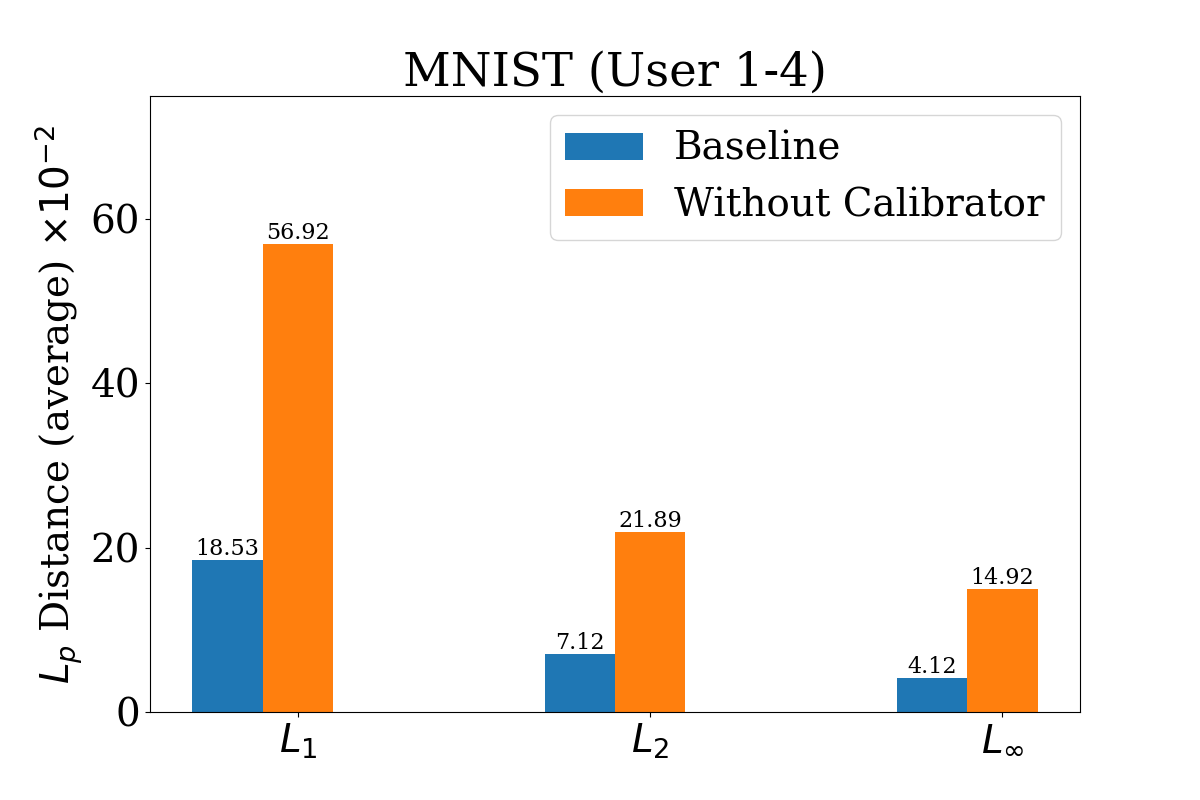}
	\caption{\texttt{Decaf} without the calibrator can not predict well when users' data is distributed evenly.}
    \label{Fig_without_calibrator}
\end{figure}

In this section, we validate the significance of the calibrator $g_u^{t}$ mentioned in \autoref{RCD}, as shown in \autoref{Fig_without_calibrator} (numerically detailed in \autoref{tab:without_calibrator}). In contrast to \autoref{sec:without_removal}, \texttt{Decaf} can not even work well on users 1-4 who have all classes. The $L_p$ distance is also times larger than the baseline attack. From \autoref{tab:without_calibrator}, many non-null classes are wrongly inferred as null (not at the step of Null Classes Removal, but at Remaining Classes Decomposition). Nonetheless, the accuracy of inferring real null classes can still be maintained at 100\%. A potential reason is that users train local models with their data containing all non-classes, and their gradient changes will show a balance between each class rather than the gradient bases constructed by the auxiliary per-class data samples. Thus, there will be extra accuracy losses if we use $g_c^{t}$ (the single-class gradient bases) to perform the step of remaining non-classes decomposition without the calibrator $g_u^{t}$. In other words, the calibrator $g_u^{t}$ can simulate users' mixed data that consists of all classes to a large extent as a reference.

\section{Discussion}
\begin{figure*}[h]
  \centering
  \includegraphics[width=\linewidth]{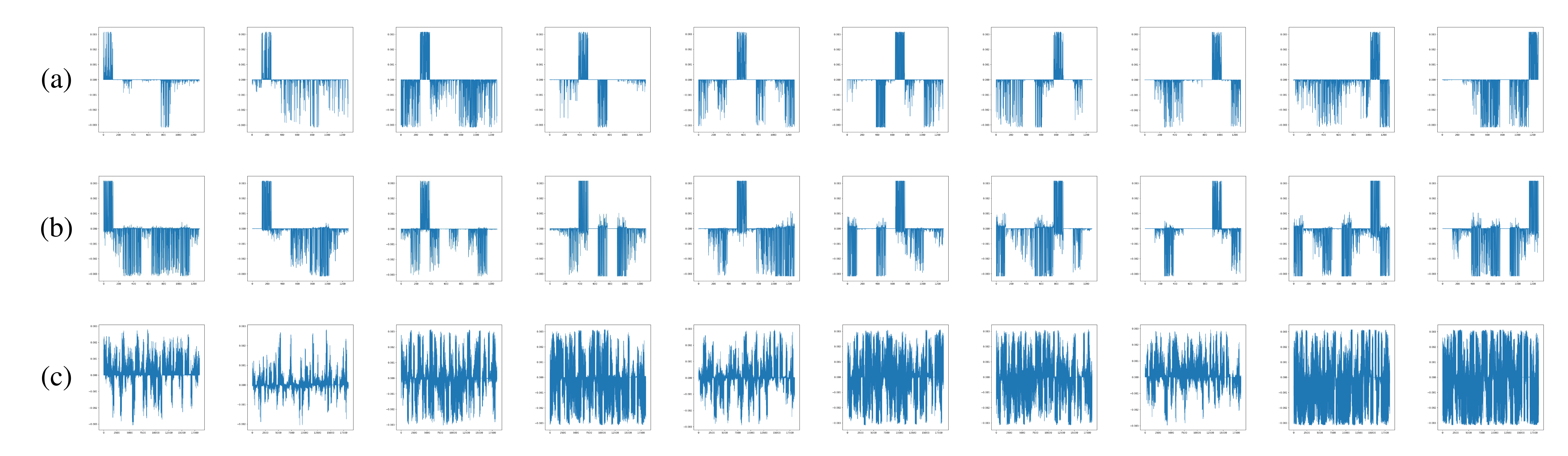}
  \caption{Figure (a) is the gradient changes of the last fully connected layer with the \textsf{relu} activation function, Figure (b) is the gradient changes of the last fully connected layer with the \textsf{leakyrelu} activation function, and Figure (c) is the gradient changes of the last convolution layer with the \textsf{relu} activation function. The horizontal coordinate is the neuron position, and the vertical coordinate is the gradient changes. The above plots are the same model trained using 0-9 single-class data, respectively.}
  \label{Fig_regularity}
\end{figure*}
\begin{figure}[h]
  \centering
  \includegraphics[width=0.90\linewidth]{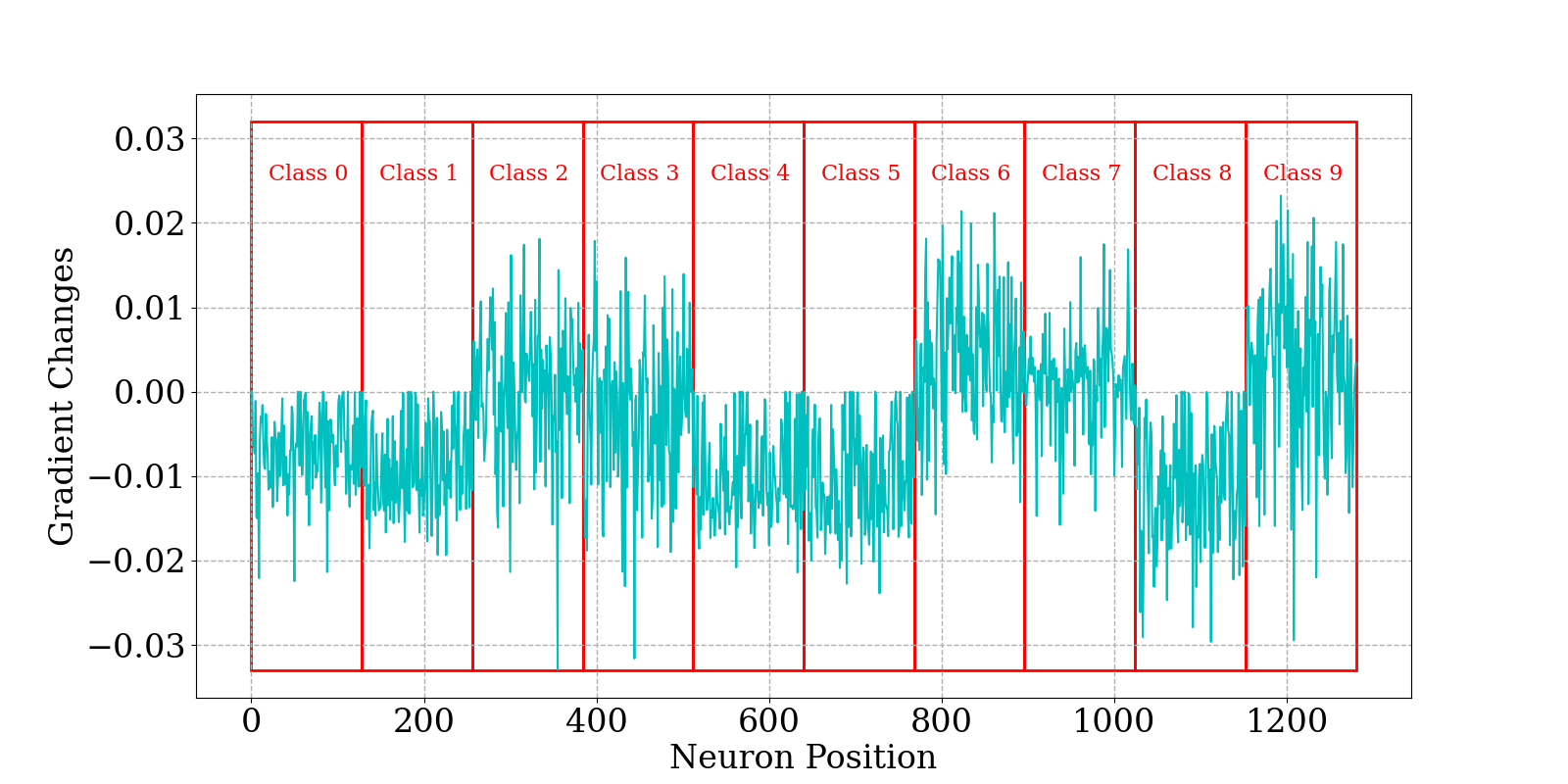}
  \caption{Gradient changes of the fully connected layer. Null classes exhibit non positive gradient changes on the corresponding neuron position.}
  \label{Fig_fully_connected_layer}
\end{figure}
\subsection{Target Layer Selection}
\label{sec:target_layer}
As proved in \autoref{sec:null_classes}, we find that a specific group of neurons in the last fully connected layer of a classification model is specifically responsible for a class. Furthermore, we find that the gradient changes will not exhibit positive values given that a class is null in the victim's local data. Here, we will provide more specific experimental proof.

We validate this in \autoref{Fig_regularity}(a), the gradient changes of the last fully connected layer (i.e., the classification layer) are obtained by retraining an already-trained model with per-class samples, respectively. The dataset is MNIST, and the model architecture is shown in \autoref{tab:model_structure}. The \textsf{relu} activation function increases the nonlinearity between the neural network layers. We can find that the corresponding group of neurons has positive gradient changes when we train the model only with data from a single class, which is consistent with our proof in \autoref{sec:null_classes}. The \textsf{relu} (same with \textsf{sigmod} etc.) activation function trims all negative values from the previous layer's output, making the model vulnerable to our attack. We indicate that activation functions like \textsf{leakyrelu} and others that significantly rectify the negative values also render the same privacy leakage problem, shown in  \autoref{Fig_regularity}(b). In this case, we can improve the accuracy of our attack by increasing the gradient change threshold for determining the user's null classes, e.g., from 0 to 0.001---note that the attacker/server knows the activation function being used in FL.

We have also performed gradient change extraction on other layers, such as the last convolution layer of the model. As shown in \autoref{Fig_regularity}(c), we find that the gradient changes of neurons in this layer do not exhibit a (visually) notable pattern when the model is retrained with per-class samples. The reason is that the last fully connected layer is mainly responsible for classifying the samples into each class so that they will be more related to the data distribution characteristics, as we have proved before. However, the last convolution layer of the model is mainly responsible for feature extraction, i.e., the parameters of the last convolution layer of the model will be more related to general data features of input samples~\cite{26DBLP:conf/iccv/SelvarajuCDVPB17}. 
\autoref{Fig_fully_connected_layer} shows the gradient changes of the last fully connected layer trained with five null classes, namely digits 0, 1, 4, 5, and 8. It is obvious that there are no positive changes in the gradient of neurons at the corresponding neuron positions of these null classes.
\begin{figure}[h]
	\centering
	\includegraphics[width=\linewidth]{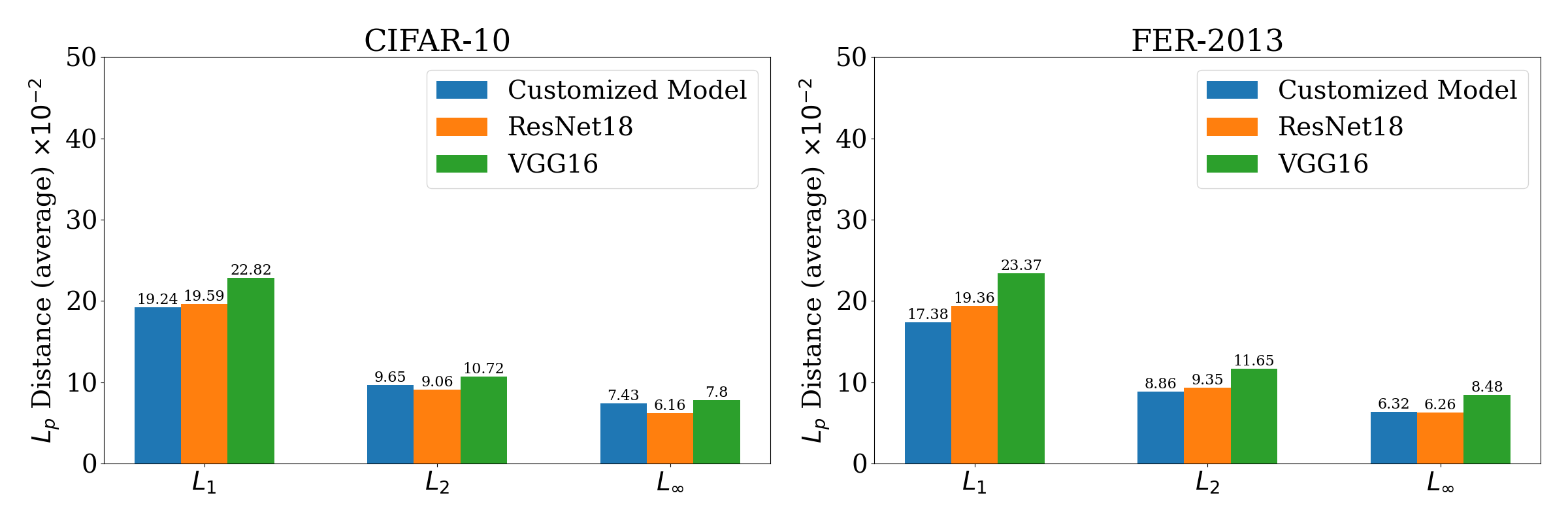}
	\caption{$L_p$ distances under different model architectures (customized CNN model, ResNet18 and VGG16) on CIFAR-10 and FER-2013.}
    \label{Fig_complex_model}
\end{figure}

\begin{table*}[h]
\centering
\caption{The results on ResNet18 (FER-2013).}
\resizebox{0.88\linewidth}{!}{
\begin{tabular}{|c|ccccccc|cllcll|}
\hline
                             & \multicolumn{7}{c|}{}                                                                                                                                                                                                                                                                                                                                       & \multicolumn{6}{c|}{\begin{tabular}[c]{@{}c@{}}\textbf{$L_1$/$L_2$/$L_\infty$($\times 10^{-2}$)}\end{tabular}}                                                                                           \\ \cline{9-14} 
\multirow{-2}{*}{\textbf{User}}       & \multicolumn{7}{c|}{\multirow{-2}{*}{\textbf{Data Composition (\%)}}}                                                                                                                                                                                                                                                                                       & \multicolumn{3}{c|}{Decaf}                                     & \multicolumn{3}{c|}{Random Guess}                \\ \hline
                             & \multicolumn{1}{c|}{14.29}                         & \multicolumn{1}{c|}{14.29}                         & \multicolumn{1}{c|}{14.29}                         & \multicolumn{1}{c|}{14.29}                         & \multicolumn{1}{c|}{14.29}                         & \multicolumn{1}{c|}{14.29}                         & 14.29                         & \multicolumn{3}{c|}{}                                                   & \multicolumn{3}{c|}{}                                     \\ \cline{2-8}
\multirow{-2}{*}{\textbf{1}} & \multicolumn{1}{c|}{\cellcolor[HTML]{C0C0C0}19.4}  & \multicolumn{1}{c|}{\cellcolor[HTML]{C0C0C0}18.03} & \multicolumn{1}{c|}{\cellcolor[HTML]{C0C0C0}10.9}  & \multicolumn{1}{c|}{\cellcolor[HTML]{C0C0C0}13.53} & \multicolumn{1}{c|}{\cellcolor[HTML]{C0C0C0}16.34} & \multicolumn{1}{c|}{\cellcolor[HTML]{C0C0C0}10.9}  & \cellcolor[HTML]{C0C0C0}10.9  & \multicolumn{3}{c|}{\multirow{-2}{*}{{\ul{21.83/8.91/5.11}}}}   & \multicolumn{3}{c|}{\multirow{-2}{*}{30.29/13.07/8.37}}   \\ \hline
                             & \multicolumn{1}{c|}{19.23}                         & \multicolumn{1}{c|}{5.77}                          & \multicolumn{1}{c|}{15.38}                         & \multicolumn{1}{c|}{3.85}                          & \multicolumn{1}{c|}{26.92}                         & \multicolumn{1}{c|}{13.46}                         & 15.38                         & \multicolumn{3}{c|}{}                                                   & \multicolumn{3}{c|}{}                                     \\ \cline{2-8}
\multirow{-2}{*}{\textbf{2}} & \multicolumn{1}{c|}{\cellcolor[HTML]{C0C0C0}19.73} & \multicolumn{1}{c|}{\cellcolor[HTML]{C0C0C0}16.16} & \multicolumn{1}{c|}{\cellcolor[HTML]{C0C0C0}19.84} & \multicolumn{1}{c|}{\cellcolor[HTML]{C0C0C0}8.74}  & \multicolumn{1}{c|}{\cellcolor[HTML]{C0C0C0}17.74} & \multicolumn{1}{c|}{\cellcolor[HTML]{C0C0C0}9.05}  & \cellcolor[HTML]{C0C0C0}8.74  & \multicolumn{3}{c|}{\multirow{-2}{*}{{\ul{40.47/17.32/10.39}}}} & \multicolumn{3}{c|}{\multirow{-2}{*}{59.25/26.2/14.44}}   \\ \hline
                             & \multicolumn{1}{c|}{5.17}                          & \multicolumn{1}{c|}{7.76}                          & \multicolumn{1}{c|}{13.79}                         & \multicolumn{1}{c|}{24.14}                         & \multicolumn{1}{c|}{12.07}                         & \multicolumn{1}{c|}{19.83}                         & 17.24                         & \multicolumn{3}{c|}{}                                                   & \multicolumn{3}{c|}{}                                     \\ \cline{2-8}
\multirow{-2}{*}{\textbf{3}} & \multicolumn{1}{c|}{\cellcolor[HTML]{C0C0C0}10.56} & \multicolumn{1}{c|}{\cellcolor[HTML]{C0C0C0}12.63} & \multicolumn{1}{c|}{\cellcolor[HTML]{C0C0C0}10.56} & \multicolumn{1}{c|}{\cellcolor[HTML]{C0C0C0}23.61} & \multicolumn{1}{c|}{\cellcolor[HTML]{C0C0C0}10.56} & \multicolumn{1}{c|}{\cellcolor[HTML]{C0C0C0}13.66} & \cellcolor[HTML]{C0C0C0}18.41 & \multicolumn{3}{c|}{\multirow{-2}{*}{{\ul{22.87/10.26/6.17}}}}  & \multicolumn{3}{c|}{\multirow{-2}{*}{80.1/36.61/23.81}}   \\ \hline
                             & \multicolumn{1}{c|}{29.2}                          & \multicolumn{1}{c|}{-}                             & \multicolumn{1}{c|}{7.3}                           & \multicolumn{1}{c|}{21.9}                          & \multicolumn{1}{c|}{8.76}                          & \multicolumn{1}{c|}{10.95}                         & 21.9                          & \multicolumn{3}{c|}{}                                                   & \multicolumn{3}{c|}{}                                     \\ \cline{2-8}
\multirow{-2}{*}{\textbf{4}} & \multicolumn{1}{c|}{\cellcolor[HTML]{C0C0C0}29.71} & \multicolumn{1}{c|}{\cellcolor[HTML]{C0C0C0}-}     & \multicolumn{1}{c|}{\cellcolor[HTML]{C0C0C0}10.95} & \multicolumn{1}{c|}{\cellcolor[HTML]{C0C0C0}16.01} & \multicolumn{1}{c|}{\cellcolor[HTML]{C0C0C0}10.95} & \multicolumn{1}{c|}{\cellcolor[HTML]{C0C0C0}10.95} & \cellcolor[HTML]{C0C0C0}21.41 & \multicolumn{3}{c|}{\multirow{-2}{*}{{\ul{12.73/7.30/5.89}}}}   & \multicolumn{3}{c|}{\multirow{-2}{*}{45.41/20.43/12.15}}  \\ \hline
                             & \multicolumn{1}{c|}{-}                             & \multicolumn{1}{c|}{8.04}                          & \multicolumn{1}{c|}{33.93}                         & \multicolumn{1}{c|}{-}                             & \multicolumn{1}{c|}{26.79}                         & \multicolumn{1}{c|}{31.25}                         & -                             & \multicolumn{3}{c|}{}                                                   & \multicolumn{3}{c|}{}                                     \\ \cline{2-8}
\multirow{-2}{*}{\textbf{5}} & \multicolumn{1}{c|}{\cellcolor[HTML]{C0C0C0}-}     & \multicolumn{1}{c|}{\cellcolor[HTML]{C0C0C0}19.07} & \multicolumn{1}{c|}{\cellcolor[HTML]{C0C0C0}24.69} & \multicolumn{1}{c|}{\cellcolor[HTML]{C0C0C0}-}     & \multicolumn{1}{c|}{\cellcolor[HTML]{C0C0C0}29.3}  & \multicolumn{1}{c|}{\cellcolor[HTML]{C0C0C0}26.94} & \cellcolor[HTML]{C0C0C0}-     & \multicolumn{3}{c|}{\multirow{-2}{*}{{\ul{27.09/15.23/11.03}}}} & \multicolumn{3}{c|}{\multirow{-2}{*}{85.95/35.82/19.34}}  \\ \hline
                             & \multicolumn{1}{c|}{27.4}                          & \multicolumn{1}{c|}{-}                             & \multicolumn{1}{c|}{47.95}                         & \multicolumn{1}{c|}{-}                             & \multicolumn{1}{c|}{24.66}                         & \multicolumn{1}{c|}{-}                             & -                             & \multicolumn{3}{c|}{}                                                   & \multicolumn{3}{c|}{}                                     \\ \cline{2-8}
\multirow{-2}{*}{\textbf{6}} & \multicolumn{1}{c|}{\cellcolor[HTML]{C0C0C0}24.35} & \multicolumn{1}{c|}{\cellcolor[HTML]{C0C0C0}-}     & \multicolumn{1}{c|}{\cellcolor[HTML]{C0C0C0}45.73} & \multicolumn{1}{c|}{\cellcolor[HTML]{C0C0C0}-}     & \multicolumn{1}{c|}{\cellcolor[HTML]{C0C0C0}29.92} & \multicolumn{1}{c|}{\cellcolor[HTML]{C0C0C0}-}     & \cellcolor[HTML]{C0C0C0}-     & \multicolumn{3}{c|}{\multirow{-2}{*}{{\ul{10.53/6.47/5.26}}}}   & \multicolumn{3}{c|}{\multirow{-2}{*}{72.69/32.05/17.52}}  \\ \hline
                             & \multicolumn{1}{c|}{-}                             & \multicolumn{1}{c|}{-}                             & \multicolumn{1}{c|}{-}                             & \multicolumn{1}{c|}{-}                             & \multicolumn{1}{c|}{-}                             & \multicolumn{1}{c|}{-}                             & 100                           & \multicolumn{3}{c|}{}                                                   & \multicolumn{3}{c|}{}                                     \\ \cline{2-8}
\multirow{-2}{*}{\textbf{7}} & \multicolumn{1}{c|}{\cellcolor[HTML]{C0C0C0}-}     & \multicolumn{1}{c|}{\cellcolor[HTML]{C0C0C0}-}     & \multicolumn{1}{c|}{\cellcolor[HTML]{C0C0C0}-}     & \multicolumn{1}{c|}{\cellcolor[HTML]{C0C0C0}-}     & \multicolumn{1}{c|}{\cellcolor[HTML]{C0C0C0}-}     & \multicolumn{1}{c|}{\cellcolor[HTML]{C0C0C0}-}     & \cellcolor[HTML]{C0C0C0}100   & \multicolumn{3}{c|}{\multirow{-2}{*}{{\ul{0/0/0}}}}             & \multicolumn{3}{c|}{\multirow{-2}{*}{158.64/87.94/79.32}} \\ \hline
\end{tabular}
}
\label{tab:fer2013_resnet18}
\end{table*}

\begin{table*}[h]
\centering
\caption{The results on VGG16 (FER-2013).}
\resizebox{0.88\linewidth}{!}{
\begin{tabular}{|c|ccccccc|cllcll|}
\hline
                             & \multicolumn{7}{c|}{}                                                                                                                                                                                                                                                                                                                                       & \multicolumn{6}{c|}{\begin{tabular}[c]{@{}c@{}}\textbf{$L_1$/$L_2$/$L_\infty$($\times 10^{-2}$)}\end{tabular}}                                                                                           \\ \cline{9-14} 
\multirow{-2}{*}{\textbf{User}}       & \multicolumn{7}{c|}{\multirow{-2}{*}{\textbf{Data Composition (\%)}}}                                                                                                                                                                                                                                                                                       & \multicolumn{3}{c|}{Decaf}                                     & \multicolumn{3}{c|}{Random Guess}                \\ \hline
                             & \multicolumn{1}{c|}{14.29}                         & \multicolumn{1}{c|}{14.29}                         & \multicolumn{1}{c|}{14.29}                         & \multicolumn{1}{c|}{14.29}                         & \multicolumn{1}{c|}{14.29}                         & \multicolumn{1}{c|}{14.29}                         & 14.29                         & \multicolumn{3}{c|}{}                                                   & \multicolumn{3}{c|}{}                                     \\ \cline{2-8}
\multirow{-2}{*}{\textbf{1}} & \multicolumn{1}{c|}{\cellcolor[HTML]{C0C0C0}17.03} & \multicolumn{1}{c|}{\cellcolor[HTML]{C0C0C0}17.93} & \multicolumn{1}{c|}{\cellcolor[HTML]{C0C0C0}11.31} & \multicolumn{1}{c|}{\cellcolor[HTML]{C0C0C0}15.74} & \multicolumn{1}{c|}{\cellcolor[HTML]{C0C0C0}12.34} & \multicolumn{1}{c|}{\cellcolor[HTML]{C0C0C0}14.35} & \cellcolor[HTML]{C0C0C0}11.31 & \multicolumn{3}{c|}{\multirow{-2}{*}{{\ul{15.8/6.67/3.64}}}}    & \multicolumn{3}{c|}{\multirow{-2}{*}{30.29/13.07/8.37}}   \\ \hline
                             & \multicolumn{1}{c|}{19.23}                         & \multicolumn{1}{c|}{5.77}                          & \multicolumn{1}{c|}{15.38}                         & \multicolumn{1}{c|}{3.85}                          & \multicolumn{1}{c|}{26.92}                         & \multicolumn{1}{c|}{13.46}                         & 15.38                         & \multicolumn{3}{c|}{}                                                   & \multicolumn{3}{c|}{}                                     \\ \cline{2-8}
\multirow{-2}{*}{\textbf{2}} & \multicolumn{1}{c|}{\cellcolor[HTML]{C0C0C0}25.45} & \multicolumn{1}{c|}{\cellcolor[HTML]{C0C0C0}22}    & \multicolumn{1}{c|}{\cellcolor[HTML]{C0C0C0}11.24} & \multicolumn{1}{c|}{\cellcolor[HTML]{C0C0C0}4.38}  & \multicolumn{1}{c|}{\cellcolor[HTML]{C0C0C0}15.39} & \multicolumn{1}{c|}{\cellcolor[HTML]{C0C0C0}8.59}  & \cellcolor[HTML]{C0C0C0}12.94 & \multicolumn{3}{c|}{\multirow{-2}{*}{{\ul{45.96/21.96/16.23}}}} & \multicolumn{3}{c|}{\multirow{-2}{*}{59.25/26.2/14.44}}   \\ \hline
                             & \multicolumn{1}{c|}{5.17}                          & \multicolumn{1}{c|}{7.76}                          & \multicolumn{1}{c|}{13.79}                         & \multicolumn{1}{c|}{24.14}                         & \multicolumn{1}{c|}{12.07}                         & \multicolumn{1}{c|}{19.83}                         & 17.24                         & \multicolumn{3}{c|}{}                                                   & \multicolumn{3}{c|}{}                                     \\ \cline{2-8}
\multirow{-2}{*}{\textbf{3}} & \multicolumn{1}{c|}{\cellcolor[HTML]{C0C0C0}9.04}  & \multicolumn{1}{c|}{\cellcolor[HTML]{C0C0C0}14.84} & \multicolumn{1}{c|}{\cellcolor[HTML]{C0C0C0}9.5}   & \multicolumn{1}{c|}{\cellcolor[HTML]{C0C0C0}25.34} & \multicolumn{1}{c|}{\cellcolor[HTML]{C0C0C0}9.19}  & \multicolumn{1}{c|}{\cellcolor[HTML]{C0C0C0}18.26} & \cellcolor[HTML]{C0C0C0}13.83 & \multicolumn{3}{c|}{\multirow{-2}{*}{{\ul{24.3/10.36/7.08}}}}   & \multicolumn{3}{c|}{\multirow{-2}{*}{80.1/36.61/23.81}}   \\ \hline
                             & \multicolumn{1}{c|}{29.2}                          & \multicolumn{1}{c|}{-}                             & \multicolumn{1}{c|}{7.3}                           & \multicolumn{1}{c|}{21.9}                          & \multicolumn{1}{c|}{8.76}                          & \multicolumn{1}{c|}{10.95}                         & 21.9                          & \multicolumn{3}{c|}{}                                                   & \multicolumn{3}{c|}{}                                     \\ \cline{2-8}
\multirow{-2}{*}{\textbf{4}} & \multicolumn{1}{c|}{\cellcolor[HTML]{C0C0C0}20.55} & \multicolumn{1}{c|}{\cellcolor[HTML]{C0C0C0}-}     & \multicolumn{1}{c|}{\cellcolor[HTML]{C0C0C0}13.86} & \multicolumn{1}{c|}{\cellcolor[HTML]{C0C0C0}20.1}  & \multicolumn{1}{c|}{\cellcolor[HTML]{C0C0C0}12.63} & \multicolumn{1}{c|}{\cellcolor[HTML]{C0C0C0}15.53} & \cellcolor[HTML]{C0C0C0}17.33 & \multicolumn{3}{c|}{\multirow{-2}{*}{{\ul{30.03/13.34/8.65}}}}  & \multicolumn{3}{c|}{\multirow{-2}{*}{45.41/20.43/12.15}}  \\ \hline
                             & \multicolumn{1}{c|}{-}                             & \multicolumn{1}{c|}{8.04}                          & \multicolumn{1}{c|}{33.93}                         & \multicolumn{1}{c|}{-}                             & \multicolumn{1}{c|}{26.79}                         & \multicolumn{1}{c|}{31.25}                         & -                             & \multicolumn{3}{c|}{}                                                   & \multicolumn{3}{c|}{}                                     \\ \cline{2-8}
\multirow{-2}{*}{\textbf{5}} & \multicolumn{1}{c|}{\cellcolor[HTML]{C0C0C0}-}     & \multicolumn{1}{c|}{\cellcolor[HTML]{C0C0C0}20.34} & \multicolumn{1}{c|}{\cellcolor[HTML]{C0C0C0}26.52} & \multicolumn{1}{c|}{\cellcolor[HTML]{C0C0C0}-}     & \multicolumn{1}{c|}{\cellcolor[HTML]{C0C0C0}23}    & \multicolumn{1}{c|}{\cellcolor[HTML]{C0C0C0}30.15} & \cellcolor[HTML]{C0C0C0}-     & \multicolumn{3}{c|}{\multirow{-2}{*}{{\ul{24.6/14.89/12.3}}}}   & \multicolumn{3}{c|}{\multirow{-2}{*}{85.95/35.82/19.34}}  \\ \hline
                             & \multicolumn{1}{c|}{27.4}                          & \multicolumn{1}{c|}{-}                             & \multicolumn{1}{c|}{47.95}                         & \multicolumn{1}{c|}{-}                             & \multicolumn{1}{c|}{24.66}                         & \multicolumn{1}{c|}{-}                             & -                             & \multicolumn{3}{c|}{}                                                   & \multicolumn{3}{c|}{}                                     \\ \cline{2-8}
\multirow{-2}{*}{\textbf{6}} & \multicolumn{1}{c|}{\cellcolor[HTML]{C0C0C0}35.29} & \multicolumn{1}{c|}{\cellcolor[HTML]{C0C0C0}-}     & \multicolumn{1}{c|}{\cellcolor[HTML]{C0C0C0}36.49} & \multicolumn{1}{c|}{\cellcolor[HTML]{C0C0C0}-}     & \multicolumn{1}{c|}{\cellcolor[HTML]{C0C0C0}28.22} & \multicolumn{1}{c|}{\cellcolor[HTML]{C0C0C0}-}     & \cellcolor[HTML]{C0C0C0}-     & \multicolumn{3}{c|}{\multirow{-2}{*}{{\ul{22.91/14.36/11.46}}}} & \multicolumn{3}{c|}{\multirow{-2}{*}{72.69/32.05/17.52}}  \\ \hline
                             & \multicolumn{1}{c|}{-}                             & \multicolumn{1}{c|}{-}                             & \multicolumn{1}{c|}{-}                             & \multicolumn{1}{c|}{-}                             & \multicolumn{1}{c|}{-}                             & \multicolumn{1}{c|}{-}                             & 100                           & \multicolumn{3}{c|}{}                                                   & \multicolumn{3}{c|}{}                                     \\ \cline{2-8}
\multirow{-2}{*}{\textbf{7}} & \multicolumn{1}{c|}{\cellcolor[HTML]{C0C0C0}-}     & \multicolumn{1}{c|}{\cellcolor[HTML]{C0C0C0}-}     & \multicolumn{1}{c|}{\cellcolor[HTML]{C0C0C0}-}     & \multicolumn{1}{c|}{\cellcolor[HTML]{C0C0C0}-}     & \multicolumn{1}{c|}{\cellcolor[HTML]{C0C0C0}-}     & \multicolumn{1}{c|}{\cellcolor[HTML]{C0C0C0}-}     & \cellcolor[HTML]{C0C0C0}100   & \multicolumn{3}{c|}{\multirow{-2}{*}{{\ul{0/0/0}}}}             & \multicolumn{3}{c|}{\multirow{-2}{*}{158.64/87.94/79.32}} \\ \hline
\end{tabular}
}
\label{tab:fer2013_vgg16}
\end{table*}

\subsection{Model Complexity}
\label{sec:complex_model}
We now evaluate \texttt{Decaf} on standard ResNet18 and VGG16---more complicated models compared to those customized models used in previous experiments, using the CIFAR-10 and FER-2013 datasets with customized CNN models---relatively simple. The experimental setups are consistent with the setups for customized models used in previous experiments. 
As shown in \autoref{Fig_complex_model}, it depicts the computed average $L_p$ distance of all 10 victims with diverse data distribution (detailed in \autoref{tab:fer2013_resnet18}-\autoref{tab:fer2013_vgg16} and \autoref{tab:cifar10_resnet18}-\autoref{tab:cifar10_vgg16}). 

On the CIFAR-10 and FER-2013, the $L_1$ distance increases with the model complexity. It is potentially because the VGG16 has the maximum number of model parameters, e.g., its last fully connected layer has the highest number of neurons of $1000 \times N$ ($N$ is the number of classes in the classification task). It can render higher accuracy loss of remaining classes decomposition compared to ResNet18 that has $512 \times N$ neurons and the customized CNN model that has $128 \times N$ neurons in the last fully connected layer. More detailed albation studies on the relationship between the \texttt{Decaf} accuracy and number of neurons in the last fully connected layers can be found in \autoref{sec:layernum}. Nonetheless, the $L_1$ distance is still around 0.2, and the $L_\infty$ distance is always less than 0.1. It means that \texttt{Decaf} can still work well, although the model complexity becomes complicated.
\begin{table}[h]
\centering
\caption{Model accuracy with different defenses.}
\resizebox{0.6\linewidth}{!}{
\begin{tabular}{|cc|c|}
\hline
\multicolumn{2}{|c|}{\textbf{Defense}}                                                                              & \textbf{Model accuracy} \\ \hline
\multicolumn{2}{|c|}{Baseline}                                                                                 & 98.79\%        \\ \hline
\multicolumn{2}{|c|}{Dropout}                                                                              & 98.6\%         \\ \hline
\multicolumn{1}{|c|}{\multirow{5}{*}{\begin{tabular}[c]{@{}c@{}}Differential\\ Privacy\end{tabular}}} & $\epsilon_1$ & 93.11\%        \\ \cline{2-3} 
\multicolumn{1}{|c|}{}                                                                                & $\epsilon_2$ & 92.16\%        \\ \cline{2-3} 
\multicolumn{1}{|c|}{}                                                                                & $\epsilon_3$ & 87.09\%        \\ \cline{2-3} 
\multicolumn{1}{|c|}{}                                                                                & $\epsilon_4$ & 79.63\%        \\ \cline{2-3} 
\multicolumn{1}{|c|}{}                                                                                & $\epsilon_5$ & 58.9\%         \\ \hline
\end{tabular}
}
\label{tab:acc_with_defenses}
\end{table}
\begin{figure}[h]
	\centering
	\includegraphics[width=0.85\linewidth]{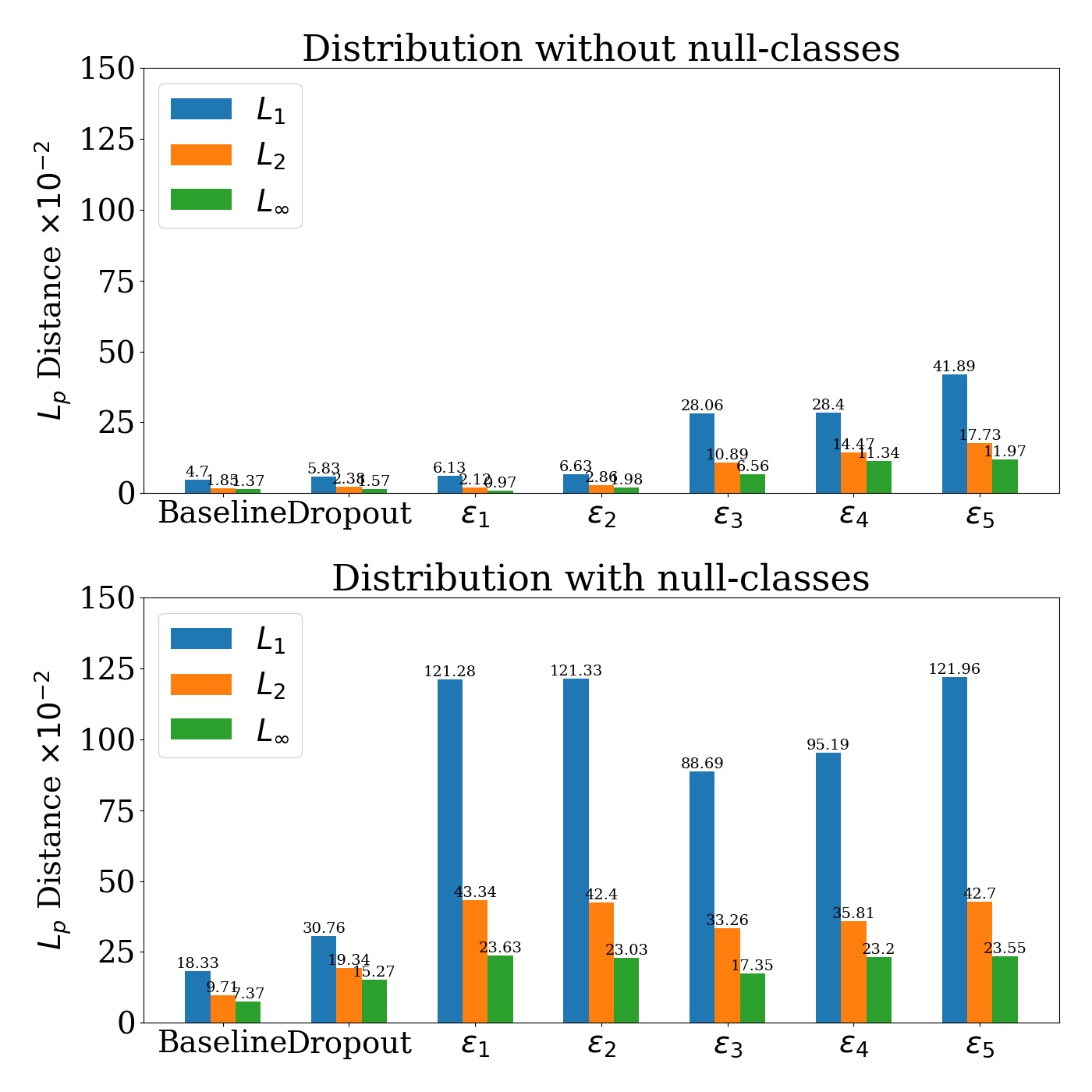}
	\caption{$L_p$ distance when defenses of dropout and differential privacy with varying budget $\epsilon$ are applied. Data distributions are set to be with and without null-classes.}
    \label{Fig_defenses}
\end{figure}
\subsection{Countermeasure}
We also perform \texttt{Decaf} when common defenses, such as dropout \cite{41DBLP:journals/jmlr/SrivastavaHKSS14} and differential privacy \cite{42DBLP:conf/icalp/Dwork06, 43DBLP:conf/ccs/AbadiCGMMT016} are applied. In differential privacy, we set the privacy budget $\epsilon_1 = 355.07$, $\epsilon_2 = 45.69$, $\epsilon_3 = 1.03$, $\epsilon_4 = 0.12$, $\epsilon_5 = 0.03$ under the constraints of noise multiplier = 0.1, 0.25, 1, 4 and 16 respectively, and $\delta = 10^{-5}$.

\autoref{Fig_defenses} shows attack results when dropout and differential privacy are applied. Among them, the dropout defense does not have obvious effects on \texttt{Decaf}. It is almost the same as the baseline attack without any defense. On the contrary, differential privacy notably decreases \texttt{Decaf} performance. When $\epsilon$ is large, like $\epsilon_1=355.07$ and $\epsilon_2=45.69$, \texttt{Decaf} can still predict it accurately if the user's data is distributed evenly (without null-classes). However, \texttt{Decaf} can not work well if the user's data is distributed unevenly (with null-classes) whenever the $\epsilon$ is large or small. Because differential privacy directly injects noise to users' model gradients, significantly affecting the determination of users' null classes in the first attack stage of \texttt{Decaf}. 

However, the utility of the global witnesses significant deterioration when more noises are leveraged to prevent \texttt{Decaf}. \autoref{tab:acc_with_defenses} shows the global model accuracy when different defenses are applied. Differential privacy protects user privacy with a trade-off of hurting utility, which is not applicable in practice. Besides, encrypt defenses like homomorphic encryption \cite{44DBLP:journals/tifs/PhongAHWM18} and secure multi-party computation \cite{45DBLP:conf/ccs/BonawitzIKMMPRS17} also can resist \texttt{Decaf}. However, these crypto-enabled defenses often neither adversely affect the accuracy or efficiency of the default FL to a greater or lesser extent. Thus, FL systems using plaintext transmission are still relatively common, and our attack is still well worth studying. Moreover, our research is more focused on some vulnerabilities in deep learning models themselves that will cause serious privacy leakage, which is aligned with these related privacy leakage investigations \cite{1DBLP:conf/ndss/ZhouGFCD0X023, 9DBLP:conf/sp/ShokriSSS17, 31DBLP:conf/sp/NasrSH19, 17DBLP:conf/ccs/GanjuWYGB18, 18DBLP:conf/sp/MelisSCS19, 4DBLP:conf/ccs/HitajAP17}.

\section{Conclusion}
This work has revealed an innovative type of FL privacy inference attack, \texttt{Decaf}, which can stealthily decompose per-class proportion owned by a victim FL user. The foundational insight is that local model neurons memorize quantifiable gradient changes in relation to local data distribution. In addition to formal proof of null-classes identification, expensive experiments on five benchmark datasets and various model architectures affirmed the accuracy, stealthiness, and stability of the \texttt{Decaf}.





%
\bibliographystyle{IEEEtranS}
\bibliography{bibtex/bib/Decaf}



\appendices
\section{Detailed Results of \texttt{Decaf}}
\label{detailed_results}
\autoref{tab:cifar10_resnet18}-\autoref{tab:without_calibrator} are the detailed results of \texttt{Decaf} on complex model structure, without null classes removal, and without calibrator/$g^t_u$.

\begin{table*}[h]
\centering
\caption{The results on ResNet18 (CIFAR-10).}
\resizebox{\linewidth}{!}{
\begin{tabular}{|c|cccccccccc|cllcll|}
\hline
                              & \multicolumn{10}{c|}{}                                                                                                                                                                                                                                                                                                                                                                                                                                                                                                     & \multicolumn{6}{c|}{\begin{tabular}[c]{@{}c@{}}\textbf{$L_1$/$L_2$/$L_\infty$($\times 10^{-2}$)}\end{tabular}}                                                                                            \\ \cline{12-17} 
\multirow{-2}{*}{\textbf{User}}        & \multicolumn{10}{c|}{\multirow{-2}{*}{\textbf{Data Composition (\%)}}}                                                                                                                                                                                                                                                                                                                                                                                                                                                     & \multicolumn{3}{c|}{Decaf}                                     & \multicolumn{3}{c|}{Random Guess}                 \\ \hline
                              & \multicolumn{1}{c|}{10}                            & \multicolumn{1}{c|}{10}                            & \multicolumn{1}{c|}{10}                            & \multicolumn{1}{c|}{10}                            & \multicolumn{1}{c|}{10}                            & \multicolumn{1}{c|}{10}                            & \multicolumn{1}{c|}{10}                            & \multicolumn{1}{c|}{10}                            & \multicolumn{1}{c|}{10}                            & 10                            & \multicolumn{3}{c|}{}                                                   & \multicolumn{3}{c|}{}                                      \\ \cline{2-11}
\multirow{-2}{*}{\textbf{1}}  & \multicolumn{1}{c|}{\cellcolor[HTML]{C0C0C0}10.57} & \multicolumn{1}{c|}{\cellcolor[HTML]{C0C0C0}10.98} & \multicolumn{1}{c|}{\cellcolor[HTML]{C0C0C0}13.38} & \multicolumn{1}{c|}{\cellcolor[HTML]{C0C0C0}9.34}  & \multicolumn{1}{c|}{\cellcolor[HTML]{C0C0C0}9}     & \multicolumn{1}{c|}{\cellcolor[HTML]{C0C0C0}10.16} & \multicolumn{1}{c|}{\cellcolor[HTML]{C0C0C0}9.46}  & \multicolumn{1}{c|}{\cellcolor[HTML]{C0C0C0}11.21} & \multicolumn{1}{c|}{\cellcolor[HTML]{C0C0C0}8.21}  & \cellcolor[HTML]{C0C0C0}7.68  & \multicolumn{3}{c|}{\multirow{-2}{*}{{\ul{12.61/4.95/3.38}}}}   & \multicolumn{3}{c|}{\multirow{-2}{*}{34.88/12.78/7.53}}    \\ \hline
                              & \multicolumn{1}{c|}{6}                             & \multicolumn{1}{c|}{10}                            & \multicolumn{1}{c|}{8}                             & \multicolumn{1}{c|}{7}                             & \multicolumn{1}{c|}{14}                            & \multicolumn{1}{c|}{17}                            & \multicolumn{1}{c|}{5}                             & \multicolumn{1}{c|}{10}                            & \multicolumn{1}{c|}{13}                            & 10                            & \multicolumn{3}{c|}{}                                                   & \multicolumn{3}{c|}{}                                      \\ \cline{2-11}
\multirow{-2}{*}{\textbf{2}}  & \multicolumn{1}{c|}{\cellcolor[HTML]{C0C0C0}9.44}  & \multicolumn{1}{c|}{\cellcolor[HTML]{C0C0C0}10.49} & \multicolumn{1}{c|}{\cellcolor[HTML]{C0C0C0}11.99} & \multicolumn{1}{c|}{\cellcolor[HTML]{C0C0C0}10.91} & \multicolumn{1}{c|}{\cellcolor[HTML]{C0C0C0}12.41} & \multicolumn{1}{c|}{\cellcolor[HTML]{C0C0C0}13.06} & \multicolumn{1}{c|}{\cellcolor[HTML]{C0C0C0}5.24}  & \multicolumn{1}{c|}{\cellcolor[HTML]{C0C0C0}9.57}  & \multicolumn{1}{c|}{\cellcolor[HTML]{C0C0C0}11.66} & \cellcolor[HTML]{C0C0C0}5.24  & \multicolumn{3}{c|}{\multirow{-2}{*}{{\ul{24.13/9.28/4.76}}}}   & \multicolumn{3}{c|}{\multirow{-2}{*}{52.44/19.27/12.65}}   \\ \hline
                              & \multicolumn{1}{c|}{16}                            & \multicolumn{1}{c|}{17}                            & \multicolumn{1}{c|}{7}                             & \multicolumn{1}{c|}{4}                             & \multicolumn{1}{c|}{16}                            & \multicolumn{1}{c|}{13}                            & \multicolumn{1}{c|}{14}                            & \multicolumn{1}{c|}{6}                             & \multicolumn{1}{c|}{3}                             & 4                             & \multicolumn{3}{c|}{}                                                   & \multicolumn{3}{c|}{}                                      \\ \cline{2-11}
\multirow{-2}{*}{\textbf{3}}  & \multicolumn{1}{c|}{\cellcolor[HTML]{C0C0C0}17.15} & \multicolumn{1}{c|}{\cellcolor[HTML]{C0C0C0}16.16} & \multicolumn{1}{c|}{\cellcolor[HTML]{C0C0C0}10.16} & \multicolumn{1}{c|}{\cellcolor[HTML]{C0C0C0}5.01}  & \multicolumn{1}{c|}{\cellcolor[HTML]{C0C0C0}13.51} & \multicolumn{1}{c|}{\cellcolor[HTML]{C0C0C0}12.29} & \multicolumn{1}{c|}{\cellcolor[HTML]{C0C0C0}14.65} & \multicolumn{1}{c|}{\cellcolor[HTML]{C0C0C0}8.82}  & \multicolumn{1}{c|}{\cellcolor[HTML]{C0C0C0}1.13}  & \cellcolor[HTML]{C0C0C0}1.13  & \multicolumn{3}{c|}{\multirow{-2}{*}{{\ul{17.57/6.31/3.16}}}}   & \multicolumn{3}{c|}{\multirow{-2}{*}{71.64/26.98/15.04}}   \\ \hline
                              & \multicolumn{1}{c|}{14}                            & \multicolumn{1}{c|}{2}                             & \multicolumn{1}{c|}{1}                             & \multicolumn{1}{c|}{20}                            & \multicolumn{1}{c|}{10}                            & \multicolumn{1}{c|}{14}                            & \multicolumn{1}{c|}{5}                             & \multicolumn{1}{c|}{2}                             & \multicolumn{1}{c|}{8}                             & 24                            & \multicolumn{3}{c|}{}                                                   & \multicolumn{3}{c|}{}                                      \\ \cline{2-11}
\multirow{-2}{*}{\textbf{4}}  & \multicolumn{1}{c|}{\cellcolor[HTML]{C0C0C0}16.57} & \multicolumn{1}{c|}{\cellcolor[HTML]{C0C0C0}6.42}  & \multicolumn{1}{c|}{\cellcolor[HTML]{C0C0C0}0}     & \multicolumn{1}{c|}{\cellcolor[HTML]{C0C0C0}18.17} & \multicolumn{1}{c|}{\cellcolor[HTML]{C0C0C0}11.09} & \multicolumn{1}{c|}{\cellcolor[HTML]{C0C0C0}13.15} & \multicolumn{1}{c|}{\cellcolor[HTML]{C0C0C0}2.47}  & \multicolumn{1}{c|}{\cellcolor[HTML]{C0C0C0}2.49}  & \multicolumn{1}{c|}{\cellcolor[HTML]{C0C0C0}9.02}  & \cellcolor[HTML]{C0C0C0}20.62 & \multicolumn{3}{c|}{\multirow{-2}{*}{{\ul{19.18/7.18/4.42}}}}   & \multicolumn{3}{c|}{\multirow{-2}{*}{86.54/31/17.1}}       \\ \hline
                              & \multicolumn{1}{c|}{4}                             & \multicolumn{1}{c|}{18}                            & \multicolumn{1}{c|}{-}                             & \multicolumn{1}{c|}{8}                             & \multicolumn{1}{c|}{10}                            & \multicolumn{1}{c|}{13}                            & \multicolumn{1}{c|}{28}                            & \multicolumn{1}{c|}{8}                             & \multicolumn{1}{c|}{6}                             & 5                             & \multicolumn{3}{c|}{}                                                   & \multicolumn{3}{c|}{}                                      \\ \cline{2-11}
\multirow{-2}{*}{\textbf{5}}  & \multicolumn{1}{c|}{\cellcolor[HTML]{C0C0C0}6.22}  & \multicolumn{1}{c|}{\cellcolor[HTML]{C0C0C0}18.6}  & \multicolumn{1}{c|}{\cellcolor[HTML]{C0C0C0}-}     & \multicolumn{1}{c|}{\cellcolor[HTML]{C0C0C0}10.1}  & \multicolumn{1}{c|}{\cellcolor[HTML]{C0C0C0}8.72}  & \multicolumn{1}{c|}{\cellcolor[HTML]{C0C0C0}12.51} & \multicolumn{1}{c|}{\cellcolor[HTML]{C0C0C0}23.49} & \multicolumn{1}{c|}{\cellcolor[HTML]{C0C0C0}9.74}  & \multicolumn{1}{c|}{\cellcolor[HTML]{C0C0C0}7.22}  & \cellcolor[HTML]{C0C0C0}3.38  & \multicolumn{3}{c|}{\multirow{-2}{*}{{\ul{15.78/6.25/4.51}}}}   & \multicolumn{3}{c|}{\multirow{-2}{*}{93.24/35.32/25.16}}   \\ \hline
                              & \multicolumn{1}{c|}{6}                             & \multicolumn{1}{c|}{11}                            & \multicolumn{1}{c|}{16}                            & \multicolumn{1}{c|}{4}                             & \multicolumn{1}{c|}{16}                            & \multicolumn{1}{c|}{-}                             & \multicolumn{1}{c|}{13}                            & \multicolumn{1}{c|}{8}                             & \multicolumn{1}{c|}{26}                            & -                             & \multicolumn{3}{c|}{}                                                   & \multicolumn{3}{c|}{}                                      \\ \cline{2-11}
\multirow{-2}{*}{\textbf{6}}  & \multicolumn{1}{c|}{\cellcolor[HTML]{C0C0C0}7.56}  & \multicolumn{1}{c|}{\cellcolor[HTML]{C0C0C0}12.95} & \multicolumn{1}{c|}{\cellcolor[HTML]{C0C0C0}13.5}  & \multicolumn{1}{c|}{\cellcolor[HTML]{C0C0C0}7.56}  & \multicolumn{1}{c|}{\cellcolor[HTML]{C0C0C0}13.21} & \multicolumn{1}{c|}{\cellcolor[HTML]{C0C0C0}-}     & \multicolumn{1}{c|}{\cellcolor[HTML]{C0C0C0}9.5}   & \multicolumn{1}{c|}{\cellcolor[HTML]{C0C0C0}9.63}  & \multicolumn{1}{c|}{\cellcolor[HTML]{C0C0C0}26.08} & \cellcolor[HTML]{C0C0C0}-     & \multicolumn{3}{c|}{\multirow{-2}{*}{{\ul{17.57/6.92/3.56}}}}   & \multicolumn{3}{c|}{\multirow{-2}{*}{103.04/34.62/15.41}}  \\ \hline
                              & \multicolumn{1}{c|}{18}                            & \multicolumn{1}{c|}{7}                             & \multicolumn{1}{c|}{14}                            & \multicolumn{1}{c|}{-}                             & \multicolumn{1}{c|}{7}                             & \multicolumn{1}{c|}{24}                            & \multicolumn{1}{c|}{-}                             & \multicolumn{1}{c|}{-}                             & \multicolumn{1}{c|}{26}                            & 4                             & \multicolumn{3}{c|}{}                                                   & \multicolumn{3}{c|}{}                                      \\ \cline{2-11}
\multirow{-2}{*}{\textbf{7}}  & \multicolumn{1}{c|}{\cellcolor[HTML]{C0C0C0}17.81} & \multicolumn{1}{c|}{\cellcolor[HTML]{C0C0C0}10.44} & \multicolumn{1}{c|}{\cellcolor[HTML]{C0C0C0}12.59} & \multicolumn{1}{c|}{\cellcolor[HTML]{C0C0C0}-}     & \multicolumn{1}{c|}{\cellcolor[HTML]{C0C0C0}11.83} & \multicolumn{1}{c|}{\cellcolor[HTML]{C0C0C0}19.16} & \multicolumn{1}{c|}{\cellcolor[HTML]{C0C0C0}-}     & \multicolumn{1}{c|}{\cellcolor[HTML]{C0C0C0}-}     & \multicolumn{1}{c|}{\cellcolor[HTML]{C0C0C0}19.34} & \cellcolor[HTML]{C0C0C0}8.83  & \multicolumn{3}{c|}{\multirow{-2}{*}{{\ul{26.20/11.33/6.66}}}}  & \multicolumn{3}{c|}{\multirow{-2}{*}{102/37.22/17.28}}     \\ \hline
                              & \multicolumn{1}{c|}{-}                             & \multicolumn{1}{c|}{-}                             & \multicolumn{1}{c|}{34}                            & \multicolumn{1}{c|}{4}                             & \multicolumn{1}{c|}{-}                             & \multicolumn{1}{c|}{-}                             & \multicolumn{1}{c|}{26}                            & \multicolumn{1}{c|}{2}                             & \multicolumn{1}{c|}{-}                             & 34                            & \multicolumn{3}{c|}{}                                                   & \multicolumn{3}{c|}{}                                      \\ \cline{2-11}
\multirow{-2}{*}{\textbf{8}}  & \multicolumn{1}{c|}{\cellcolor[HTML]{C0C0C0}-}     & \multicolumn{1}{c|}{\cellcolor[HTML]{C0C0C0}-}     & \multicolumn{1}{c|}{\cellcolor[HTML]{C0C0C0}27.05} & \multicolumn{1}{c|}{\cellcolor[HTML]{C0C0C0}19.14} & \multicolumn{1}{c|}{\cellcolor[HTML]{C0C0C0}-}     & \multicolumn{1}{c|}{\cellcolor[HTML]{C0C0C0}-}     & \multicolumn{1}{c|}{\cellcolor[HTML]{C0C0C0}26.25} & \multicolumn{1}{c|}{\cellcolor[HTML]{C0C0C0}0}     & \multicolumn{1}{c|}{\cellcolor[HTML]{C0C0C0}-}     & \cellcolor[HTML]{C0C0C0}27.55 & \multicolumn{3}{c|}{\multirow{-2}{*}{{\ul{30.79/17.98/15.14}}}} & \multicolumn{3}{c|}{\multirow{-2}{*}{140.5/49.69/28.83}}   \\ \hline
                              & \multicolumn{1}{c|}{-}                             & \multicolumn{1}{c|}{-}                             & \multicolumn{1}{c|}{-}                             & \multicolumn{1}{c|}{40}                            & \multicolumn{1}{c|}{-}                             & \multicolumn{1}{c|}{10}                            & \multicolumn{1}{c|}{-}                             & \multicolumn{1}{c|}{-}                             & \multicolumn{1}{c|}{-}                             & 50                            & \multicolumn{3}{c|}{}                                                   & \multicolumn{3}{c|}{}                                      \\ \cline{2-11}
\multirow{-2}{*}{\textbf{9}}  & \multicolumn{1}{c|}{\cellcolor[HTML]{C0C0C0}-}     & \multicolumn{1}{c|}{\cellcolor[HTML]{C0C0C0}-}     & \multicolumn{1}{c|}{\cellcolor[HTML]{C0C0C0}-}     & \multicolumn{1}{c|}{\cellcolor[HTML]{C0C0C0}35.96} & \multicolumn{1}{c|}{\cellcolor[HTML]{C0C0C0}-}     & \multicolumn{1}{c|}{\cellcolor[HTML]{C0C0C0}26.04} & \multicolumn{1}{c|}{\cellcolor[HTML]{C0C0C0}-}     & \multicolumn{1}{c|}{\cellcolor[HTML]{C0C0C0}-}     & \multicolumn{1}{c|}{\cellcolor[HTML]{C0C0C0}-}     & \cellcolor[HTML]{C0C0C0}38.01 & \multicolumn{3}{c|}{\multirow{-2}{*}{{\ul{32.07/20.43/16.04}}}} & \multicolumn{3}{c|}{\multirow{-2}{*}{141.5/57.42/39.44}}   \\ \hline
                              & \multicolumn{1}{c|}{-}                             & \multicolumn{1}{c|}{-}                             & \multicolumn{1}{c|}{-}                             & \multicolumn{1}{c|}{-}                             & \multicolumn{1}{c|}{-}                             & \multicolumn{1}{c|}{-}                             & \multicolumn{1}{c|}{-}                             & \multicolumn{1}{c|}{100}                           & \multicolumn{1}{c|}{-}                             & -                             & \multicolumn{3}{c|}{}                                                   & \multicolumn{3}{c|}{}                                      \\ \cline{2-11}
\multirow{-2}{*}{\textbf{10}} & \multicolumn{1}{c|}{\cellcolor[HTML]{C0C0C0}-}     & \multicolumn{1}{c|}{\cellcolor[HTML]{C0C0C0}-}     & \multicolumn{1}{c|}{\cellcolor[HTML]{C0C0C0}-}     & \multicolumn{1}{c|}{\cellcolor[HTML]{C0C0C0}-}     & \multicolumn{1}{c|}{\cellcolor[HTML]{C0C0C0}-}     & \multicolumn{1}{c|}{\cellcolor[HTML]{C0C0C0}-}     & \multicolumn{1}{c|}{\cellcolor[HTML]{C0C0C0}-}     & \multicolumn{1}{c|}{\cellcolor[HTML]{C0C0C0}100}   & \multicolumn{1}{c|}{\cellcolor[HTML]{C0C0C0}-}     & \cellcolor[HTML]{C0C0C0}-     & \multicolumn{3}{c|}{\multirow{-2}{*}{{\ul{0/0/0}}}}             & \multicolumn{3}{c|}{\multirow{-2}{*}{187.76/100.41/93.88}} \\ \hline
\end{tabular}
}
\label{tab:cifar10_resnet18}
\end{table*}

\begin{table*}[h]
\centering
\caption{The results on VGG16 (CIFAR-10).}
\resizebox{\linewidth}{!}{
\begin{tabular}{|c|cccccccccc|cllcll|}
\hline
                              & \multicolumn{10}{c|}{}                                                                                                                                                                                                                                                                                                                                                                                                                                                                                                     & \multicolumn{6}{c|}{\begin{tabular}[c]{@{}c@{}}\textbf{$L_1$/$L_2$/$L_\infty$($\times 10^{-2}$)}\end{tabular}}                                                                                            \\ \cline{12-17} 
\multirow{-2}{*}{\textbf{User}}        & \multicolumn{10}{c|}{\multirow{-2}{*}{\textbf{Data Composition (\%)}}}                                                                                                                                                                                                                                                                                                                                                                                                                                                     & \multicolumn{3}{c|}{Decaf}                                     & \multicolumn{3}{c|}{Random Guess}                 \\ \hline
                              & \multicolumn{1}{c|}{10}                            & \multicolumn{1}{c|}{10}                            & \multicolumn{1}{c|}{10}                            & \multicolumn{1}{c|}{10}                            & \multicolumn{1}{c|}{10}                            & \multicolumn{1}{c|}{10}                            & \multicolumn{1}{c|}{10}                            & \multicolumn{1}{c|}{10}                            & \multicolumn{1}{c|}{10}                            & 10                            & \multicolumn{3}{c|}{}                                                   & \multicolumn{3}{c|}{}                                      \\ \cline{2-11}
\multirow{-2}{*}{\textbf{1}}  & \multicolumn{1}{c|}{\cellcolor[HTML]{C0C0C0}9.66}  & \multicolumn{1}{c|}{\cellcolor[HTML]{C0C0C0}11.3}  & \multicolumn{1}{c|}{\cellcolor[HTML]{C0C0C0}11}    & \multicolumn{1}{c|}{\cellcolor[HTML]{C0C0C0}10.47} & \multicolumn{1}{c|}{\cellcolor[HTML]{C0C0C0}8.74}  & \multicolumn{1}{c|}{\cellcolor[HTML]{C0C0C0}9.69}  & \multicolumn{1}{c|}{\cellcolor[HTML]{C0C0C0}8.93}  & \multicolumn{1}{c|}{\cellcolor[HTML]{C0C0C0}11.3}  & \multicolumn{1}{c|}{\cellcolor[HTML]{C0C0C0}8.74}  & \cellcolor[HTML]{C0C0C0}10.15 & \multicolumn{3}{c|}{\multirow{-2}{*}{{\ul{8.46/3.03/1.30}}}}    & \multicolumn{3}{c|}{\multirow{-2}{*}{34.88/12.78/7.53}}    \\ \hline
                              & \multicolumn{1}{c|}{6}                             & \multicolumn{1}{c|}{10}                            & \multicolumn{1}{c|}{8}                             & \multicolumn{1}{c|}{7}                             & \multicolumn{1}{c|}{14}                            & \multicolumn{1}{c|}{17}                            & \multicolumn{1}{c|}{5}                             & \multicolumn{1}{c|}{10}                            & \multicolumn{1}{c|}{13}                            & 10                            & \multicolumn{3}{c|}{}                                                   & \multicolumn{3}{c|}{}                                      \\ \cline{2-11}
\multirow{-2}{*}{\textbf{2}}  & \multicolumn{1}{c|}{\cellcolor[HTML]{C0C0C0}7.04}  & \multicolumn{1}{c|}{\cellcolor[HTML]{C0C0C0}10.81} & \multicolumn{1}{c|}{\cellcolor[HTML]{C0C0C0}9.91}  & \multicolumn{1}{c|}{\cellcolor[HTML]{C0C0C0}8.21}  & \multicolumn{1}{c|}{\cellcolor[HTML]{C0C0C0}12.12} & \multicolumn{1}{c|}{\cellcolor[HTML]{C0C0C0}12.28} & \multicolumn{1}{c|}{\cellcolor[HTML]{C0C0C0}5.96}  & \multicolumn{1}{c|}{\cellcolor[HTML]{C0C0C0}10.89} & \multicolumn{1}{c|}{\cellcolor[HTML]{C0C0C0}12.83} & \cellcolor[HTML]{C0C0C0}9.95  & \multicolumn{3}{c|}{\multirow{-2}{*}{{\ul{13.67/5.87/4.72}}}}   & \multicolumn{3}{c|}{\multirow{-2}{*}{52.44/19.27/12.65}}   \\ \hline
                              & \multicolumn{1}{c|}{16}                            & \multicolumn{1}{c|}{17}                            & \multicolumn{1}{c|}{7}                             & \multicolumn{1}{c|}{4}                             & \multicolumn{1}{c|}{16}                            & \multicolumn{1}{c|}{13}                            & \multicolumn{1}{c|}{14}                            & \multicolumn{1}{c|}{6}                             & \multicolumn{1}{c|}{3}                             & 4                             & \multicolumn{3}{c|}{}                                                   & \multicolumn{3}{c|}{}                                      \\ \cline{2-11}
\multirow{-2}{*}{\textbf{3}}  & \multicolumn{1}{c|}{\cellcolor[HTML]{C0C0C0}13.67} & \multicolumn{1}{c|}{\cellcolor[HTML]{C0C0C0}14.91} & \multicolumn{1}{c|}{\cellcolor[HTML]{C0C0C0}11.33} & \multicolumn{1}{c|}{\cellcolor[HTML]{C0C0C0}5.51}  & \multicolumn{1}{c|}{\cellcolor[HTML]{C0C0C0}12.75} & \multicolumn{1}{c|}{\cellcolor[HTML]{C0C0C0}12.52} & \multicolumn{1}{c|}{\cellcolor[HTML]{C0C0C0}14.25} & \multicolumn{1}{c|}{\cellcolor[HTML]{C0C0C0}9.04}  & \multicolumn{1}{c|}{\cellcolor[HTML]{C0C0C0}3.01}  & \cellcolor[HTML]{C0C0C0}3.01  & \multicolumn{3}{c|}{\multirow{-2}{*}{{\ul{18.28/7.20/4.33}}}}   & \multicolumn{3}{c|}{\multirow{-2}{*}{71.64/26.98/15.04}}   \\ \hline
                              & \multicolumn{1}{c|}{14}                            & \multicolumn{1}{c|}{2}                             & \multicolumn{1}{c|}{1}                             & \multicolumn{1}{c|}{20}                            & \multicolumn{1}{c|}{10}                            & \multicolumn{1}{c|}{14}                            & \multicolumn{1}{c|}{5}                             & \multicolumn{1}{c|}{2}                             & \multicolumn{1}{c|}{8}                             & 24                            & \multicolumn{3}{c|}{}                                                   & \multicolumn{3}{c|}{}                                      \\ \cline{2-11}
\multirow{-2}{*}{\textbf{4}}  & \multicolumn{1}{c|}{\cellcolor[HTML]{C0C0C0}17.24} & \multicolumn{1}{c|}{\cellcolor[HTML]{C0C0C0}0}     & \multicolumn{1}{c|}{\cellcolor[HTML]{C0C0C0}0}     & \multicolumn{1}{c|}{\cellcolor[HTML]{C0C0C0}16.53} & \multicolumn{1}{c|}{\cellcolor[HTML]{C0C0C0}15.87} & \multicolumn{1}{c|}{\cellcolor[HTML]{C0C0C0}14.21} & \multicolumn{1}{c|}{\cellcolor[HTML]{C0C0C0}9.52}  & \multicolumn{1}{c|}{\cellcolor[HTML]{C0C0C0}0}     & \multicolumn{1}{c|}{\cellcolor[HTML]{C0C0C0}11.43} & \cellcolor[HTML]{C0C0C0}15.21 & \multicolumn{3}{c|}{\multirow{-2}{*}{{\ul{34.53/13.25/8.79}}}}  & \multicolumn{3}{c|}{\multirow{-2}{*}{86.54/31/17.1}}       \\ \hline
                              & \multicolumn{1}{c|}{4}                             & \multicolumn{1}{c|}{18}                            & \multicolumn{1}{c|}{-}                             & \multicolumn{1}{c|}{8}                             & \multicolumn{1}{c|}{10}                            & \multicolumn{1}{c|}{13}                            & \multicolumn{1}{c|}{28}                            & \multicolumn{1}{c|}{8}                             & \multicolumn{1}{c|}{6}                             & 5                             & \multicolumn{3}{c|}{}                                                   & \multicolumn{3}{c|}{}                                      \\ \cline{2-11}
\multirow{-2}{*}{\textbf{5}}  & \multicolumn{1}{c|}{\cellcolor[HTML]{C0C0C0}8.94}  & \multicolumn{1}{c|}{\cellcolor[HTML]{C0C0C0}14.93} & \multicolumn{1}{c|}{\cellcolor[HTML]{C0C0C0}-}     & \multicolumn{1}{c|}{\cellcolor[HTML]{C0C0C0}9.26}  & \multicolumn{1}{c|}{\cellcolor[HTML]{C0C0C0}11.09} & \multicolumn{1}{c|}{\cellcolor[HTML]{C0C0C0}11.75} & \multicolumn{1}{c|}{\cellcolor[HTML]{C0C0C0}15.2}  & \multicolumn{1}{c|}{\cellcolor[HTML]{C0C0C0}10.79} & \multicolumn{1}{c|}{\cellcolor[HTML]{C0C0C0}9.1}   & \cellcolor[HTML]{C0C0C0}8.94  & \multicolumn{3}{c|}{\multirow{-2}{*}{{\ul{34.24/15.33/12.80}}}} & \multicolumn{3}{c|}{\multirow{-2}{*}{93.24/35.32/25.16}}   \\ \hline
                              & \multicolumn{1}{c|}{6}                             & \multicolumn{1}{c|}{11}                            & \multicolumn{1}{c|}{16}                            & \multicolumn{1}{c|}{4}                             & \multicolumn{1}{c|}{16}                            & \multicolumn{1}{c|}{-}                             & \multicolumn{1}{c|}{13}                            & \multicolumn{1}{c|}{8}                             & \multicolumn{1}{c|}{26}                            & -                             & \multicolumn{3}{c|}{}                                                   & \multicolumn{3}{c|}{}                                      \\ \cline{2-11}
\multirow{-2}{*}{\textbf{6}}  & \multicolumn{1}{c|}{\cellcolor[HTML]{C0C0C0}10.29} & \multicolumn{1}{c|}{\cellcolor[HTML]{C0C0C0}10.32} & \multicolumn{1}{c|}{\cellcolor[HTML]{C0C0C0}14.65} & \multicolumn{1}{c|}{\cellcolor[HTML]{C0C0C0}11.49} & \multicolumn{1}{c|}{\cellcolor[HTML]{C0C0C0}13.13} & \multicolumn{1}{c|}{\cellcolor[HTML]{C0C0C0}-}     & \multicolumn{1}{c|}{\cellcolor[HTML]{C0C0C0}13.9}  & \multicolumn{1}{c|}{\cellcolor[HTML]{C0C0C0}10.09} & \multicolumn{1}{c|}{\cellcolor[HTML]{C0C0C0}16.13} & \cellcolor[HTML]{C0C0C0}-     & \multicolumn{3}{c|}{\multirow{-2}{*}{{\ul{29.54/13.70/9.87}}}}  & \multicolumn{3}{c|}{\multirow{-2}{*}{103.04/34.62/15.41}}  \\ \hline
                              & \multicolumn{1}{c|}{18}                            & \multicolumn{1}{c|}{7}                             & \multicolumn{1}{c|}{14}                            & \multicolumn{1}{c|}{-}                             & \multicolumn{1}{c|}{7}                             & \multicolumn{1}{c|}{24}                            & \multicolumn{1}{c|}{-}                             & \multicolumn{1}{c|}{-}                             & \multicolumn{1}{c|}{26}                            & 4                             & \multicolumn{3}{c|}{}                                                   & \multicolumn{3}{c|}{}                                      \\ \cline{2-11}
\multirow{-2}{*}{\textbf{7}}  & \multicolumn{1}{c|}{\cellcolor[HTML]{C0C0C0}16.47} & \multicolumn{1}{c|}{\cellcolor[HTML]{C0C0C0}10.89} & \multicolumn{1}{c|}{\cellcolor[HTML]{C0C0C0}15.05} & \multicolumn{1}{c|}{\cellcolor[HTML]{C0C0C0}-}     & \multicolumn{1}{c|}{\cellcolor[HTML]{C0C0C0}11.55} & \multicolumn{1}{c|}{\cellcolor[HTML]{C0C0C0}17.03} & \multicolumn{1}{c|}{\cellcolor[HTML]{C0C0C0}-}     & \multicolumn{1}{c|}{\cellcolor[HTML]{C0C0C0}-}     & \multicolumn{1}{c|}{\cellcolor[HTML]{C0C0C0}18.13} & \cellcolor[HTML]{C0C0C0}10.89 & \multicolumn{3}{c|}{\multirow{-2}{*}{{\ul{32.75/14.05/7.87}}}}  & \multicolumn{3}{c|}{\multirow{-2}{*}{102/37.22/17.28}}     \\ \hline
                              & \multicolumn{1}{c|}{-}                             & \multicolumn{1}{c|}{-}                             & \multicolumn{1}{c|}{34}                            & \multicolumn{1}{c|}{4}                             & \multicolumn{1}{c|}{-}                             & \multicolumn{1}{c|}{-}                             & \multicolumn{1}{c|}{26}                            & \multicolumn{1}{c|}{2}                             & \multicolumn{1}{c|}{-}                             & 34                            & \multicolumn{3}{c|}{}                                                   & \multicolumn{3}{c|}{}                                      \\ \cline{2-11}
\multirow{-2}{*}{\textbf{8}}  & \multicolumn{1}{c|}{\cellcolor[HTML]{C0C0C0}-}     & \multicolumn{1}{c|}{\cellcolor[HTML]{C0C0C0}-}     & \multicolumn{1}{c|}{\cellcolor[HTML]{C0C0C0}33.83} & \multicolumn{1}{c|}{\cellcolor[HTML]{C0C0C0}0}     & \multicolumn{1}{c|}{\cellcolor[HTML]{C0C0C0}-}     & \multicolumn{1}{c|}{\cellcolor[HTML]{C0C0C0}-}     & \multicolumn{1}{c|}{\cellcolor[HTML]{C0C0C0}34.91} & \multicolumn{1}{c|}{\cellcolor[HTML]{C0C0C0}0}     & \multicolumn{1}{c|}{\cellcolor[HTML]{C0C0C0}-}     & \cellcolor[HTML]{C0C0C0}31.25 & \multicolumn{3}{c|}{\multirow{-2}{*}{{\ul{17.83/10.34/8.91}}}}  & \multicolumn{3}{c|}{\multirow{-2}{*}{140.5/49.69/28.83}}   \\ \hline
                              & \multicolumn{1}{c|}{-}                             & \multicolumn{1}{c|}{-}                             & \multicolumn{1}{c|}{-}                             & \multicolumn{1}{c|}{40}                            & \multicolumn{1}{c|}{-}                             & \multicolumn{1}{c|}{10}                            & \multicolumn{1}{c|}{-}                             & \multicolumn{1}{c|}{-}                             & \multicolumn{1}{c|}{-}                             & 50                            & \multicolumn{3}{c|}{}                                                   & \multicolumn{3}{c|}{}                                      \\ \cline{2-11}
\multirow{-2}{*}{\textbf{9}}  & \multicolumn{1}{c|}{\cellcolor[HTML]{C0C0C0}-}     & \multicolumn{1}{c|}{\cellcolor[HTML]{C0C0C0}-}     & \multicolumn{1}{c|}{\cellcolor[HTML]{C0C0C0}-}     & \multicolumn{1}{c|}{\cellcolor[HTML]{C0C0C0}34.12} & \multicolumn{1}{c|}{\cellcolor[HTML]{C0C0C0}-}     & \multicolumn{1}{c|}{\cellcolor[HTML]{C0C0C0}29.43} & \multicolumn{1}{c|}{\cellcolor[HTML]{C0C0C0}-}     & \multicolumn{1}{c|}{\cellcolor[HTML]{C0C0C0}-}     & \multicolumn{1}{c|}{\cellcolor[HTML]{C0C0C0}-}     & \cellcolor[HTML]{C0C0C0}36.45 & \multicolumn{3}{c|}{\multirow{-2}{*}{{\ul{38.86/24.41/19.43}}}} & \multicolumn{3}{c|}{\multirow{-2}{*}{141.5/57.42/39.44}}   \\ \hline
                              & \multicolumn{1}{c|}{-}                             & \multicolumn{1}{c|}{-}                             & \multicolumn{1}{c|}{-}                             & \multicolumn{1}{c|}{-}                             & \multicolumn{1}{c|}{-}                             & \multicolumn{1}{c|}{-}                             & \multicolumn{1}{c|}{-}                             & \multicolumn{1}{c|}{100}                           & \multicolumn{1}{c|}{-}                             & -                             & \multicolumn{3}{c|}{}                                                   & \multicolumn{3}{c|}{}                                      \\ \cline{2-11}
\multirow{-2}{*}{\textbf{10}} & \multicolumn{1}{c|}{\cellcolor[HTML]{C0C0C0}-}     & \multicolumn{1}{c|}{\cellcolor[HTML]{C0C0C0}-}     & \multicolumn{1}{c|}{\cellcolor[HTML]{C0C0C0}-}     & \multicolumn{1}{c|}{\cellcolor[HTML]{C0C0C0}-}     & \multicolumn{1}{c|}{\cellcolor[HTML]{C0C0C0}-}     & \multicolumn{1}{c|}{\cellcolor[HTML]{C0C0C0}-}     & \multicolumn{1}{c|}{\cellcolor[HTML]{C0C0C0}-}     & \multicolumn{1}{c|}{\cellcolor[HTML]{C0C0C0}100}   & \multicolumn{1}{c|}{\cellcolor[HTML]{C0C0C0}-}     & \cellcolor[HTML]{C0C0C0}-     & \multicolumn{3}{c|}{\multirow{-2}{*}{{\ul{0/0/0}}}}             & \multicolumn{3}{c|}{\multirow{-2}{*}{187.76/100.41/93.88}} \\ \hline
\end{tabular}
}
\label{tab:cifar10_vgg16}
\end{table*}

\begin{table*}[h]
\centering
\caption{The results without null classes removal (MNIST).}
\resizebox{\linewidth}{!}{
\begin{tabular}{|c|cccccccccc|cllcll|}
\hline
                                & \multicolumn{10}{c|}{}                                                                                                                                                                                                                                                                                                                                                                                                                                                                                                     & \multicolumn{6}{c|}{\begin{tabular}[c]{@{}c@{}}\textbf{$L_1$/$L_2$/$L_\infty$($\times 10^{-2}$)}\end{tabular}}                                                                                            \\ \cline{12-17} 
\multirow{-2}{*}{\textbf{User}} & \multicolumn{10}{c|}{\multirow{-2}{*}{\textbf{Data Composition (\%)}}}                                                                                                                                                                                                                                                                                                                                                                                                                                                     & \multicolumn{3}{c|}{Decaf}                                      & \multicolumn{3}{c|}{Random Guess}                \\ \hline
                                & \multicolumn{1}{c|}{10}                            & \multicolumn{1}{c|}{10}                            & \multicolumn{1}{c|}{10}                            & \multicolumn{1}{c|}{10}                            & \multicolumn{1}{c|}{10}                            & \multicolumn{1}{c|}{10}                            & \multicolumn{1}{c|}{10}                            & \multicolumn{1}{c|}{10}                            & \multicolumn{1}{c|}{10}                            & 10                            & \multicolumn{3}{c|}{}                                                    & \multicolumn{3}{c|}{}                                     \\ \cline{2-11}
\multirow{-2}{*}{\textbf{1}}    & \multicolumn{1}{c|}{\cellcolor[HTML]{C0C0C0}9.29}  & \multicolumn{1}{c|}{\cellcolor[HTML]{C0C0C0}10.93} & \multicolumn{1}{c|}{\cellcolor[HTML]{C0C0C0}10.34} & \multicolumn{1}{c|}{\cellcolor[HTML]{C0C0C0}11.28} & \multicolumn{1}{c|}{\cellcolor[HTML]{C0C0C0}10.49} & \multicolumn{1}{c|}{\cellcolor[HTML]{C0C0C0}10.18} & \multicolumn{1}{c|}{\cellcolor[HTML]{C0C0C0}10.72} & \multicolumn{1}{c|}{\cellcolor[HTML]{C0C0C0}8.53}  & \multicolumn{1}{c|}{\cellcolor[HTML]{C0C0C0}8.53}  & \cellcolor[HTML]{C0C0C0}9.69  & \multicolumn{3}{c|}{\multirow{-2}{*}{{\ul{7.9/2.89/1.47}}}}      & \multicolumn{3}{c|}{\multirow{-2}{*}{34.62/14.34/9.38}}   \\ \hline
                                & \multicolumn{1}{c|}{6.67}                          & \multicolumn{1}{c|}{10}                            & \multicolumn{1}{c|}{8.33}                          & \multicolumn{1}{c|}{7.5}                           & \multicolumn{1}{c|}{13.33}                         & \multicolumn{1}{c|}{15.83}                         & \multicolumn{1}{c|}{5.83}                          & \multicolumn{1}{c|}{10}                            & \multicolumn{1}{c|}{12.5}                          & 10                            & \multicolumn{3}{c|}{}                                                    & \multicolumn{3}{c|}{}                                     \\ \cline{2-11}
\multirow{-2}{*}{\textbf{2}}    & \multicolumn{1}{c|}{\cellcolor[HTML]{C0C0C0}10.88} & \multicolumn{1}{c|}{\cellcolor[HTML]{C0C0C0}10.43} & \multicolumn{1}{c|}{\cellcolor[HTML]{C0C0C0}8.33}  & \multicolumn{1}{c|}{\cellcolor[HTML]{C0C0C0}8.53}  & \multicolumn{1}{c|}{\cellcolor[HTML]{C0C0C0}10.25} & \multicolumn{1}{c|}{\cellcolor[HTML]{C0C0C0}15.06} & \multicolumn{1}{c|}{\cellcolor[HTML]{C0C0C0}4.41}  & \multicolumn{1}{c|}{\cellcolor[HTML]{C0C0C0}12.18} & \multicolumn{1}{c|}{\cellcolor[HTML]{C0C0C0}8.9}   & \cellcolor[HTML]{C0C0C0}11.03 & \multicolumn{3}{c|}{\multirow{-2}{*}{{\ul{17.77/7.06/4.21}}}}    & \multicolumn{3}{c|}{\multirow{-2}{*}{47.59/17.85/10.53}}  \\ \hline
                                & \multicolumn{1}{c|}{15}                            & \multicolumn{1}{c|}{15.83}                         & \multicolumn{1}{c|}{7.5}                           & \multicolumn{1}{c|}{5}                             & \multicolumn{1}{c|}{15}                            & \multicolumn{1}{c|}{12.5}                          & \multicolumn{1}{c|}{13.33}                         & \multicolumn{1}{c|}{6.67}                          & \multicolumn{1}{c|}{4.17}                          & 5                             & \multicolumn{3}{c|}{}                                                    & \multicolumn{3}{c|}{}                                     \\ \cline{2-11}
\multirow{-2}{*}{\textbf{3}}    & \multicolumn{1}{c|}{\cellcolor[HTML]{C0C0C0}8.06}  & \multicolumn{1}{c|}{\cellcolor[HTML]{C0C0C0}12.78} & \multicolumn{1}{c|}{\cellcolor[HTML]{C0C0C0}10.29} & \multicolumn{1}{c|}{\cellcolor[HTML]{C0C0C0}9.87}  & \multicolumn{1}{c|}{\cellcolor[HTML]{C0C0C0}11.61} & \multicolumn{1}{c|}{\cellcolor[HTML]{C0C0C0}9.18}  & \multicolumn{1}{c|}{\cellcolor[HTML]{C0C0C0}10.67} & \multicolumn{1}{c|}{\cellcolor[HTML]{C0C0C0}9.51}  & \multicolumn{1}{c|}{\cellcolor[HTML]{C0C0C0}11.11} & \cellcolor[HTML]{C0C0C0}6.92  & \multicolumn{3}{c|}{\multirow{-2}{*}{{\ul{38.73/13.36/6.94}}}}   & \multicolumn{3}{c|}{\multirow{-2}{*}{60.08/21.71/11.98}}  \\ \hline
                                & \multicolumn{1}{c|}{13.33}                         & \multicolumn{1}{c|}{3.33}                          & \multicolumn{1}{c|}{2.5}                           & \multicolumn{1}{c|}{18.33}                         & \multicolumn{1}{c|}{10}                            & \multicolumn{1}{c|}{13.33}                         & \multicolumn{1}{c|}{5.83}                          & \multicolumn{1}{c|}{1.67}                          & \multicolumn{1}{c|}{8.33}                          & 23.33                         & \multicolumn{3}{c|}{}                                                    & \multicolumn{3}{c|}{}                                     \\ \cline{2-11}
\multirow{-2}{*}{\textbf{4}}    & \multicolumn{1}{c|}{\cellcolor[HTML]{C0C0C0}11.53} & \multicolumn{1}{c|}{\cellcolor[HTML]{C0C0C0}12.3}  & \multicolumn{1}{c|}{\cellcolor[HTML]{C0C0C0}7.43}  & \multicolumn{1}{c|}{\cellcolor[HTML]{C0C0C0}10.34} & \multicolumn{1}{c|}{\cellcolor[HTML]{C0C0C0}10.85} & \multicolumn{1}{c|}{\cellcolor[HTML]{C0C0C0}10.9}  & \multicolumn{1}{c|}{\cellcolor[HTML]{C0C0C0}7.03}  & \multicolumn{1}{c|}{\cellcolor[HTML]{C0C0C0}5.34}  & \multicolumn{1}{c|}{\cellcolor[HTML]{C0C0C0}9.01}  & \cellcolor[HTML]{C0C0C0}15.29 & \multicolumn{3}{c|}{\multirow{-2}{*}{{\ul{40.57/16.08/8.97}}}}   & \multicolumn{3}{c|}{\multirow{-2}{*}{97.96/37.78/23.33}}  \\ \hline
                                & \multicolumn{1}{c|}{5}                             & \multicolumn{1}{c|}{18.33}                         & \multicolumn{1}{c|}{-}                             & \multicolumn{1}{c|}{8.33}                          & \multicolumn{1}{c|}{10}                            & \multicolumn{1}{c|}{12.5}                          & \multicolumn{1}{c|}{25}                            & \multicolumn{1}{c|}{8.33}                          & \multicolumn{1}{c|}{6.67}                          & 5.83                          & \multicolumn{3}{c|}{}                                                    & \multicolumn{3}{c|}{}                                     \\ \cline{2-11}
\multirow{-2}{*}{\textbf{5}}    & \multicolumn{1}{c|}{\cellcolor[HTML]{C0C0C0}16.6}  & \multicolumn{1}{c|}{\cellcolor[HTML]{C0C0C0}16.58} & \multicolumn{1}{c|}{\cellcolor[HTML]{C0C0C0}7.41}  & \multicolumn{1}{c|}{\cellcolor[HTML]{C0C0C0}10.2}  & \multicolumn{1}{c|}{\cellcolor[HTML]{C0C0C0}8.89}  & \multicolumn{1}{c|}{\cellcolor[HTML]{C0C0C0}8.57}  & \multicolumn{1}{c|}{\cellcolor[HTML]{C0C0C0}9.53}  & \multicolumn{1}{c|}{\cellcolor[HTML]{C0C0C0}7.41}  & \multicolumn{1}{c|}{\cellcolor[HTML]{C0C0C0}7.41}  & \cellcolor[HTML]{C0C0C0}7.41  & \multicolumn{3}{c|}{\multirow{-2}{*}{{\ul{46.39/21.35/15.47}}}}  & \multicolumn{3}{c|}{\multirow{-2}{*}{80.91/34.45/24.43}}  \\ \hline
                                & \multicolumn{1}{c|}{6.67}                          & \multicolumn{1}{c|}{10.83}                         & \multicolumn{1}{c|}{15}                            & \multicolumn{1}{c|}{5}                             & \multicolumn{1}{c|}{16.67}                         & \multicolumn{1}{c|}{-}                             & \multicolumn{1}{c|}{12.5}                          & \multicolumn{1}{c|}{8.33}                          & \multicolumn{1}{c|}{25}                            & -                             & \multicolumn{3}{c|}{}                                                    & \multicolumn{3}{c|}{}                                     \\ \cline{2-11}
\multirow{-2}{*}{\textbf{6}}    & \multicolumn{1}{c|}{\cellcolor[HTML]{C0C0C0}9.88}  & \multicolumn{1}{c|}{\cellcolor[HTML]{C0C0C0}10.19} & \multicolumn{1}{c|}{\cellcolor[HTML]{C0C0C0}12.78} & \multicolumn{1}{c|}{\cellcolor[HTML]{C0C0C0}12.41} & \multicolumn{1}{c|}{\cellcolor[HTML]{C0C0C0}14.3}  & \multicolumn{1}{c|}{\cellcolor[HTML]{C0C0C0}5.8}   & \multicolumn{1}{c|}{\cellcolor[HTML]{C0C0C0}8}     & \multicolumn{1}{c|}{\cellcolor[HTML]{C0C0C0}8.33}  & \multicolumn{1}{c|}{\cellcolor[HTML]{C0C0C0}13.68} & \cellcolor[HTML]{C0C0C0}4.64  & \multicolumn{3}{c|}{\multirow{-2}{*}{{\ul{42.12/16.73/11.32}}}}  & \multicolumn{3}{c|}{\multirow{-2}{*}{74.34/28.52/15.85}}  \\ \hline
                                & \multicolumn{1}{c|}{16.67}                         & \multicolumn{1}{c|}{7.5}                           & \multicolumn{1}{c|}{13.33}                         & \multicolumn{1}{c|}{-}                             & \multicolumn{1}{c|}{7.5}                           & \multicolumn{1}{c|}{25}                            & \multicolumn{1}{c|}{-}                             & \multicolumn{1}{c|}{-}                             & \multicolumn{1}{c|}{26.67}                         & 3.33                          & \multicolumn{3}{c|}{}                                                    & \multicolumn{3}{c|}{}                                     \\ \cline{2-11}
\multirow{-2}{*}{\textbf{7}}    & \multicolumn{1}{c|}{\cellcolor[HTML]{C0C0C0}14.1}  & \multicolumn{1}{c|}{\cellcolor[HTML]{C0C0C0}12.11} & \multicolumn{1}{c|}{\cellcolor[HTML]{C0C0C0}15.37} & \multicolumn{1}{c|}{\cellcolor[HTML]{C0C0C0}4.32}  & \multicolumn{1}{c|}{\cellcolor[HTML]{C0C0C0}11.38} & \multicolumn{1}{c|}{\cellcolor[HTML]{C0C0C0}14.52} & \multicolumn{1}{c|}{\cellcolor[HTML]{C0C0C0}0.99}  & \multicolumn{1}{c|}{\cellcolor[HTML]{C0C0C0}0.94}  & \multicolumn{1}{c|}{\cellcolor[HTML]{C0C0C0}12.47} & \cellcolor[HTML]{C0C0C0}13.79 & \multicolumn{3}{c|}{\multirow{-2}{*}{{\ul{54.48 /22.10/14.20}}}} & \multicolumn{3}{c|}{\multirow{-2}{*}{107.36/37.67/18.42}} \\ \hline
                                & \multicolumn{1}{c|}{-}                             & \multicolumn{1}{c|}{-}                             & \multicolumn{1}{c|}{33.33}                         & \multicolumn{1}{c|}{5}                             & \multicolumn{1}{c|}{-}                             & \multicolumn{1}{c|}{-}                             & \multicolumn{1}{c|}{26.67}                         & \multicolumn{1}{c|}{3.33}                          & \multicolumn{1}{c|}{-}                             & 31.67                         & \multicolumn{3}{c|}{}                                                    & \multicolumn{3}{c|}{}                                     \\ \cline{2-11}
\multirow{-2}{*}{\textbf{8}}    & \multicolumn{1}{c|}{\cellcolor[HTML]{C0C0C0}5.91}  & \multicolumn{1}{c|}{\cellcolor[HTML]{C0C0C0}10.41} & \multicolumn{1}{c|}{\cellcolor[HTML]{C0C0C0}25.15} & \multicolumn{1}{c|}{\cellcolor[HTML]{C0C0C0}16.44} & \multicolumn{1}{c|}{\cellcolor[HTML]{C0C0C0}0.13}  & \multicolumn{1}{c|}{\cellcolor[HTML]{C0C0C0}-}     & \multicolumn{1}{c|}{\cellcolor[HTML]{C0C0C0}19.15} & \multicolumn{1}{c|}{\cellcolor[HTML]{C0C0C0}8.44}  & \multicolumn{1}{c|}{\cellcolor[HTML]{C0C0C0}0.01}  & \cellcolor[HTML]{C0C0C0}14.36 & \multicolumn{3}{c|}{\multirow{-2}{*}{{\ul{66.02/26.90/17.31}}}}  & \multicolumn{3}{c|}{\multirow{-2}{*}{137.2/49.05/28.62}}  \\ \hline
                                & \multicolumn{1}{c|}{-}                             & \multicolumn{1}{c|}{-}                             & \multicolumn{1}{c|}{-}                             & \multicolumn{1}{c|}{41.67}                         & \multicolumn{1}{c|}{-}                             & \multicolumn{1}{c|}{8.33}                          & \multicolumn{1}{c|}{-}                             & \multicolumn{1}{c|}{-}                             & \multicolumn{1}{c|}{-}                             & 50                            & \multicolumn{3}{c|}{}                                                    & \multicolumn{3}{c|}{}                                     \\ \cline{2-11}
\multirow{-2}{*}{\textbf{9}}    & \multicolumn{1}{c|}{\cellcolor[HTML]{C0C0C0}5.09}  & \multicolumn{1}{c|}{\cellcolor[HTML]{C0C0C0}4.47}  & \multicolumn{1}{c|}{\cellcolor[HTML]{C0C0C0}5.24}  & \multicolumn{1}{c|}{\cellcolor[HTML]{C0C0C0}29.51} & \multicolumn{1}{c|}{\cellcolor[HTML]{C0C0C0}3.76}  & \multicolumn{1}{c|}{\cellcolor[HTML]{C0C0C0}27.82} & \multicolumn{1}{c|}{\cellcolor[HTML]{C0C0C0}-}     & \multicolumn{1}{c|}{\cellcolor[HTML]{C0C0C0}-}     & \multicolumn{1}{c|}{\cellcolor[HTML]{C0C0C0}-}     & \cellcolor[HTML]{C0C0C0}24.12 & \multicolumn{3}{c|}{\multirow{-2}{*}{{\ul{76.08/35.85/25.88}}}}  & \multicolumn{3}{c|}{\multirow{-2}{*}{161.18/66.39/48.42}} \\ \hline
                                & \multicolumn{1}{c|}{-}                             & \multicolumn{1}{c|}{-}                             & \multicolumn{1}{c|}{-}                             & \multicolumn{1}{c|}{-}                             & \multicolumn{1}{c|}{-}                             & \multicolumn{1}{c|}{-}                             & \multicolumn{1}{c|}{-}                             & \multicolumn{1}{c|}{100}                           & \multicolumn{1}{c|}{-}                             & -                             & \multicolumn{3}{c|}{}                                                    & \multicolumn{3}{c|}{}                                     \\ \cline{2-11}
\multirow{-2}{*}{\textbf{10}}   & \multicolumn{1}{c|}{\cellcolor[HTML]{C0C0C0}2.76}  & \multicolumn{1}{c|}{\cellcolor[HTML]{C0C0C0}2.09}  & \multicolumn{1}{c|}{\cellcolor[HTML]{C0C0C0}4.81}  & \multicolumn{1}{c|}{\cellcolor[HTML]{C0C0C0}4.17}  & \multicolumn{1}{c|}{\cellcolor[HTML]{C0C0C0}3.4}   & \multicolumn{1}{c|}{\cellcolor[HTML]{C0C0C0}3.32}  & \multicolumn{1}{c|}{\cellcolor[HTML]{C0C0C0}6.23}  & \multicolumn{1}{c|}{\cellcolor[HTML]{C0C0C0}63.48} & \multicolumn{1}{c|}{\cellcolor[HTML]{C0C0C0}3.37}  & \cellcolor[HTML]{C0C0C0}6.37  & \multicolumn{3}{c|}{\multirow{-2}{*}{{\ul{73.04/38.72/36.52}}}}  & \multicolumn{3}{c|}{\multirow{-2}{*}{186.32/99.9/93.16}}  \\ \hline
\end{tabular}
}
\label{tab:without_removal}
\end{table*}

\begin{table*}[h]
\centering
\caption{The results without calibrator/$g^t_u$ (MNIST).}
\resizebox{\linewidth}{!}{
\begin{tabular}{|c|cccccccccc|cllcll|}
\hline
                                & \multicolumn{10}{c|}{}                                                                                                                                                                                                                                                                                                                                                                                                                                                                                                     & \multicolumn{6}{c|}{\begin{tabular}[c]{@{}c@{}}\textbf{$L_1$/$L_2$/$L_\infty$($\times 10^{-2}$)}\end{tabular}}                                                                                           \\ \cline{12-17} 
\multirow{-2}{*}{\textbf{User}} & \multicolumn{10}{c|}{\multirow{-2}{*}{\textbf{Data Composition (\%)}}}                                                                                                                                                                                                                                                                                                                                                                                                                                                     & \multicolumn{3}{c|}{Decaf}                                     & \multicolumn{3}{c|}{Random Guess}                \\ \hline
                                & \multicolumn{1}{c|}{10}                            & \multicolumn{1}{c|}{10}                            & \multicolumn{1}{c|}{10}                            & \multicolumn{1}{c|}{10}                            & \multicolumn{1}{c|}{10}                            & \multicolumn{1}{c|}{10}                            & \multicolumn{1}{c|}{10}                            & \multicolumn{1}{c|}{10}                            & \multicolumn{1}{c|}{10}                            & 10                            & \multicolumn{3}{c|}{}                                                   & \multicolumn{3}{c|}{}                                     \\ \cline{2-11}
\multirow{-2}{*}{\textbf{1}}    & \multicolumn{1}{c|}{\cellcolor[HTML]{C0C0C0}5.63}  & \multicolumn{1}{c|}{\cellcolor[HTML]{C0C0C0}13.89} & \multicolumn{1}{c|}{\cellcolor[HTML]{C0C0C0}11.78} & \multicolumn{1}{c|}{\cellcolor[HTML]{C0C0C0}14.36} & \multicolumn{1}{c|}{\cellcolor[HTML]{C0C0C0}7.07}  & \multicolumn{1}{c|}{\cellcolor[HTML]{C0C0C0}12.01} & \multicolumn{1}{c|}{\cellcolor[HTML]{C0C0C0}0}     & \multicolumn{1}{c|}{\cellcolor[HTML]{C0C0C0}7.74}  & \multicolumn{1}{c|}{\cellcolor[HTML]{C0C0C0}27.53} & \cellcolor[HTML]{C0C0C0}0     & \multicolumn{3}{c|}{\multirow{-2}{*}{{\ul{59.13/24.11/17.53}}}} & \multicolumn{3}{c|}{\multirow{-2}{*}{34.62/14.34/9.38}}   \\ \hline
                                & \multicolumn{1}{c|}{6.67}                          & \multicolumn{1}{c|}{10}                            & \multicolumn{1}{c|}{8.33}                          & \multicolumn{1}{c|}{7.5}                           & \multicolumn{1}{c|}{13.33}                         & \multicolumn{1}{c|}{15.83}                         & \multicolumn{1}{c|}{5.83}                          & \multicolumn{1}{c|}{10}                            & \multicolumn{1}{c|}{12.5}                          & 10                            & \multicolumn{3}{c|}{}                                                   & \multicolumn{3}{c|}{}                                     \\ \cline{2-11}
\multirow{-2}{*}{\textbf{2}}    & \multicolumn{1}{c|}{\cellcolor[HTML]{C0C0C0}13.13} & \multicolumn{1}{c|}{\cellcolor[HTML]{C0C0C0}11.02} & \multicolumn{1}{c|}{\cellcolor[HTML]{C0C0C0}2.52}  & \multicolumn{1}{c|}{\cellcolor[HTML]{C0C0C0}8.74}  & \multicolumn{1}{c|}{\cellcolor[HTML]{C0C0C0}6.28}  & \multicolumn{1}{c|}{\cellcolor[HTML]{C0C0C0}13.63} & \multicolumn{1}{c|}{\cellcolor[HTML]{C0C0C0}0}     & \multicolumn{1}{c|}{\cellcolor[HTML]{C0C0C0}28.03} & \multicolumn{1}{c|}{\cellcolor[HTML]{C0C0C0}9.02}  & \cellcolor[HTML]{C0C0C0}7.63  & \multicolumn{3}{c|}{\multirow{-2}{*}{{\ul{53.51/22.57/18.03}}}} & \multicolumn{3}{c|}{\multirow{-2}{*}{47.59/17.85/10.53}}  \\ \hline
                                & \multicolumn{1}{c|}{15}                            & \multicolumn{1}{c|}{15.83}                         & \multicolumn{1}{c|}{7.5}                           & \multicolumn{1}{c|}{5}                             & \multicolumn{1}{c|}{15}                            & \multicolumn{1}{c|}{12.5}                          & \multicolumn{1}{c|}{13.33}                         & \multicolumn{1}{c|}{6.67}                          & \multicolumn{1}{c|}{4.17}                          & 5                             & \multicolumn{3}{c|}{}                                                   & \multicolumn{3}{c|}{}                                     \\ \cline{2-11}
\multirow{-2}{*}{\textbf{3}}    & \multicolumn{1}{c|}{\cellcolor[HTML]{C0C0C0}12.15} & \multicolumn{1}{c|}{\cellcolor[HTML]{C0C0C0}22.48} & \multicolumn{1}{c|}{\cellcolor[HTML]{C0C0C0}13.59} & \multicolumn{1}{c|}{\cellcolor[HTML]{C0C0C0}0}     & \multicolumn{1}{c|}{\cellcolor[HTML]{C0C0C0}10.21} & \multicolumn{1}{c|}{\cellcolor[HTML]{C0C0C0}6.23}  & \multicolumn{1}{c|}{\cellcolor[HTML]{C0C0C0}4.32}  & \multicolumn{1}{c|}{\cellcolor[HTML]{C0C0C0}19.74} & \multicolumn{1}{c|}{\cellcolor[HTML]{C0C0C0}11.26} & \cellcolor[HTML]{C0C0C0}0     & \multicolumn{3}{c|}{\multirow{-2}{*}{{\ul{65.83/22.45/13.07}}}} & \multicolumn{3}{c|}{\multirow{-2}{*}{60.08/21.71/11.98}}  \\ \hline
                                & \multicolumn{1}{c|}{13.33}                         & \multicolumn{1}{c|}{3.33}                          & \multicolumn{1}{c|}{2.5}                           & \multicolumn{1}{c|}{18.33}                         & \multicolumn{1}{c|}{10}                            & \multicolumn{1}{c|}{13.33}                         & \multicolumn{1}{c|}{5.83}                          & \multicolumn{1}{c|}{1.67}                          & \multicolumn{1}{c|}{8.33}                          & 23.33                         & \multicolumn{3}{c|}{}                                                   & \multicolumn{3}{c|}{}                                     \\ \cline{2-11}
\multirow{-2}{*}{\textbf{4}}    & \multicolumn{1}{c|}{\cellcolor[HTML]{C0C0C0}9.24}  & \multicolumn{1}{c|}{\cellcolor[HTML]{C0C0C0}14.38} & \multicolumn{1}{c|}{\cellcolor[HTML]{C0C0C0}6.76}  & \multicolumn{1}{c|}{\cellcolor[HTML]{C0C0C0}17.78} & \multicolumn{1}{c|}{\cellcolor[HTML]{C0C0C0}15.47} & \multicolumn{1}{c|}{\cellcolor[HTML]{C0C0C0}9.85}  & \multicolumn{1}{c|}{\cellcolor[HTML]{C0C0C0}0}     & \multicolumn{1}{c|}{\cellcolor[HTML]{C0C0C0}0}     & \multicolumn{1}{c|}{\cellcolor[HTML]{C0C0C0}12.51} & \cellcolor[HTML]{C0C0C0}14.02 & \multicolumn{3}{c|}{\multirow{-2}{*}{{\ul{49.19/18.45/11.05}}}} & \multicolumn{3}{c|}{\multirow{-2}{*}{97.96/37.78/23.33}}  \\ \hline
                                & \multicolumn{1}{c|}{5}                             & \multicolumn{1}{c|}{18.33}                         & \multicolumn{1}{c|}{-}                             & \multicolumn{1}{c|}{8.33}                          & \multicolumn{1}{c|}{10}                            & \multicolumn{1}{c|}{12.5}                          & \multicolumn{1}{c|}{25}                            & \multicolumn{1}{c|}{8.33}                          & \multicolumn{1}{c|}{6.67}                          & 5.83                          & \multicolumn{3}{c|}{}                                                   & \multicolumn{3}{c|}{}                                     \\ \cline{2-11}
\multirow{-2}{*}{\textbf{5}}    & \multicolumn{1}{c|}{\cellcolor[HTML]{C0C0C0}10.55} & \multicolumn{1}{c|}{\cellcolor[HTML]{C0C0C0}27.31} & \multicolumn{1}{c|}{\cellcolor[HTML]{C0C0C0}-}     & \multicolumn{1}{c|}{\cellcolor[HTML]{C0C0C0}2.19}  & \multicolumn{1}{c|}{\cellcolor[HTML]{C0C0C0}9.05}  & \multicolumn{1}{c|}{\cellcolor[HTML]{C0C0C0}12.87} & \multicolumn{1}{c|}{\cellcolor[HTML]{C0C0C0}19.49} & \multicolumn{1}{c|}{\cellcolor[HTML]{C0C0C0}13.53} & \multicolumn{1}{c|}{\cellcolor[HTML]{C0C0C0}0}     & \cellcolor[HTML]{C0C0C0}5     & \multicolumn{3}{c|}{\multirow{-2}{*}{{\ul{40.19/15.90/8.98}}}}  & \multicolumn{3}{c|}{\multirow{-2}{*}{80.91/34.45/24.43}}  \\ \hline
                                & \multicolumn{1}{c|}{6.67}                          & \multicolumn{1}{c|}{10.83}                         & \multicolumn{1}{c|}{15}                            & \multicolumn{1}{c|}{5}                             & \multicolumn{1}{c|}{16.67}                         & \multicolumn{1}{c|}{-}                             & \multicolumn{1}{c|}{12.5}                          & \multicolumn{1}{c|}{8.33}                          & \multicolumn{1}{c|}{25}                            & -                             & \multicolumn{3}{c|}{}                                                   & \multicolumn{3}{c|}{}                                     \\ \cline{2-11}
\multirow{-2}{*}{\textbf{6}}    & \multicolumn{1}{c|}{\cellcolor[HTML]{C0C0C0}6.29}  & \multicolumn{1}{c|}{\cellcolor[HTML]{C0C0C0}8.96}  & \multicolumn{1}{c|}{\cellcolor[HTML]{C0C0C0}8.54}  & \multicolumn{1}{c|}{\cellcolor[HTML]{C0C0C0}22.19} & \multicolumn{1}{c|}{\cellcolor[HTML]{C0C0C0}18.25} & \multicolumn{1}{c|}{\cellcolor[HTML]{C0C0C0}-}     & \multicolumn{1}{c|}{\cellcolor[HTML]{C0C0C0}10.35} & \multicolumn{1}{c|}{\cellcolor[HTML]{C0C0C0}8.78}  & \multicolumn{1}{c|}{\cellcolor[HTML]{C0C0C0}16.64} & \cellcolor[HTML]{C0C0C0}-     & \multicolumn{3}{c|}{\multirow{-2}{*}{{\ul{38.44/20.45/17.19}}}} & \multicolumn{3}{c|}{\multirow{-2}{*}{74.34/28.52/15.85}}  \\ \hline
                                & \multicolumn{1}{c|}{16.67}                         & \multicolumn{1}{c|}{7.5}                           & \multicolumn{1}{c|}{13.33}                         & \multicolumn{1}{c|}{-}                             & \multicolumn{1}{c|}{7.5}                           & \multicolumn{1}{c|}{25}                            & \multicolumn{1}{c|}{-}                             & \multicolumn{1}{c|}{-}                             & \multicolumn{1}{c|}{26.67}                         & 3.33                          & \multicolumn{3}{c|}{}                                                   & \multicolumn{3}{c|}{}                                     \\ \cline{2-11}
\multirow{-2}{*}{\textbf{7}}    & \multicolumn{1}{c|}{\cellcolor[HTML]{C0C0C0}26.92} & \multicolumn{1}{c|}{\cellcolor[HTML]{C0C0C0}7.6}   & \multicolumn{1}{c|}{\cellcolor[HTML]{C0C0C0}5.36}  & \multicolumn{1}{c|}{\cellcolor[HTML]{C0C0C0}-}     & \multicolumn{1}{c|}{\cellcolor[HTML]{C0C0C0}4.44}  & \multicolumn{1}{c|}{\cellcolor[HTML]{C0C0C0}23.56} & \multicolumn{1}{c|}{\cellcolor[HTML]{C0C0C0}-}     & \multicolumn{1}{c|}{\cellcolor[HTML]{C0C0C0}-}     & \multicolumn{1}{c|}{\cellcolor[HTML]{C0C0C0}24.37} & \cellcolor[HTML]{C0C0C0}7.75  & \multicolumn{3}{c|}{\multirow{-2}{*}{{\ul{29.54/14.31/10.25}}}} & \multicolumn{3}{c|}{\multirow{-2}{*}{107.36/37.67/18.42}} \\ \hline
                                & \multicolumn{1}{c|}{-}                             & \multicolumn{1}{c|}{-}                             & \multicolumn{1}{c|}{33.33}                         & \multicolumn{1}{c|}{5}                             & \multicolumn{1}{c|}{-}                             & \multicolumn{1}{c|}{-}                             & \multicolumn{1}{c|}{26.67}                         & \multicolumn{1}{c|}{3.33}                          & \multicolumn{1}{c|}{-}                             & 31.67                         & \multicolumn{3}{c|}{}                                                   & \multicolumn{3}{c|}{}                                     \\ \cline{2-11}
\multirow{-2}{*}{\textbf{8}}    & \multicolumn{1}{c|}{\cellcolor[HTML]{C0C0C0}-}     & \multicolumn{1}{c|}{\cellcolor[HTML]{C0C0C0}-}     & \multicolumn{1}{c|}{\cellcolor[HTML]{C0C0C0}21.55} & \multicolumn{1}{c|}{\cellcolor[HTML]{C0C0C0}1.93}  & \multicolumn{1}{c|}{\cellcolor[HTML]{C0C0C0}-}     & \multicolumn{1}{c|}{\cellcolor[HTML]{C0C0C0}-}     & \multicolumn{1}{c|}{\cellcolor[HTML]{C0C0C0}32.97} & \multicolumn{1}{c|}{\cellcolor[HTML]{C0C0C0}8.39}  & \multicolumn{1}{c|}{\cellcolor[HTML]{C0C0C0}-}     & \cellcolor[HTML]{C0C0C0}35.16 & \multicolumn{3}{c|}{\multirow{-2}{*}{{\ul{29.71/15.02/11.78}}}} & \multicolumn{3}{c|}{\multirow{-2}{*}{137.2/49.05/28.62}}  \\ \hline
                                & \multicolumn{1}{c|}{-}                             & \multicolumn{1}{c|}{-}                             & \multicolumn{1}{c|}{-}                             & \multicolumn{1}{c|}{41.67}                         & \multicolumn{1}{c|}{-}                             & \multicolumn{1}{c|}{8.33}                          & \multicolumn{1}{c|}{-}                             & \multicolumn{1}{c|}{-}                             & \multicolumn{1}{c|}{-}                             & 50                            & \multicolumn{3}{c|}{}                                                   & \multicolumn{3}{c|}{}                                     \\ \cline{2-11}
\multirow{-2}{*}{\textbf{9}}    & \multicolumn{1}{c|}{\cellcolor[HTML]{C0C0C0}-}     & \multicolumn{1}{c|}{\cellcolor[HTML]{C0C0C0}-}     & \multicolumn{1}{c|}{\cellcolor[HTML]{C0C0C0}-}     & \multicolumn{1}{c|}{\cellcolor[HTML]{C0C0C0}41.75} & \multicolumn{1}{c|}{\cellcolor[HTML]{C0C0C0}-}     & \multicolumn{1}{c|}{\cellcolor[HTML]{C0C0C0}7}     & \multicolumn{1}{c|}{\cellcolor[HTML]{C0C0C0}-}     & \multicolumn{1}{c|}{\cellcolor[HTML]{C0C0C0}-}     & \multicolumn{1}{c|}{\cellcolor[HTML]{C0C0C0}-}     & \cellcolor[HTML]{C0C0C0}51.25 & \multicolumn{3}{c|}{\multirow{-2}{*}{{\ul{2.6/1.83/1.33}}}}     & \multicolumn{3}{c|}{\multirow{-2}{*}{161.18/66.39/48.42}} \\ \hline
                                & \multicolumn{1}{c|}{-}                             & \multicolumn{1}{c|}{-}                             & \multicolumn{1}{c|}{-}                             & \multicolumn{1}{c|}{-}                             & \multicolumn{1}{c|}{-}                             & \multicolumn{1}{c|}{-}                             & \multicolumn{1}{c|}{-}                             & \multicolumn{1}{c|}{100}                           & \multicolumn{1}{c|}{-}                             & -                             & \multicolumn{3}{c|}{}                                                   & \multicolumn{3}{c|}{}                                     \\ \cline{2-11}
\multirow{-2}{*}{\textbf{10}}   & \multicolumn{1}{c|}{\cellcolor[HTML]{C0C0C0}-}     & \multicolumn{1}{c|}{\cellcolor[HTML]{C0C0C0}-}     & \multicolumn{1}{c|}{\cellcolor[HTML]{C0C0C0}-}     & \multicolumn{1}{c|}{\cellcolor[HTML]{C0C0C0}-}     & \multicolumn{1}{c|}{\cellcolor[HTML]{C0C0C0}-}     & \multicolumn{1}{c|}{\cellcolor[HTML]{C0C0C0}-}     & \multicolumn{1}{c|}{\cellcolor[HTML]{C0C0C0}-}     & \multicolumn{1}{c|}{\cellcolor[HTML]{C0C0C0}100}   & \multicolumn{1}{c|}{\cellcolor[HTML]{C0C0C0}-}     & \cellcolor[HTML]{C0C0C0}-     & \multicolumn{3}{c|}{\multirow{-2}{*}{{\ul{0/0/0}}}}             & \multicolumn{3}{c|}{\multirow{-2}{*}{186.32/99.9/93.16}}  \\ \hline
\end{tabular}
}
\label{tab:without_calibrator}
\end{table*}

\end{document}